\documentclass[review]{elsarticle}

\usepackage{hyperref}

\journal{Journal of \LaTeX\ Templates}

\usepackage{graphicx}
\usepackage[cmex10]{amsmath}
\usepackage{array}
\usepackage{booktabs}
\usepackage{multirow}
\usepackage{graphicx,epstopdf,epsfig}
\graphicspath{{figures/}}
\DeclareGraphicsExtensions{.eps}
\usepackage{url}
\usepackage{float}
\usepackage[caption=false,font=footnotesize,labelfont=sf,textfont=sf]{subfig}

\begin{document}

\begin{frontmatter}

\title{A Directionally Selective Neural Network with Separated ON and OFF Pathways for Translational Motion Perception in a Visually Cluttered Environment}

\author[add1]{Qinbing Fu}
\ead{qifu@lincoln.ac.uk}
\author[add1]{Nicola Bellotto}
\ead{nbellotto@lincoln.ac.uk}
\author[add1]{Shigang Yue\corref{cor1}}
\ead{syue@lincoln.ac.uk}
\cortext[cor1]{Corresponding author}
\address[add1]{Lincoln Centre for Autonomous Systems (L-CAS), University of Lincoln, United Kingdom}

\begin{abstract}
Extracting useful motion cues from complex and dynamic scenes, both in an efficient and robust manner, is still a pronounced challenge for building artificial motion sensitive systems. Contrary to conventional computer vision methodologies, visual processing mechanisms in animals such as insects, may provide very simple and effective solutions for motion detection. With respect to biological findings underlying fly's physiology in the past decade, we present a directionally selective neural network (DSNN), with a feed-forward structure and entirely low-level visual processing, so as to implement direction selective neurons (DSNs) in the fly's visual system, which are mainly sensitive to wide-field translational movements in four cardinal directions. In this research, we highlight the functionality of ON and OFF pathways, separating motion information for parallel computation corresponding to light-on and light-off selectivity. Through this modeling study, we demonstrate several achievements compared with former bio-plausible translational motion detectors, like the elementary motion detectors (EMDs). First, we thoroughly mimic the fly's preliminary motion-detecting pathways with newly revealed fly's physiology. Second, we improve the speed response to moving dark/light features via the design of ensembles of same polarity (ON-ON/OFF-OFF) cells in the dual-pathways. Moreover, we alleviate the impact of irrelevant motion in a visually cluttered environment like the shifting of background and windblown vegetation, via the modeling of spatiotemporal dynamics. We systematically tested the DSNN against stimuli ranging from synthetic and real-world scenes, to notably a visual modality of a ground micro robot. The results demonstrated that the DSNN outperforms former bio-plausible translational motion detectors. Importantly, we verified its computational simplicity and effectiveness benefiting the building of neuromorphic vision sensor for robots.
\end{abstract}

\begin{keyword}
fly physiology\sep preliminary motion pathways\sep direction selective neurons\sep translational motion perception\sep ON and OFF pathways\sep spatiotemporal dynamics\sep neuromorphic sensor\sep neurons modeling
\end{keyword}
\end{frontmatter}


\section{Introduction}
\label{introduction}
Motion vision serves a wealth of daily tasks for animals and humans. For the vast majority of animals, a critically important feature of all visual systems is the detection and analysis of motion. Seeing the motion and direction in which a chased prey, a striking predator or a mating partner is moving, is of particular importance for the survival of any animal species. It is not only mammals but also insects that are competent in perceiving motion and distinguishing different classes of movements for decent visual course control, helping safe navigation through an environment. From biology to computational intelligence, the revealed internal neurons and mechanisms in animals' visual brains have provided us with a lot of inspirations for constructing artificial vision systems. In order for agents to initiate proper behaviors in complicated and dynamic environments, especially interacting with human hosts, a practical and robust motion-detecting system should possess the ability to extract meaningful motion cues from busy backgrounds in real time. Such an ability is of significance for both animals and intelligent machines like unmanned aerial vehicles, autonomous robots and also future robots, which are now playing crucial roles or may greatly influence our daily life in the near future.

For motion detection and estimation, there are many methodologies showing good performances. For instance, several 3-D motion segmentation based methods were proposed in the last two decades \cite{CVPR-2007}. Recently, monocular-vision based models and methods have demonstrated both high accuracy in the estimation of multi-body motion, including ego-motion and other independent motions, for example using a motion-segmentation strategy \cite{multi-motion-2016}, and good performance in the navigation control of quadrotors \cite{Quadrotor-2015,Quadrotor-2016,Quadrotor-2017}. In addition, new event-driven cameras \cite{ECCV-2016}, which directly report pixel-wise brightness changes instead of traditional intensity images, have been used for motion detection and tracking with clustering and learning algorithms in robotics \cite{EventCamera-icar-2017}. However, these segmentation, registration and learning based computer vision techniques are either computationally costly, or restricted to specific hardware that can not handle, with the degree of complexity required, real world scenarios for motion perception both in a cheap and robust manner.

In this article, we focus on neuromorphic solutions for building motion sensitive vision systems with relatively lower computational-consumption \cite{NeuroVisionSensor-2000,DeSouza_2002}. The biologically visual neural networks, have evolved and been tested over hundreds of millions of years, will be undisputedly forming solid modules to build artificial vision systems. As so far, they have provided a rich source of inspirations for perceiving motion fast and reliably, let alone its great potential in machine vision applications. Invertebrates in particular, using a relatively smaller amount of visual neurons compared to mammals and humans for motion detection, are attractive as sources of inspiration in recent decades for constructing a good number of motion detectors to simulate motion-detecting strategies, for example in insects like locusts \cite{LGMD1-1996,Yue_2013_dsn,LGMD1-Yue2006,LGMD1-DSN-competing,DSN-2007}, flies \cite{DSN-IJCNN,Harrison_2000,Iida_2000,Zanker_2005,Iida_2003,Huber_1999,Franceschini_2004,BackgroundMotion-ICDL,Wiederman-2008,STMD-IJCNN,ROBIO-2017} and reviewed by \cite{Borst_1989,Borst_2011,Borst-common,Borst-2014,Serres-2017,FlyingInsects-2010}.

In the insects' visual systems, it is believed that various groups of neurons possess specialized functionality for perceiving different motion cues, which can further act together to fuse various motion features. Different identified visual neurons, each with specifically physiological properties, motivated the creation of unique computing efficient neural networks. For instance, two lobula giant movement detectors (LGMDs) in the locust's visual system, namely LGMD1 and LGMD2, were implemented for quick and robust looming (collision) detectors in ground-vehicle scenarios \cite{LGMD1-car2006,LGMD-car-2017}, and realized as neuromorphic vision sensors for robots \cite{LGMD1-robot2005,LGMD1-Glayer,LGMD1-nonlinear,IROS-LGMDs,LGMD2-Fu,LGMD2-BMVC,Colias-Hu,LGMD1-robot2010}. The optical flow-based collision avoidance systems were widely used in near-range navigation of flying robots, e.g. \cite{OpticFlow-2008,OpticFlow-2015,Serres-2017}, which were motivated by the elementary motion detectors (EMDs) in the fly's visual system. In addition, another group of neurons, i.e. the small target motion detectors (STMDs), were revealed specific sensitivity to movements caused by dark objects with a very small or limited size, and implemented as artificial motion detectors as well, e.g. \cite{Wiederman-2008,STMD-2011,STMD-IJCNN}.

In this article, we present a visual neural network for the purpose of studying a specific group of neurons, so-called direction selective neurons (DSNs) in the fly's preliminary motion-detecting pathways, which are mainly sensitive to wide-field translational motion in a visual field. Over hundreds of millions of years of evolution, it is no surprise that DSNs were found across the animals kingdom that has been studied so far, not only in insects like locusts \cite{Rind-DSN1990a,Rind-DSN1990b} and flies \cite{Buchner_1976,Hassenstein_1956,Borst_2011,Borst-2002}, but also in mammals like rabbits \cite{Barlow_1965}, cats \cite{Cat-1998} and mice \cite{Borst-common}. In the last decade, with developments of physiological techniques, much progress has been made by biologists underlying the fly's preliminary motion-detecting pathways, \cite{Borst_2011,Borst-common,Joesch_2010,Circuit-correlation2013,Eichner_2011,Clark_2011,Vogt-2007,Joesch_2013,Maisak_2013,Rister-2007,Gabbiani_2011,James_2014,Leonhardt-2016,CommonEvo-2015}, yet despite this the fundamental cellular implementation still remains mysterious. Therefore, our computational modeling experience may provide useful insights into underlying mechanisms and circuits, and may give new biological hypotheses.
\begin{figure}[t]
	\centering
	\includegraphics[width=0.6\linewidth]{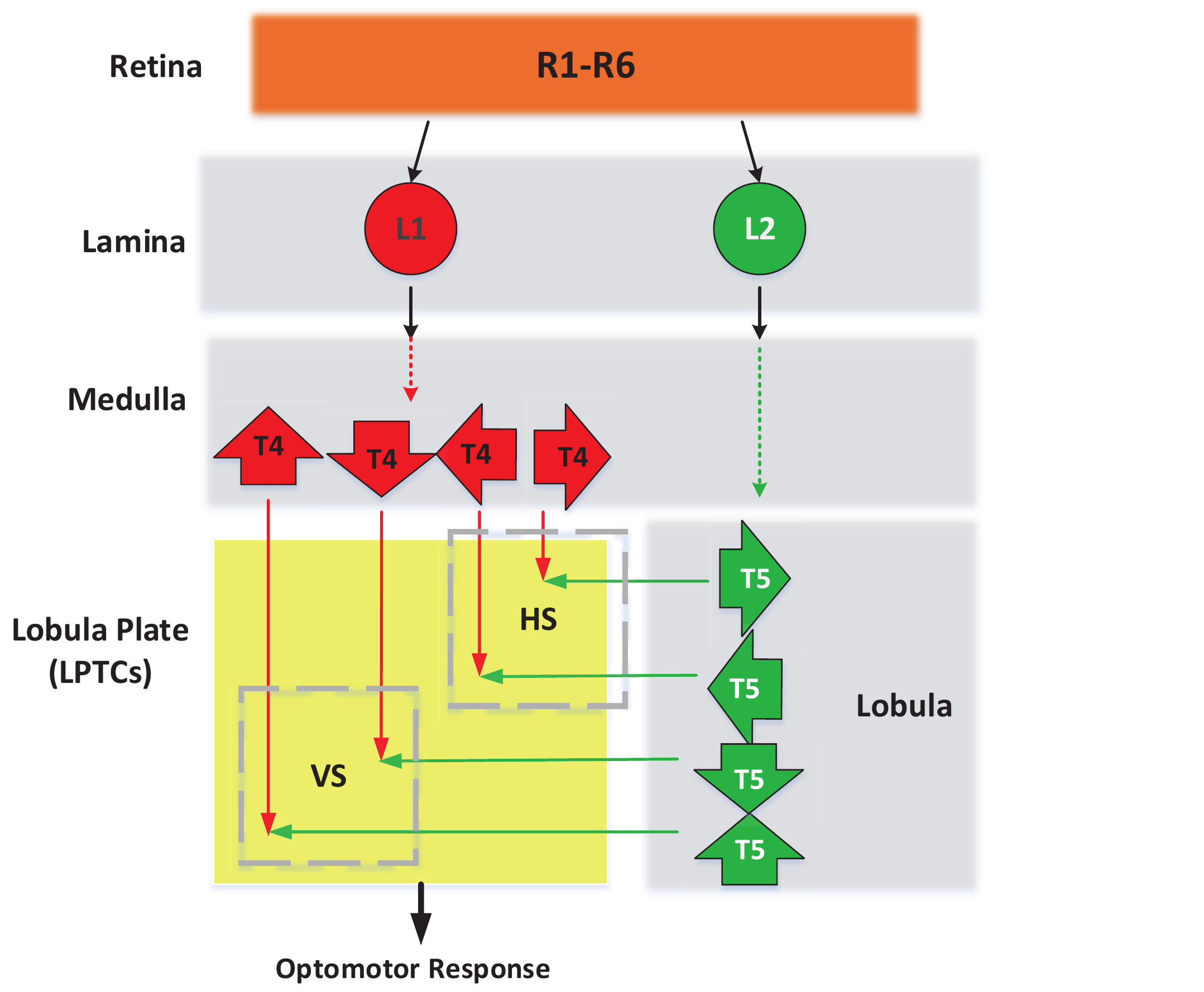}
	\caption{A diagram of fly's preliminary motion-detecting pathways, throughout five neuropile layers: the first retina layer with R1-R6 neurons denotes photoreceptors, which convey motion information to lamina monopolar cells (L1 and L2). Visual signals are thereby split into parallel ON and OFF pathways, which are indicated by red and green arrows respectively. The direction selectivity of motion information is generated in the medulla and lobula layers tuned by four groups of \textbf{directionally specific} T4 and T5 neurons. The lobula plate tangential cells (LPTCs) pool each group of directionally specific neural response to form the horizontally sensitive (HS) and vertically sensitive (VS) systems. The outputs of these two systems guide behavioral optomotor responses.}
	\label{physiology}
\end{figure}

An important biological theory guiding the proposed DSNN modeling is that visual information is separated into ON and OFF pathways for parallel computation, as shown in Fig. \ref{physiology}. The onset and offset responses, evoked by luminance increments and decrements, are conveyed to the medulla and lobula layers by the ON and OFF rectifying transient cells (RTCs) in the lamina layer. More importantly, the \textbf{direction selectivity} to ON-edge and OFF-edge movements is encoded and formed in the medulla and lobula layers. Finally, in the lobula plate layer, the lobula plate tangential cells (LPTCs) pool directionally selective motion from four groups of direction selective T4 and T5 neurons in the medulla and lobula layers to form two directionally sensitive systems, i.e., the horizontally sensitive (HS) and the vertically sensitive (VS) systems. Interestingly, both systems respond to visual motion with fixed preferred and non-preferred (or null) directions regardless of color or contrast of both the visual stimuli and the background \cite{Joesch_2010}. More specifically, DSNs are rigorously activated by motion along the rightward and downward, i.e., preferred directions, while inhibited by motion along the leftward and upward -- null directions \cite{Joesch_2010}. Despite that, how the direction selectivity forms in the dual-pathways is still controversial \cite{Joesch_2013,Gabbiani_2011}.

\begin{figure}[t]
	\begin{minipage}[t]{0.32\linewidth}
		\centering
		\centerline{\includegraphics[width=0.7in]{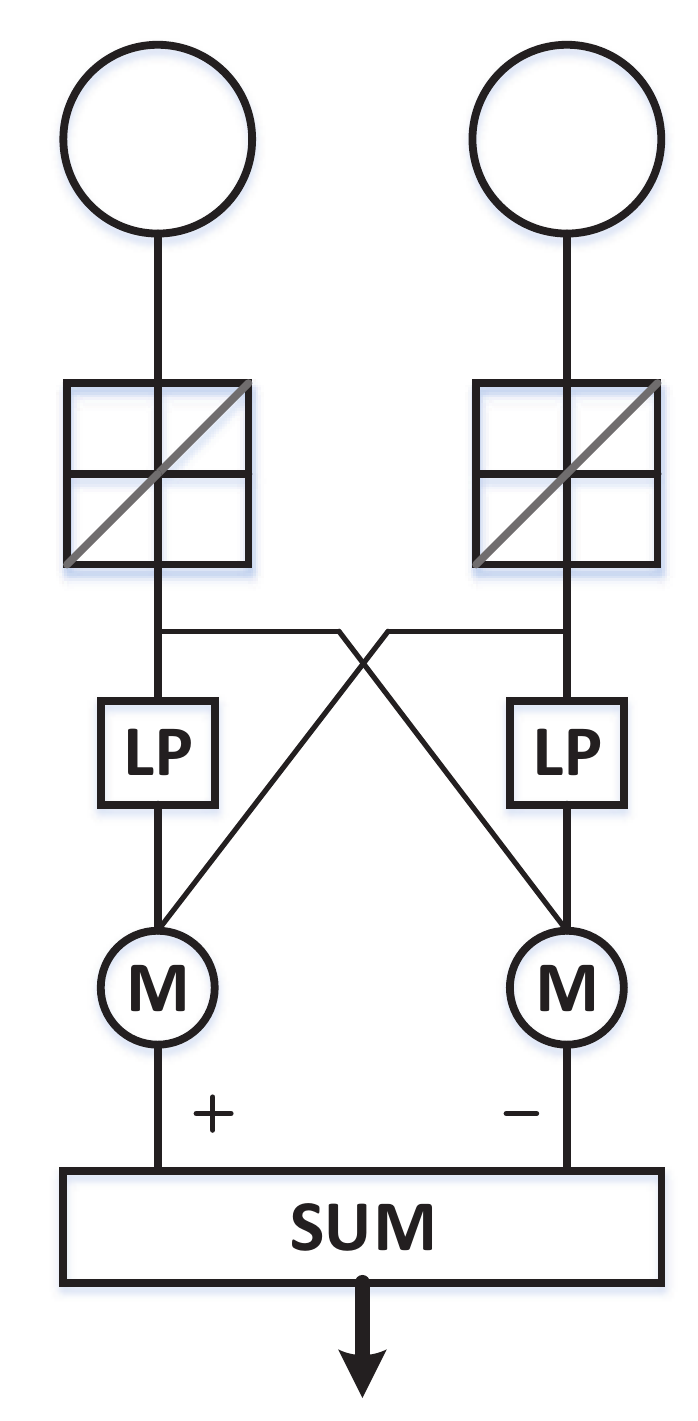}}
		\centerline{\scriptsize(a)}
	\end{minipage}
	\hfill
	\begin{minipage}[t]{0.32\linewidth}
		\centering
		\centerline{\includegraphics[width=1.5in]{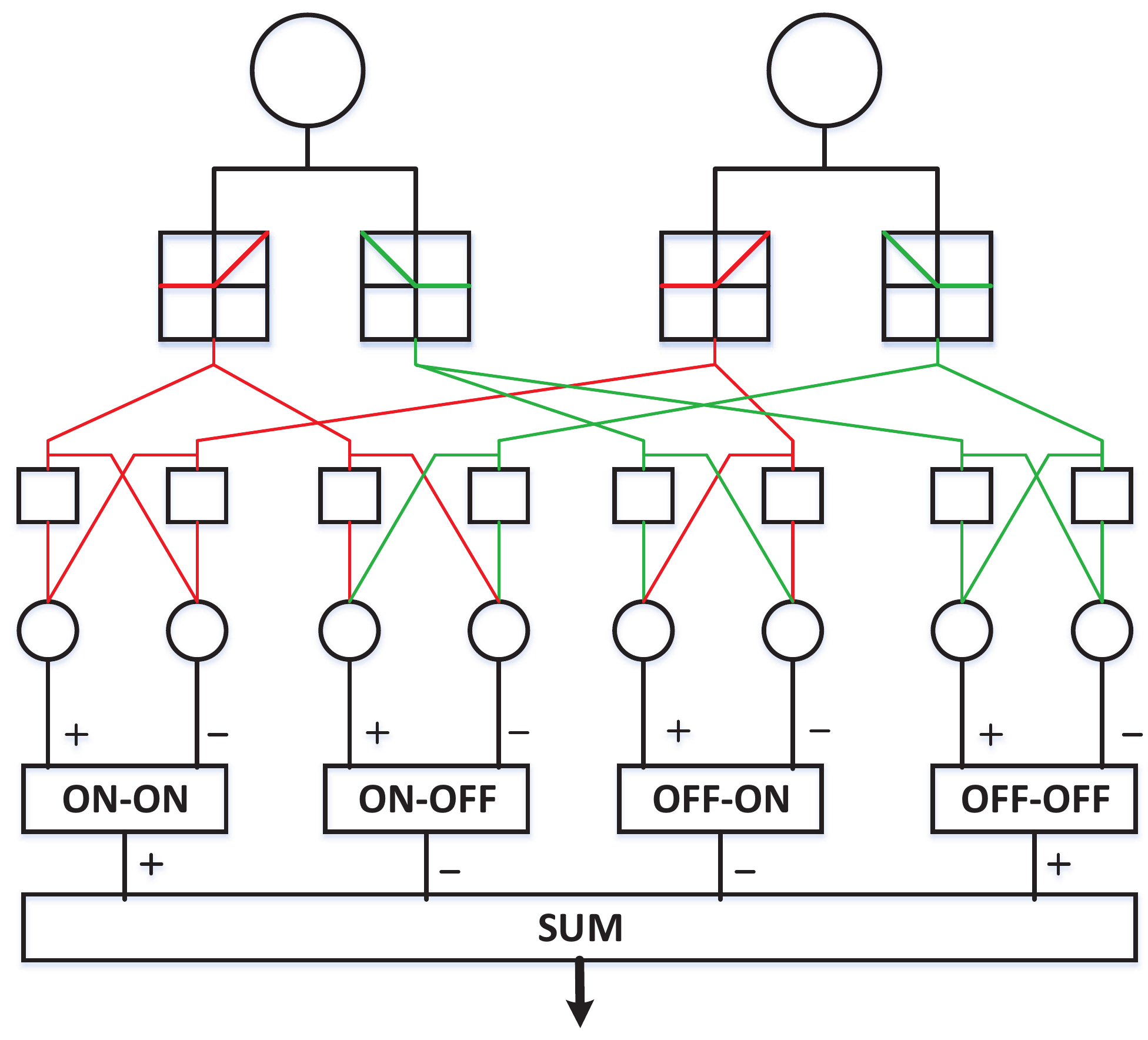}}
		\centerline{\scriptsize(b)}
	\end{minipage}
	\hfill
	\begin{minipage}[t]{0.32\textwidth}
		\centering
		\centerline{\includegraphics[width=1.2in]{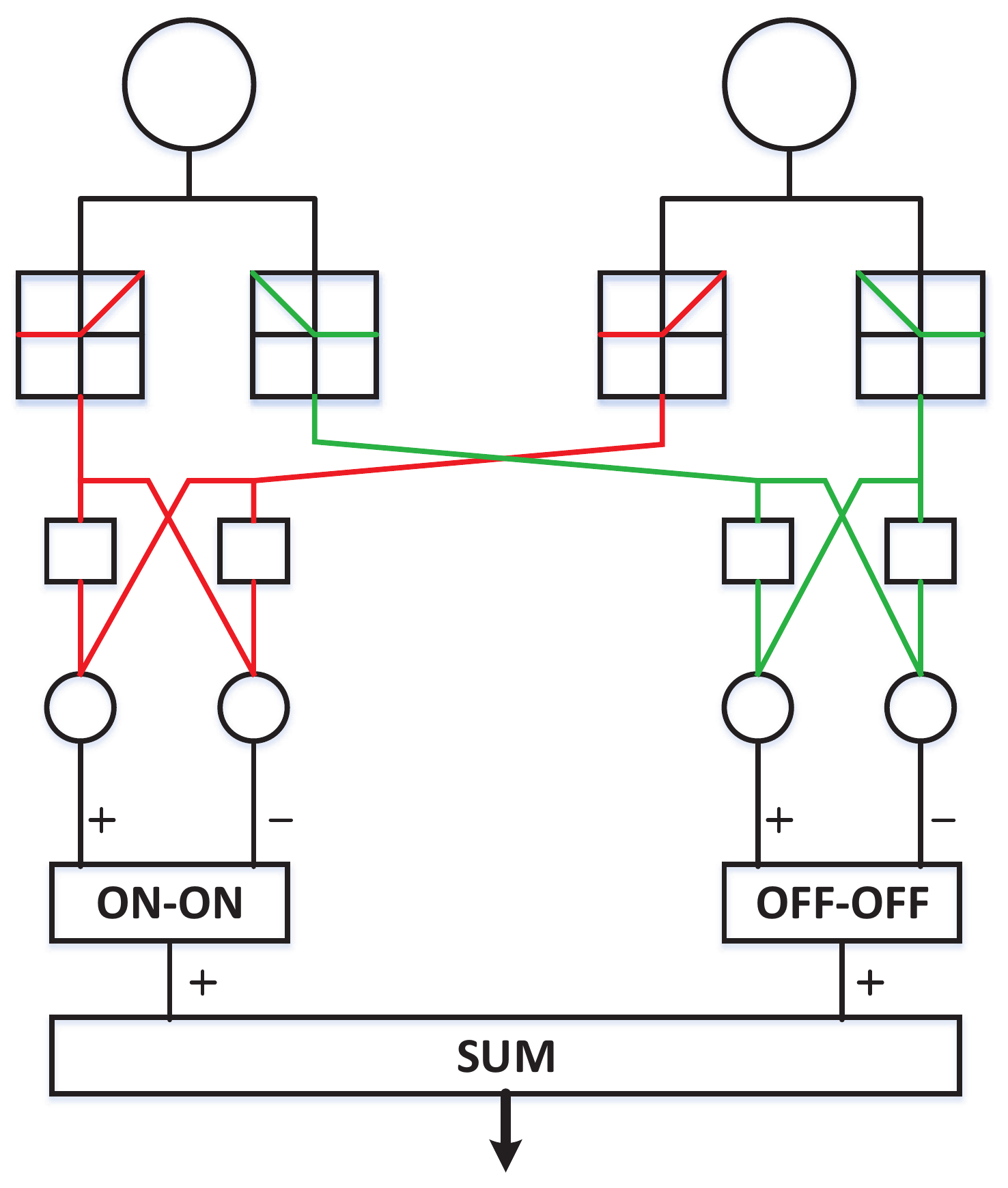}}
		\centerline{\scriptsize(c)}
	\end{minipage}
	\caption{Different biological models for motion detection: (a) The symmetric Reichardt detectors (EMDs) are revealed as a standard model for motion detection in insects. LP and M components indicate the low-pass filtering and multiplication. (b) The four-quadrant model processes input combinations of ON-ON, ON-OFF, OFF-ON, and OFF-OFF. Each combination replicates the structure of Reichardt detectors. This model is mathematically identical to the original Reichardt detectors. (c) The two-quadrant model processes only input combinations of the same sign (ON-ON, OFF-OFF).}
	\label{motion-models}
\end{figure}

With regard to our current understandings, visual neurons compute the direction of motion corresponding to the well-known Hassenstein-Reichardt Correlation (HRC) model (referred as `Reichardt detectors') \cite{Hassenstein_1956}, which mathematically explains the way of mapping nonlinear algorithm onto neuronal hardware and being implemented by neural networks (Fig. \ref{motion-models}(a)). There have been quite a lot of studies coming from such a `correlation-type' motion detector, like the elementary motion detectors (EMDs) with a symmetric structure of Reichardt detectors \cite{Borst_1989}, and its derivative models (e.g. \cite{EMD-MainArticle,Iida_2000,Iida_2003,Zanker_1999,Zanker_2005,Zanker_1996,Frye2015Elementary}) with extra spatiotemporal filters. Since biological studies demonstrate that visual signals are already directionally selective before collectively arriving at LPTCs in flies \cite{Maisak_2013}, the computational architecture of HRC detectors well explains the forming of directions in the medulla and lobula neuropil layers, which are locationally prior to the lobula plate layer in the fly's visual circuits. However, in former HRC-based computational models, visual signals are processed only in a single pathway, unlike the illustrated fly's physiology in Fig. \ref{physiology}. It appears that the separated ON/OFF pathways are playing crucial roles in the underlying preliminary motion-detecting circuits. To fill this gap, we construct fly's preliminary visual system thoroughly (layer-by-layer) to demonstrate its characteristic and significance for wide-field translational motion perception.

There are several biological models arguing for different motion detection strategies with different combinations of ON and OFF RTCs in the dual-pathways \cite{Eichner_2011,Clark_2011,Joesch_2013,Gabbiani_2011}. The first assumption is the \textbf{four-quadrant} detectors with communications between both the same and opposite polarity cells (Fig. \ref{motion-models}(b)). Technically speaking, it mathematically conforms to the symmetric HRC-based model with a single pathway \cite{Hassenstein_1956,Eichner_2011}. The second important speculation is the \textbf{two-quadrant} model that was in accordance with electro-physiological recordings from LPTCs \cite{Eichner_2011}. Contrary to the four-quadrant model, it processes only input combinations of the same sign signal, i.e., ON-ON and OFF-OFF, as shown in Fig. \ref{motion-models}(c). In the biological study \cite{Eichner_2011}, the two-quadrant instead of the four-quadrant model was recommended to exist in the motion-detecting circuitry via physiological tests.

Furthermore, there is another biological model based on behavioral experiments, which supports that either ON/OFF pathways convey motion information about both positive and negative contrast changes in the motion-detecting circuitry \cite{Clark_2011}. In this research, a framework of \textbf{six-quadrant} detectors was proposed with interactions between both polarity cells in either pathways. Compared with the structure of four-quadrant model (Fig. \ref{motion-models}(b)), it also processes light-off (OFF-OFF) response in the ON pathway and light-on (ON-ON) response in the OFF pathway. More importantly, this biological model further emphasizes the importance of edge selectivity in motion detection.

Given these biological motion detectors, the different combinations of ON and OFF transient cells all depict a picture of how the fly's neural circuits implement the Reichardt detectors to shape the direction selectivity. To decide among these alternatives, a subsequent research provided strong evidence of the existence of \textbf{two-quadrant} versus six-quadrant motion detectors for producing the directional signals, via genetically blocking either ON or OFF pathways \cite{Joesch_2013}. In the proposed DSNN modeling, we were consistent with the combinations of only same sign polarity cells, i.e., the two-quadrant model, for guiding the neural computations within the separated ON/OFF pathways. We also demonstrated the significance of edge selectivity to movements of ON-edges and OFF-edges, especially in a visually cluttered environment.

A shortcoming or unsolved problem of former HRC-based models is the speed tuning of motion detection. In other words, a biological motion-detecting circuit may not tell the true velocity of stimuli \cite{Frye2015Elementary}. The reason is that for each combination of such `delay and correlate' motion detectors, it is advisable to decide the spacing between each pairwise detectors, and the time span for the delay in follow-up nonlinear computation, each factor of which will affect the model's performance \cite{Zanker_1999}. For example, perceiving faster movements requires a larger spatial span between detectors if fixing the delay; otherwise, it requires a shorter delay when the spacing is unchanged. In this research, we found that building an \textbf{ensemble of motion detectors}, by directionally connecting multiple same-sign polarity cells, has great potential of improving the speed response of motion sensitive neural networks, even though it costs more computational energy. With this idea, we can pre-define the sampling distance in each pairwise combination of same sign motion detectors in either ON/OFF pathways, as well as the number of connected cells for each local cell for speed tuning. We also investigated the modeling of temporal dynamics within the directionally lateral interactions of ON-ON/OFF-OFF motion detectors. We found that a \textbf{dynamically temporal filtering} strategy for combinations of detectors with different spacings improves the velocity sensitivity to translational motion, as presented in a related model from our recent work \cite{DSN-IJCNN}.

Another defect of former HRC-based models is that they lack robust mechanisms for filtering out irrelevant motion from visual clutter, so that they are easily influenced by environmental motion such as the windblown vegetation, as well as the shifting of background or surroundings caused by ego-motions. Compared to various kinds of physical sensors like infrared, ultrasound, radar and laser, much richer environmental information is gathered by the visual sensing modality. As a result, how motion vision systems filter out irrelevant motion from relevant motion is still significantly challenging the computational modeling of biological neural networks, i.e, it appears that more robust motion filters are badly needed. Motivated by some physiological researches or models (e.g. \cite{Clark_2011,Circuit-correlation2013,fly-1967,Borst_1989,Wiederman-2008}), our previous research demonstrated the effectiveness of a spatial pre-filtering of motion signals prior to the dual-pathways (in the lamina layer), which can maximize the transmission of useful motion cues along with removing redundant environmental noise in a cluttered background \cite{DSN-IJCNN}. However, this research only investigated motion perception against stationary backgrounds. We found that when challenged by the shifting of a visually cluttered background, the situations which happen very frequently during navigation, signal pre-filtering only \textbf{in spatial} dose not fully satisfy the requirements of a robust motion vision system. In this research, we also incorporated in the DSNN a bio-plausible mechanism within the dual-pathways, which can further filter out irrelevant motions \textbf{in temporal}.

More details of the novelty in the architecture design of DSNN will be illustrated fully in the next section. We hope this study will provide useful conclusions and suggestions for mimicking animals' preliminary motion-detecting circuits, designing robust and efficient motion sensitive vision systems, and exploring the potential of bio-plausible neural networks in future intelligent robots and other application domains. The rest of paper is organized as follows: in Section \ref{network}, we illustrate the methodology; in Section \ref{expriments}, we present the systematic experiments with analysis and discussion; in Section \ref{conclusion}, we conclude this article and give future research directions.

\section{Framework of the directionally selective neural network}
\label{network}
In this section, we introduce the proposed directionally selective neural network (DSNN) fully. A key feature of the functionality of DSNN is that it is only sensitive to translating motion cues in four cardinal directions, unlike a few collision-detecting neural networks based on neurons in locusts' visual pathways, i.e., LGMD1 (e.g. \cite{LGMD1-Glayer,LGMD1-nonlinear,LGMD1-1996,Colias-Hu}), LGMD2 (\cite{LGMD2-BMVC,LGMD2-Fu}), DSNs (e.g. \cite{DSN-2007,Yue_2013_dsn}), as well as their combinations (e.g. \cite{IROS-LGMDs,LGMD1-Yue2006,LGMD1-DSN-competing}).

\begin{figure}[!t]
	\begin{minipage}[t]{0.59\linewidth}
		\centering
		\centerline{\includegraphics[width=2.5in]{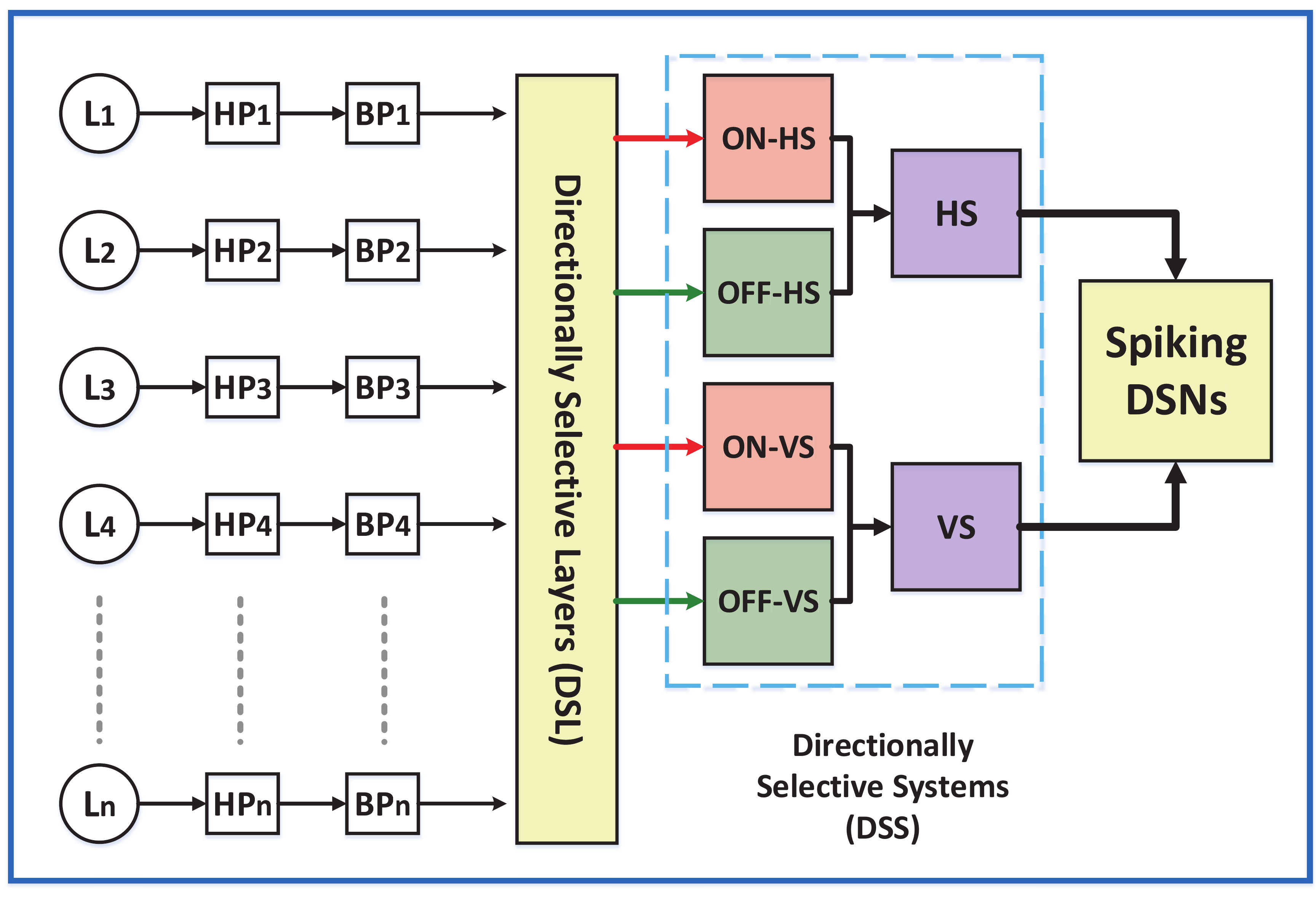}}
		\centerline{\scriptsize(a) a schema of DSNN}
	\end{minipage}
	\hfill
	\begin{minipage}[t]{0.39\linewidth}
		\centering
		\centerline{\includegraphics[width=1.8in]{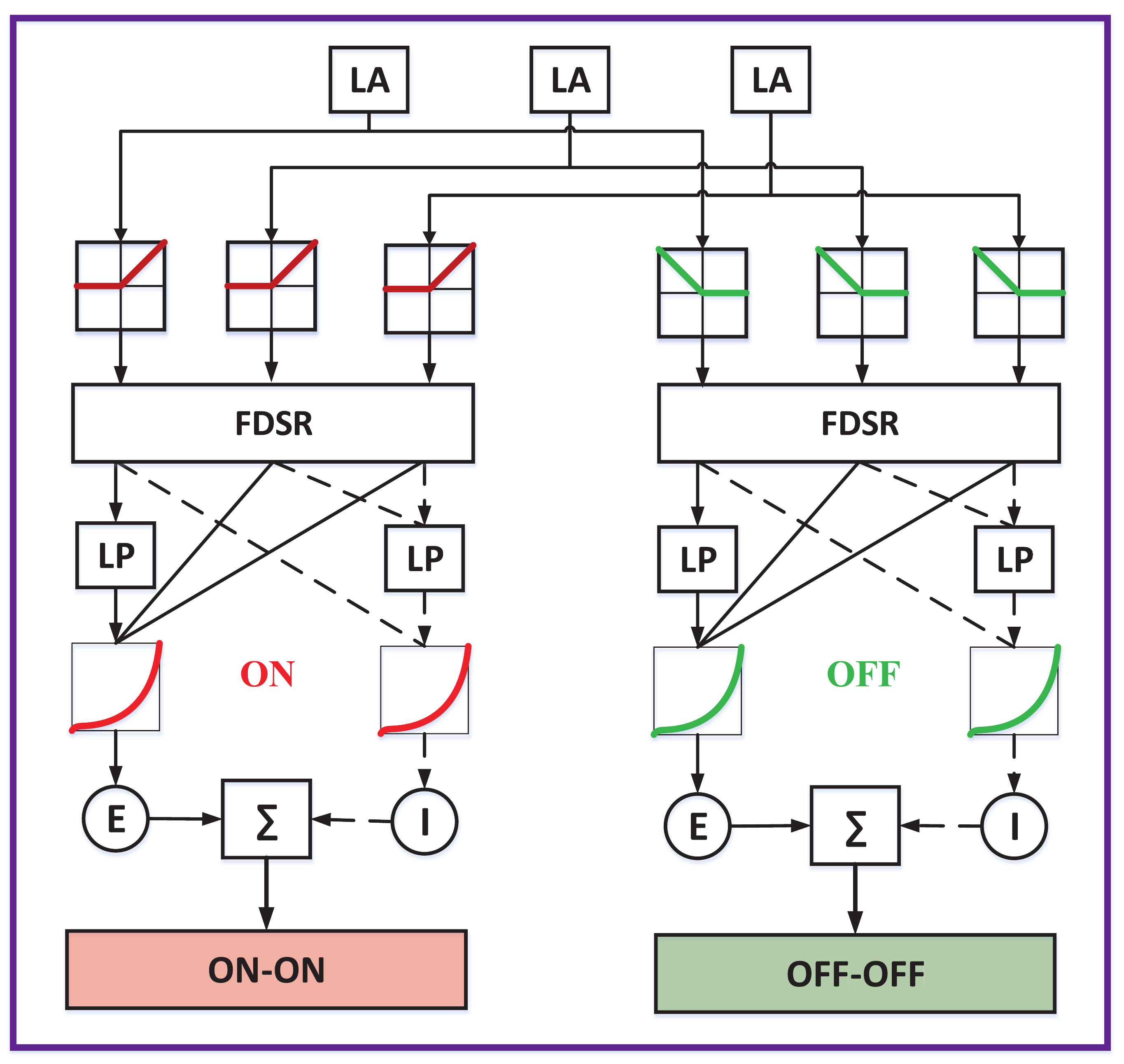}}
		\centerline{\scriptsize(b) a schema of DSL in DSNN}
	\end{minipage}
	\vfill
	\vspace{0.06in}
	\begin{minipage}[t]{0.49\linewidth}
		\centering
		\centerline{\includegraphics[width=2.1in]{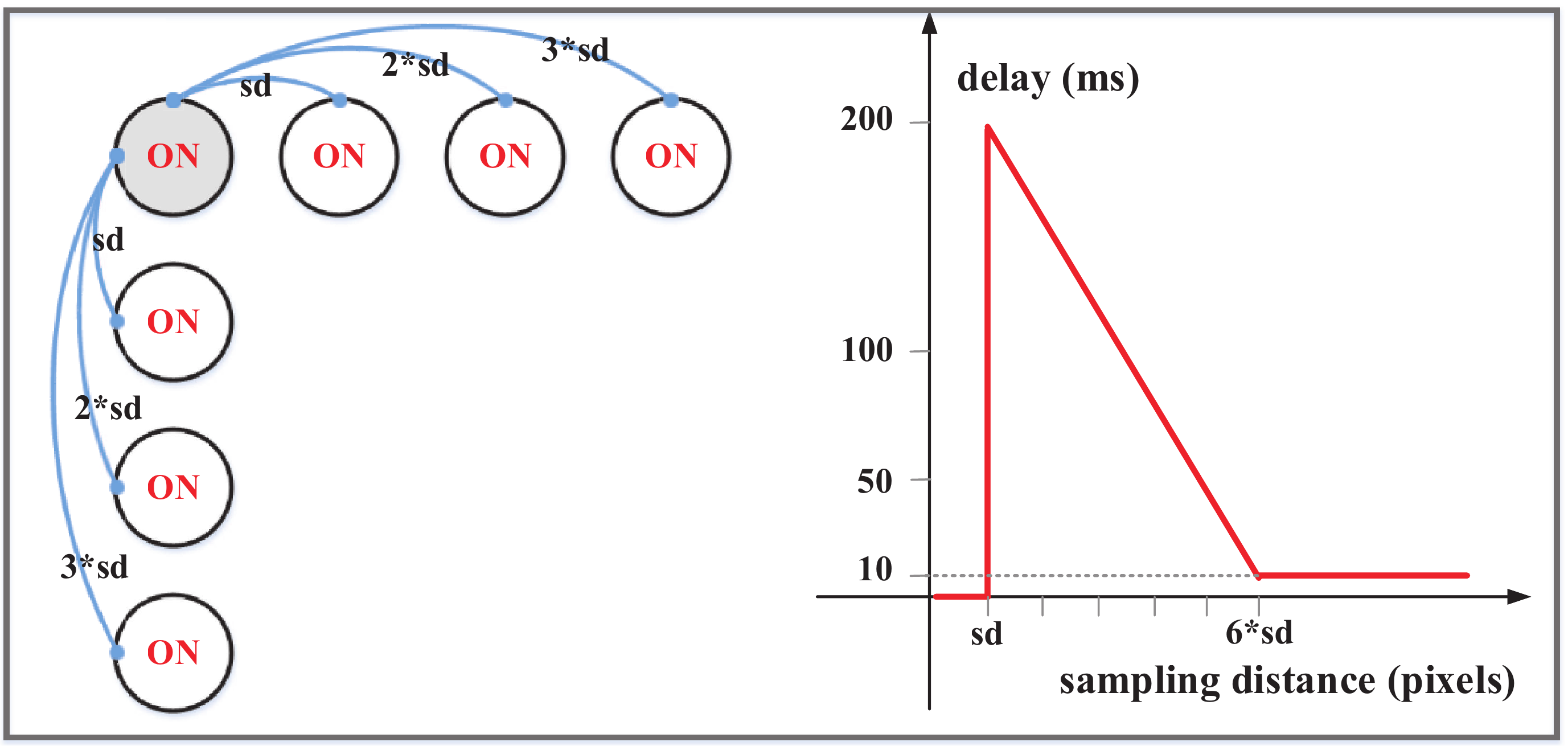}}
		\centerline{\scriptsize(c) spatiotemporal dynamics in DSL}
	\end{minipage}
	\hfill
	\begin{minipage}[t]{0.49\linewidth}
		\centering
		\centerline{\includegraphics[width=2.1in]{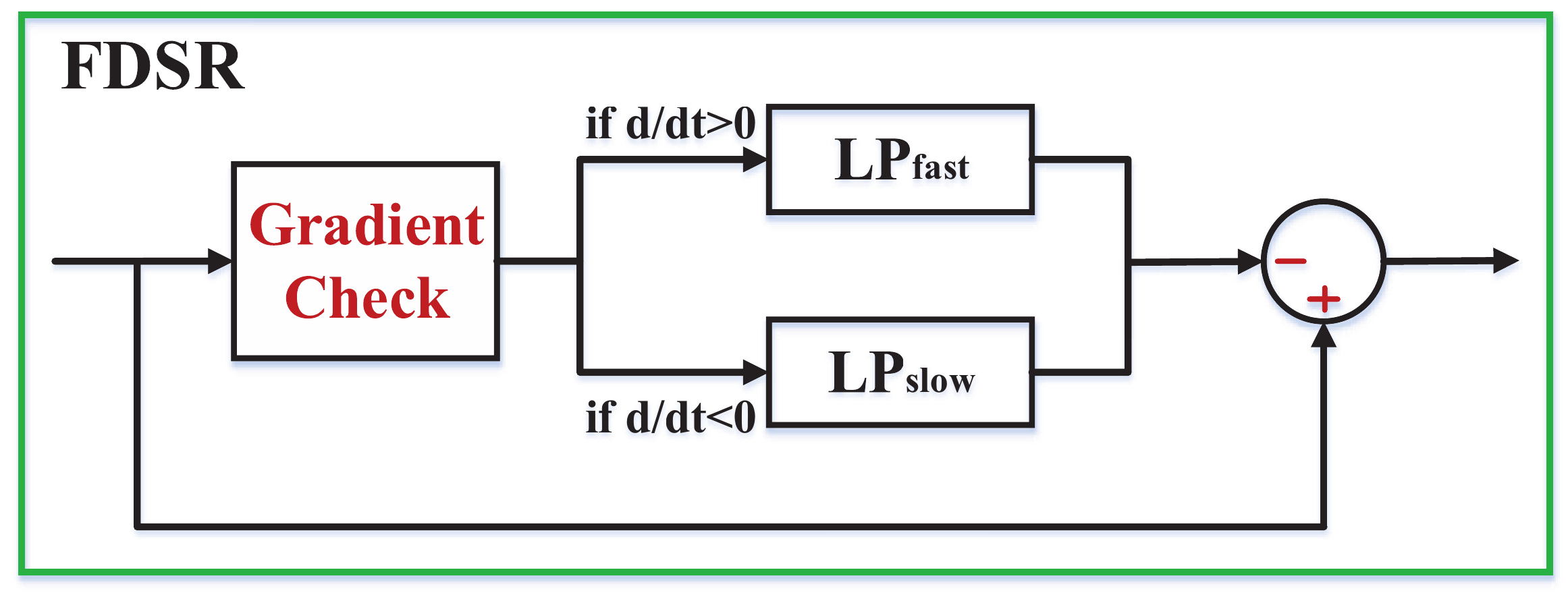}}
		\centerline{\scriptsize(d) FDSR mechanism}
	\end{minipage}
	%
	\caption{The schematic of DSNN: (a) A general signal processing and filtering flowchart: the motion information after spatial band-pass filtering goes through \textbf{directionally selective layers} (DSL) including ON and OFF pathways to form two flows of \textbf{directionally selective systems} (DSS), which map neural response to spikes. (b) A schema of DSL in DSNN taken three connected lamina cells for illustration: each pairwise interaction of ON-ON or OFF-OFF polarity cells matches the computation of a pairwise and symmetric Reichardt detectors. (c) The spatial multi-connections of ON cells for each local cell \textbf{in two directions} and the delay function for combinations with different sampling distances in DSL; similarity for connections of OFF cells.  (d) The temporal mechanism of FDSR in both ON and OFF channels in DSL. The full-names of abbreviated model components can be found in Table \ref{tableAcr}.}
	\label{dsn}
\end{figure}

In general, there are five computational neuropile-layers constituting the motivated motion-detecting pathways for mimicking DSNs in the fly's visual system. The core structure of DSNN is the separated ON and OFF pathways splitting motion information for parallel computations, encoding selectivity to moving ON-edges and OFF-edges respectively. Compared to other bio-plausible translational motion detectors, we also apply spatiotemporal filtering both prior to and within the dual-pathways, in order to achieve more robust motion detection performance even in a visually dynamic and cluttered environment. It is also worth clarifying that the whole motion-detecting pathways possess a completely feed-forward structure, and only uses low-level image processing methods, whereby those computationally expensive algorithms for objects classification, scene or activity analysis and parameters learning are hardly needed in the current DSNN modeling. It perceives motion by reacting to moving ON-edges and OFF-edges; both its computational simplicity and efficiency shed lights to build neuromorphic vision sensors for autonomous robots. A schema of DSNN along with the underlined dual-pathways and spatiotemporal mechanisms is illustrated in Fig. \ref{dsn}. The full-names of abbreviated model components in Fig. \ref{dsn} can be found in Table \ref{tableAcr}. The predefined neural network parameters are listed in Table \ref{tableParams}. 

\subsection{The computational retina layer}
In the first computational layer, there are photoreceptors arranged in a two-dimensional matrix form, which capture gray-scaled and pixel-wise luminance from video clips or visually sensing modality of robots. The brightness obtained by photoreceptors goes through a first-order high-pass filtering (HP in Fig. \ref{dsn}(a)) in order to get moving features by the differential image between every two successive frames:
\begin{equation}
P(x,y,t) = L(x,y,t) - L(x,y,t-1) + \sum_{i}^{N_p}a_i \cdot P(x,y,t-i)
\label{high-pass}
\end{equation}
where $P(x,y,t)$ is the change of luminance according to each local pixel at frame $t$. $x$ and $y$ are the abscissa and ordinate in the visual field. $L(t)$ and $L(t-1)$ are the brightness of two successive frames. The luminance change could last and decay for a short while: $N_p$ indicates the total number of frames constituting the duration of residual visual information, and the coefficient $a_i$ is defined by $a_i = (1 + e^{u \cdot i})^{-1}$ wherein $u = 1$: increasing $u$ leads to faster decay of remaining luminance change.
\begin{table}[t]
	\centering
	\caption{The DSNN filters and components in Fig. \ref{dsn}}
	\begin{tabular}{l|l|l|l}
		\toprule
		\multicolumn{4}{c}{acronym and full-name}\\
		\cmidrule{1-4}
		L	   &gray-scaled luminance&n     &number of photoreceptors\\
		HP     &high-pass filter    &FDSR  &fast depolarizing slow repolarizing\\
		LP     &low-pass filter     &HS    &horizontally sensitive system\\
		BP     &band-pass filter    &VS    &vertically sensitive system\\
		E/I    &excitation/inhibition&LA    &lamina monopolar cell\\
		\bottomrule
	\end{tabular}
	\label{tableAcr}
\end{table}

\subsection{The computational lamina layer}
After that, as depicted in Fig. \ref{dsn}(a), we apply a spatial band-pass (BP) filtering for motion features, which is mathematically represented by a two-dimensional form of `Difference of Gaussians' (DoGs) algorithm, so as to enhance the underlined edge selectivity in the motion-detecting circuitry, and maximize information transmission by spatially removing redundant environmental noise. Such a mechanism embodies the biological functions of large monopolar cells in the lamina neuropil layer, which was considered a suitable filter prior to the site of motion detection in insects' vision system \cite{fly-1967,LGMD1-nonlinear,Wiederman-2008}. With this mechanism, we can realize the center-surrounding antagonism for each local lamina cell, with the center-positive and surrounding-negative Gaussians representing the excitatory and inhibitory fields respectively:
\begin{equation}
P_{e}(x,y,t) = P(x,y,t) \overset{x,y}{*} G_{\sigma_e}(x,y),\ 
P_{i}(x,y,t) = P(x,y,t) \overset{x,y}{*} G_{\sigma_i}(x,y)
\label{two-gauss}
\end{equation}
where $\overset{x,y}{*}$ indicates the convolution at local cell $(x,y)$ in the visual field, $\sigma_e$ and $\sigma_i$ indicate the excitatory and inhibitory standard deviations. $G$ is the convolution kernel, which satisfies with a two-dimensional Gaussian distribution:
\begin{equation}
G_{\sigma}(x,y) = \frac{1}{2\pi \sigma^2}\ exp(-\frac{x^2 + y^2}{2 \sigma^2})
\label{gauss-function}
\end{equation}
Therefore, the center-surround field is created by having each point to be the weighted average of the points surrounding it, and the weightings take a form of two Gaussian distributions respectively. In the DoGs algorithm, the broader inhibitory Gaussian is subtracted from the narrower excitatory one, along with the polarity selectivity to fit the functionality of the following first-order ON and OFF RTCs:
\begin{equation}
LA(x,y,t) = \left\{
\begin{aligned}
|P_{e}(x,y,t) - P_{i}(x,y,t)|,\ &\text{if}\ P_{e}(x,y,t) \geq 0\ \&\ P_{i}(x,y,t) \geq 0\\
-|P_{e}(x,y,t) - P_{i}(x,y,t)|,\ &\text{if}\ P_{e}(x,y,t) < 0\ \&\ P_{i}(x,y,t) < 0
\end{aligned}
\right.
\label{polar-dogs}
\end{equation}
The RTCs split spatially filtered signals into separated ON/OFF channels, encoding light-on and light-off responses in ON and OFF pathways respectively. Technically speaking, such neural mechanisms fulfill the 'half-wave' rectifiers (Fig. \ref{motion-models}(b), (c) and Fig. \ref{dsn}(b)), filtering out negative and positive input for ON and OFF channels respectively, as well as inverting negative information for OFF channels. Each lamina monopolar cell corresponds to a pairwise ON and OFF RTCs:
\begin{equation}
\begin{aligned}
&LA_{on}(x,y,t) = (LA(x,y,t) + |LA(x,y,t)|) / 2 + \sigma_{l} \cdot LA_{on}(x,y,t-1)\\
&LA_{off}(x,y,t) = |(LA(x,y,t) - |LA(x,y,t)|)| / 2 + \sigma_{l} \cdot LA_{off}(x,y,t-1)
\end{aligned}
\label{halfwave}
\end{equation}
where $LA_{on}(x,y,t)$ denotes the ON cell value, and similarly for the OFF cell value. In addition, we allow a small fraction ($\sigma_{l}$) of original information in parallel to pass through, mimicking the residual visual information in the motion-detecting circuitry of insects \cite{Eichner_2011}.

For each independent polarity neuron, an `adaptation state' is formed by a biologically plausible mechanism, i.e., the temporal dynamics of `fast depolarizing slow repolarizing' (FDSR in Fig. \ref{dsn}(b)), which matches the neural characteristic of `fast onset and slow decay' phenomenons. As depicted in Fig. \ref{dsn}(d), we do the gradient check for relayed signals from RTCs before the processing of low-pass filtering:
\begin{equation}
\frac{d D(x,y,t)}{d t} = \left\{
\begin{aligned}
&\frac{1}{\tau_{fast}}(LA^{'}(x,y,t) - D(x,y,t)),\ \text{if}\ \frac{d LA^{'}(x,y,t)}{d t} \ge 0\\
&\frac{1}{\tau_{slow}}(LA^{'}(x,y,t) - D(x,y,t)),\ \text{if}\ \frac{d LA^{'}(x,y,t)}{d t} < 0
\end{aligned}
\right.
\label{fdsr-lowpass}
\end{equation}
where $LA^{'}$ designates the input from either ON/OFF RTCs, and $D$ denotes the delayed polarity signals. Intuitively, if the gradient is nonnegative, we employ a very short delay -- $\tau_{fast}$ ($1$ millisecond in our case) -- realizing the `fast onset' response; otherwise, the delay is set to $100$ms for the `slow decay'. Because the digital signal does not have a continuous derivative, we do check the gradient through comparative analysis between discrete frames. After that, in the FDSR mechanism, the delayed signal is subtracted to the original passed one:
\begin{equation}
\begin{aligned}
&F_{on}(x,y,t) = LA_{on}(x,y,t) - D_{on}(x,y,t),\\
&F_{off}(x,y,t) = LA_{off}(x,y,t) - D_{off}(x,y,t)
\end{aligned}
\label{fdsr-subtraction}
\end{equation}
Such a mechanism contributes to temporally filter out irrelevant motion from relevant motion in dynamic and complex environments.

\subsection{The computational medulla and lobula layers}
Next, the medulla and lobula neuropil layers (Fig. \ref{physiology}) have been proposed to be the most likely places where neighboring interneurons interact with each other in a nonlinear way producing directionally selective signal to the following lobula plate \cite{Maisak_2013}. We computationally model these two layers as the directionally selective layers shown in Fig. \ref{dsn}. To be more specific, there are two kinds of flows -- excitation and inhibition (E, I in Fig. \ref{dsn}(b)) -- being generated in ON/OFF channels of the medulla/lobula layers respectively. Importantly, compared with the DSNs modeling works motivated by neurons in locusts' visual pathways \cite{DSN-2007,Yue_2013_dsn}, wherein the inhibitory connections are modeled in four or eight directions to generate the directionally selective information, we shape the directional tuning in the proposed DSNN via the mapping of connections of same-polarity (ON-ON/OFF-OFF) cells in only two orientations by similarly nonlinear computation of Reichardt detectors: the excitation and inhibitions form in the start and adjacent connected cells respectively (Fig. \ref{dsn}(c)).

Contrary to a number of EMDs-based models (e.g. \cite{Clark_2011,Eichner_2011,Borst_1989,Zanker_1996,Iida_2000,EMD-MainArticle}), we not only implemented the lateral multi-connections for each local cell in the computational medulla and lobula layers, but also adopt dynamically temporal filtering, wherein the delays vary in each directional combination of ON-ON/OFF-OFF motion detectors depending on different spacings and obey a linearly decaying function, as shown in Fig. \ref{dsn}(c). Such a structure has demonstrated great potential of enhancing the speed response to translational movements \cite{DSN-IJCNN}. Firstly, we illustrate calculations of the HS system for the ON pathway:
\begin{equation}
\begin{aligned}
&E_{on}^{hs}(x,y,t) = \sum_{i=d}^{d \cdot N_{con}}D_{on}(x,y,t) \cdot F_{on}(x+i,y,t),\\
&I_{on}^{hs}(x,y,t) = \sum_{i=d}^{d \cdot N_{con}}D_{on}(x+i,y,t) \cdot F_{on}(x,y,t),\\
&ME^{hs}(x,y,t) = E_{on}^{hs}(x,y,t) - w_{i} \cdot I_{on}^{hs}(x,y,t)
\end{aligned}
\label{medulla-HS-EI}
\end{equation}
where $N_{con}$ and $d$ designate the number of connected polarity cells and the increment of spacings in sampling distance respectively. $w_i$ is a local bias to form a partially balanced model with stronger response to the preferred directional motion. The delay function in either ON/OFF pathways conforms to Eq. \ref{fdsr-lowpass} -- a low-pass filtering -- with a dynamic time parameter $\tau_s$, which can vary from tens to hundreds of milliseconds, as illustrated in Fig. \ref{dsn}(c):
\begin{equation}
\begin{aligned}
&\frac{d D_{on}(x,y,t)}{d t} = \frac{F_{on}(x,y,t) - D_{on}(x,y,t)}{\tau_s},\\
&\frac{d D_{off}(x,y,t)}{d t} = \frac{F_{off}(x,y,t) - D_{off}(x,y,t)}{\tau_s}
\end{aligned}
\label{dynamic-lowpass}
\end{equation}
And similarly for computations of the VS system for the ON pathway:
\begin{equation}
\begin{aligned}
&E_{on}^{vs}(x,y,t) = \sum_{i=d}^{d \cdot N_{con}}D_{on}(x,y,t) \cdot F_{on}(x,y+i,t),\\
&I_{on}^{vs}(x,y,t) = \sum_{i=d}^{d \cdot N_{con}}D_{on}(x,y+i,t) \cdot F_{on}(x,y,t),\\
&ME^{vs}(x,y,t) = E_{on}^{vs}(x,y,t) - w_{i} \cdot I_{on}^{vs}(x,y,t)
\end{aligned}
\label{medulla-VS-EI}
\end{equation}
With similar ideas, in the lobula layer, the HS system for the OFF pathway is computed as:
\begin{equation}
\begin{aligned}
&E_{off}^{hs}(x,y,t) = \sum_{i=d}^{d \cdot N_{con}}D_{off}(x,y,t) \cdot F_{off}(x+i,y,t),\\
&I_{off}^{hs}(x,y,t) = \sum_{i=d}^{d \cdot N_{con}}D_{off}(x+i,y,t) \cdot F_{off}(x,y,t),\\
&LO^{hs}(x,y,t) = E_{off}^{hs}(x,y,t) - w_{i} \cdot I_{off}^{hs}(x,y,t)
\end{aligned}
\label{lobula-HS-EI}
\end{equation}
and calculations of the VS system for the OFF pathway are defined as:
\begin{equation}
\begin{aligned}
&E_{off}^{vs}(x,y,t) = \sum_{i=d}^{d \cdot N_{con}}D_{off}(x,y,t) \cdot F_{off}(x,y+i,t),\\
&I_{off}^{vs}(x,y,t) = \sum_{i=d}^{d \cdot N_{con}}D_{off}(x,y+i,t) \cdot F_{off}(x,y,t),\\
&LO^{vs}(x,y,t) = E_{off}^{vs}(x,y,t) - w_{i} \cdot I_{off}^{vs}(x,y,t)
\end{aligned}
\label{lobula-VS-EI}
\end{equation}

\subsection{The computational lobula plate layer}
In the final layer of the motion-detecting pathways in the fly's visual circuits, i.e. the lobula plate, there are four groups of LPTCs. Each group of neurons have specifically directional selectivity to one of the four cardinal orientations respectively, as illustrated in Fig. \ref{physiology}. We computationally model these LPTCs as the directionally selective systems (DSS in Fig. \ref{dsn}(a)), via linearly integrating relayed excitations from ON and OFF pathways forming the neural responses represented by membrane potentials in four cardinal directions:
\begin{equation}
\begin{aligned}
&LP_{on}^{hs}(t) = \sum_{1}^{C}\sum_{1}^{R}ME^{hs}(x,y,t),\ LP_{on}^{vs}(t) = \sum_{1}^{C}\sum_{1}^{R}ME^{vs}(x,y,t),\\
&LP_{off}^{hs}(t) = \sum_{1}^{C}\sum_{1}^{R}LO^{hs}(x,y,t),\ LP_{off}^{vs}(t) = \sum_{1}^{C}\sum_{1}^{R}LO^{vs}(x,y,t)
\end{aligned}
\label{LPTCs}
\end{equation}
where $C$ and $R$ indicate the numbers of columns and rows in the two-dimensional visual field. Importantly, with regard to the symmetrically nonlinear processing in the medulla and lobula layers, the global membrane potentials of four groups of LPTCs are rigorously tuned to be positive by preferred directions, i.e., rightward and downward motion, and negative by opposite or null directions - leftward and upward motion. To further reduce noise, we low-pass filter the membrane potential of each group of directionally specific LPTCs, the equation of which is similar to Eq. \ref{dynamic-lowpass} but with a fixed time parameter $\tau_{mp}$ in milliseconds.

Moreover, like other artificial neurons (e.g. \cite{LGMD1-Glayer,LGMD1-nonlinear,IROS-LGMDs,Colias-Hu}), we apply an activation function to realize spiking DSNs with an exponential relationship between the neural response and the firing frequency, which could be explained in terms of the sigmoid transformation function \cite{Biophysics_1998}. Let the membrane potential of each group of LPTCs be $x$, the activation function is expressed as:
\begin{equation}
f(x) = \operatorname{sgn}(x) \cdot ((1 + e^{-|x| \cdot (C \cdot R \cdot K_{sig})^{-1}})^{-1} - \Delta_{C})
\label{sigmoid}
\end{equation}
where $K_{sig}$ is a small coefficient. The output is normalized to $[0,0.5)$ for the positive input, and $(-0.5,0]$ for the negative input, by setting $\Delta_{C}$ to $0.5$: without such a coefficient, the output is within the range of $(-1,-0.5]$ for the negative input and $[0.5,1)$ for the positive input, which are not successive. Therefore, as depicted in Fig. \ref{dsn}(a), the sigmoid membrane potential of four groups of LPTCs ($\hat{LP}$) congregate at HS and VS systems separately, each output of which is within the range of $(-1,1)$:
\begin{equation}
HS(t) = \hat{LP}_{on}^{hs}(t) + \hat{LP}_{off}^{hs}(t),\quad VS(t) = \hat{LP}_{on}^{vs}(t) + \hat{LP}_{off}^{vs}(t)
\label{HS-VS}
\end{equation}

\subsection{Spiking DSNs}
In the proposed DSNN, we implement the DSNs as spiking neurons by exponentially mapping the sigmoid membrane potential of either HS/VS systems to different number of spikes at each discrete frame:
\begin{equation}
S^{spike}_{hs}(t) = \left \lfloor{e^{[K_{sp} \cdot (|HS(t)| - |T_{sp}|)]}}\right \rfloor,\ S^{spike}_{vs}(t) = \left \lfloor{e^{[K_{sp} \cdot (|VS(t)| - |T_{sp}|)]}}\right \rfloor
\label{spiking}
\end{equation}
where $\left \lfloor{x}\right \rfloor$ indicates a `floor' function to obtain the largest integer less than or equal the input. $K_{sp}$ denotes a coefficient, which can directly affect the firing rate, i.e., increasing it will lead to higher firing rate. $T_{sp}$ designates the spiking threshold, which is positive to preferred-directional output yet negative to null-directional output. Through such a spiking mechanism, more than one spikes could be generated at each frame.

\subsection{The selection of DSNN parameters}
\begin{table}[t]
	\caption{The predefined parameters of DSNN}
	\centering
	\begin{tabular}{ll|ll|ll}
		\toprule
		Name      &Value     &Name      &Value       &Name      &Value\\
		\hline
		$C$,$\ R$ &adaptable &$K_{sp}$  &$1 \sim 3$  &$\tau_s$  &$10 \sim 200$ms\\
		$w_i$     &$0.9$     &$N_{con}$ &$4 \sim 8$  &$d$       &$1 \sim 4$\\
		$\sigma_e,\ \sigma_i$&$d,\ 2 \cdot d$&$N_p$&$0 \sim 6$&$\sigma_l$&$0.1$\\
		$\tau_{fast}$&$1$ms  &$\tau_{slow}$&$100$ms  &$\tau_{mp}$ &$10$ms\\
		$K_{sig}$ &$0.01$    &$\Delta_C$&$0.5$       &$T_{sp}$  &$\pm0.16 \sim \pm0.2$\\
		\bottomrule
	\end{tabular}
	\label{tableParams}
\end{table}
All model parameters of the proposed DSNN are decided empirically with considerations of the functionality of biological DSNs for translational motion detection in dynamic and complex scenes, as well as the implementation as an embedded vision system in a micro robot. There are currently no parameters training methods involved in this framework. Table \ref{tableParams} presents the predefined parameters of DSNN. The adaptable parameters $C$ and $R$ are decided by the resolution of input images. In the DoGs algorithm, we shape the Gaussians by balancing the standard deviations on two dimensions, and make the outer negative Gaussian twice the size of the inner positive Gaussian for forming selectivity to ON-edges and OFF-edges. It also appears that the widths of Gaussians depend on the spacing between the nearest neighboring ON/OFF motion detectors, i.e., it is essentially determining the spatial frequency resolution in the band-pass filtering of the computational lamina layer. In addition, as mentioned above, a critically important feature of this neural network is the building of ensembles of motion detectors in ON and OFF pathways. Increasing the number of connected cells ($N_{con}$) for each local unit in the dual-pathways could further improve the speed response to moving dark/light features, at the cost though of more computational consumption.

In the next section, we will represent the systematic experiments, the results of which clearly demonstrate how the outputs of DSNN, i.e., membrane potential and spiking frequency of DSNs, reflect the direction and magnitude information of foreground translational motion against visually cluttered backgrounds.

\section{Experimental evaluation}
\label{expriments}
In this section, we present systematic experiments along with analysis and discussion. The main objectives were firstly to assess the fundamental functionality and effectiveness of the proposed DSNN on translational motion perception; second, we systematically investigated its internal properties, and compared with an EMDs-based model \cite{Iida_2000} as well as a preliminary modeling work of this research \cite{DSN-IJCNN}. Importantly, we also tested its feasibility and robustness as an embedded vision system in an autonomous micro-robot. All the experiments can be categorized into two types of tests: off-line and on-line tests. In the off-line tests, the visual stimuli comprise computer-simulated and real physical scenarios. In the on-line tests, the embedded DSNN was systematically tested.

\subsection{Experimental setting}
We first introduce the software and hardware set-ups. In the off-line tests, the frameworks of DSNN and two comparative models were all set up in Visual Studio 2015 (Microsoft Corporation). Data analysis and representations were accomplished in Matlab 2015 (The MathWorks, Inc. Natick, USA). The resolutions of synthetic visual streams are $320 \times 180$ and $540 \times 180$ for translational movements embedded in clean and natural backgrounds respectively. The resolution of real-world visual stimuli is $320 \times 180$. All the video images are converted to the grayscale format at the sampling frequency of $30$Hz for the processing of neural networks.
\begin{figure}[t]
	\centering
	\includegraphics[width=0.7\linewidth]{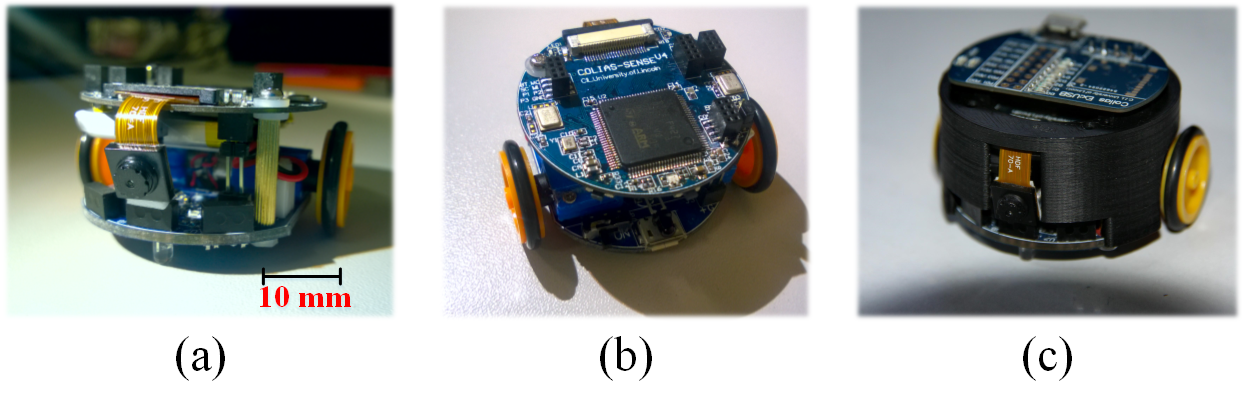}
	\caption{The Colias robot prototype: (a) -- (b) It mainly consists of two boards - \textbf{the upper board is equipped with a RGB camera and can execute vision based models}; the bottom board actuates motion control with two wheels and a battery. (c) We can decorate it with a 3D-printed black shell in robot experiments as translation `stimuli'.}
	\label{colias}
\end{figure}

In the on-line tests, the mobile robot platform is a low-cost micro robot named `Colias' \cite{Colias-Hu,Colias-introduction} with a small color camera, which is the only sensor used in this research. It has been developed for swarm robotic applications \cite{CosPhi,Colias-investigation}, as well as biologically monocular-vision based systems research \cite{Colias-Cheng,Colias-Hu,LGMD2-BMVC,LGMD2-Fu,IROS-LGMDs}. As illustrated in Fig. \ref{colias}, the robot has a small footprint of $4\ $cm in diameter and $3\ $cm in height, with two main boards or modules. The bottom board is the motion actuator with two DC motors driven differentially that provide the robot platform a maximum speed of approximately $35\ $cm/s. In addition, a $3.7\ $V, $320\ $mAh lithium battery supports the autonomy for $1\sim2$ hours.

The upper board executes vision-based models. Its processor for running neural networks, including image processing, is the ARM-Cortex M4 based MCU STM32F427 running at $180\ $MHz, with $256\ $Kbyte SRAM, 2Mbyte in-chip Flash. As depicted in Fig. \ref{colias}, the assembled camera utilized in this study is an OV7670 from Omni-vision, with approximately $70^{\circ}$ field of view. In comparison with the off-line tests, the acquired image was set to the resolution of $99 \times 72$ in YUV422 format at $30$ fps. In addition, in this research, we applied a bluetooth device connected with the visual module, for the purpose of obtaining real-time model outputs remotely, including the membrane potential and spikes from the robot.

\subsection{Synthetic stimuli tests}
First of all, our experiments started by testing the DSNN using computer-simulated visual stimuli consisting of the movements of darker and lighter objects embedded in clean and visually cluttered backgrounds respectively. All the synthetic stimuli can be categorized into the following types: the depth-movements including approaching and receding of objects, translations in both horizontal and vertical directions.

\paragraph{Visual stimuli embedded in a clean background}
\begin{figure}[!t]
	\begin{minipage}[t]{0.48\linewidth}
		\centering
		\centerline{\includegraphics[width=1in]{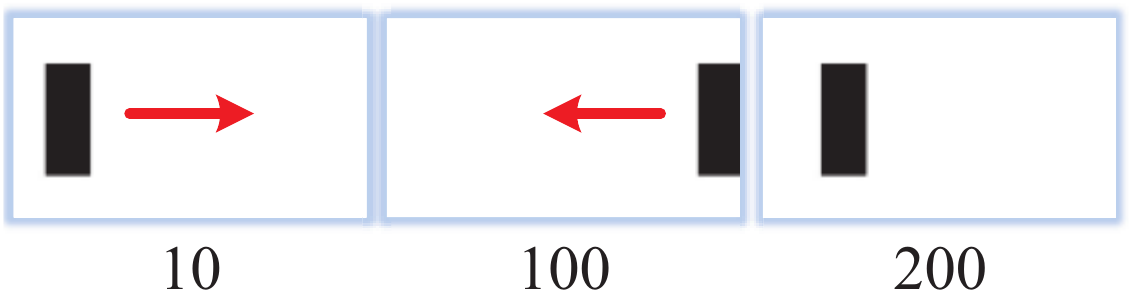}}
	\end{minipage}
	\hfill
	\begin{minipage}[t]{0.48\linewidth}
		\centering
		\centerline{\includegraphics[width=1in]{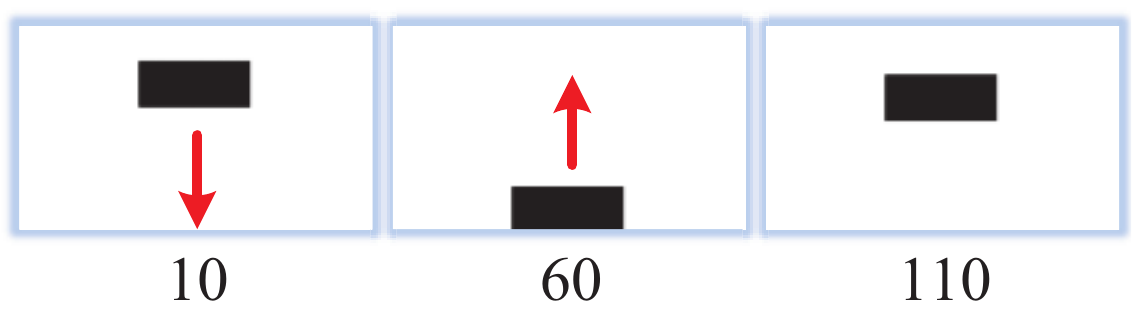}}
	\end{minipage}
	\vfill
	\begin{minipage}[t]{0.48\textwidth}
		\centering
		\centerline{\includegraphics[width=2in]{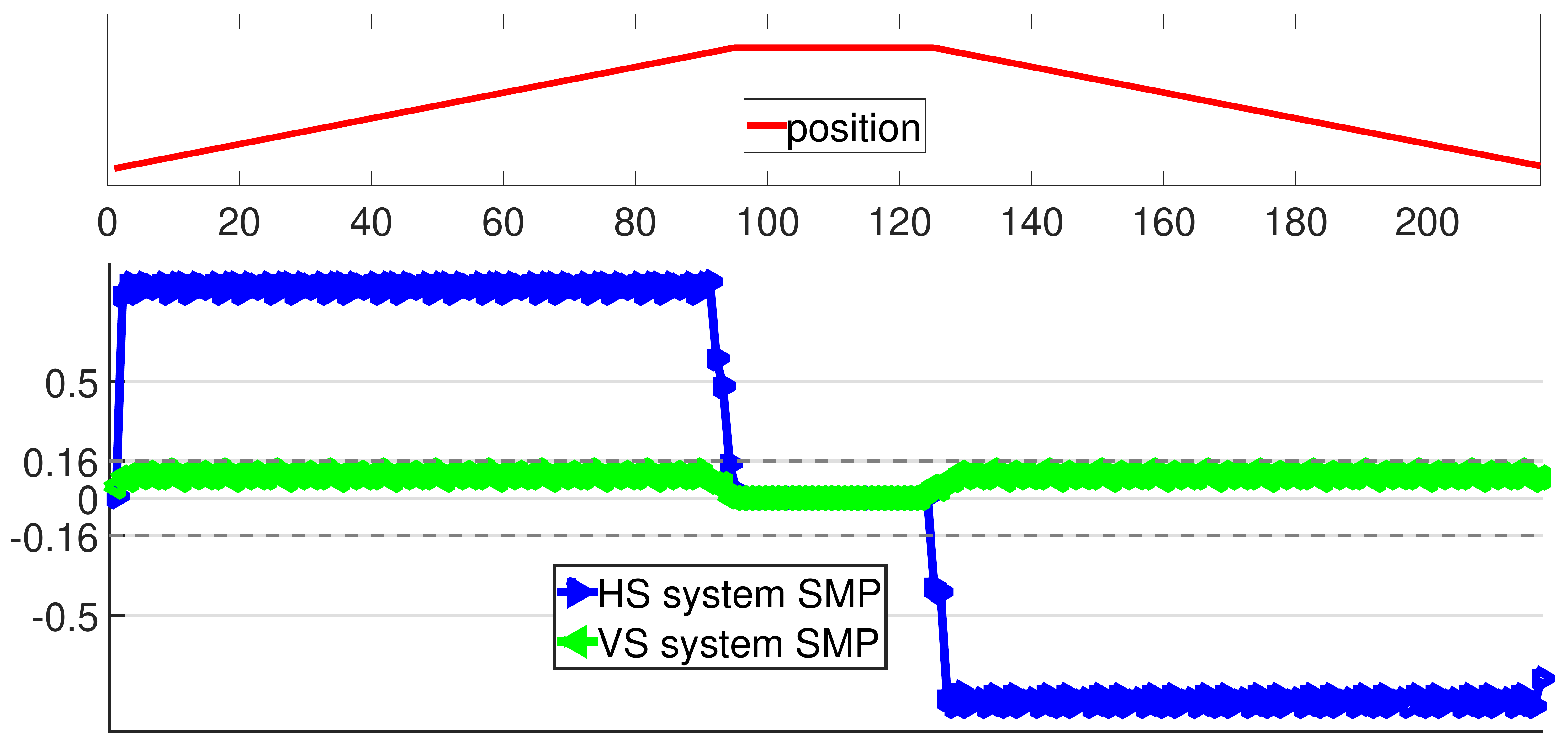}}
		\centerline{\scriptsize(a) a darker object \textbf{translating horizontally}}
	\end{minipage}
	\hfill
	\begin{minipage}[t]{0.48\textwidth}
		\centering
		\centerline{\includegraphics[width=2in]{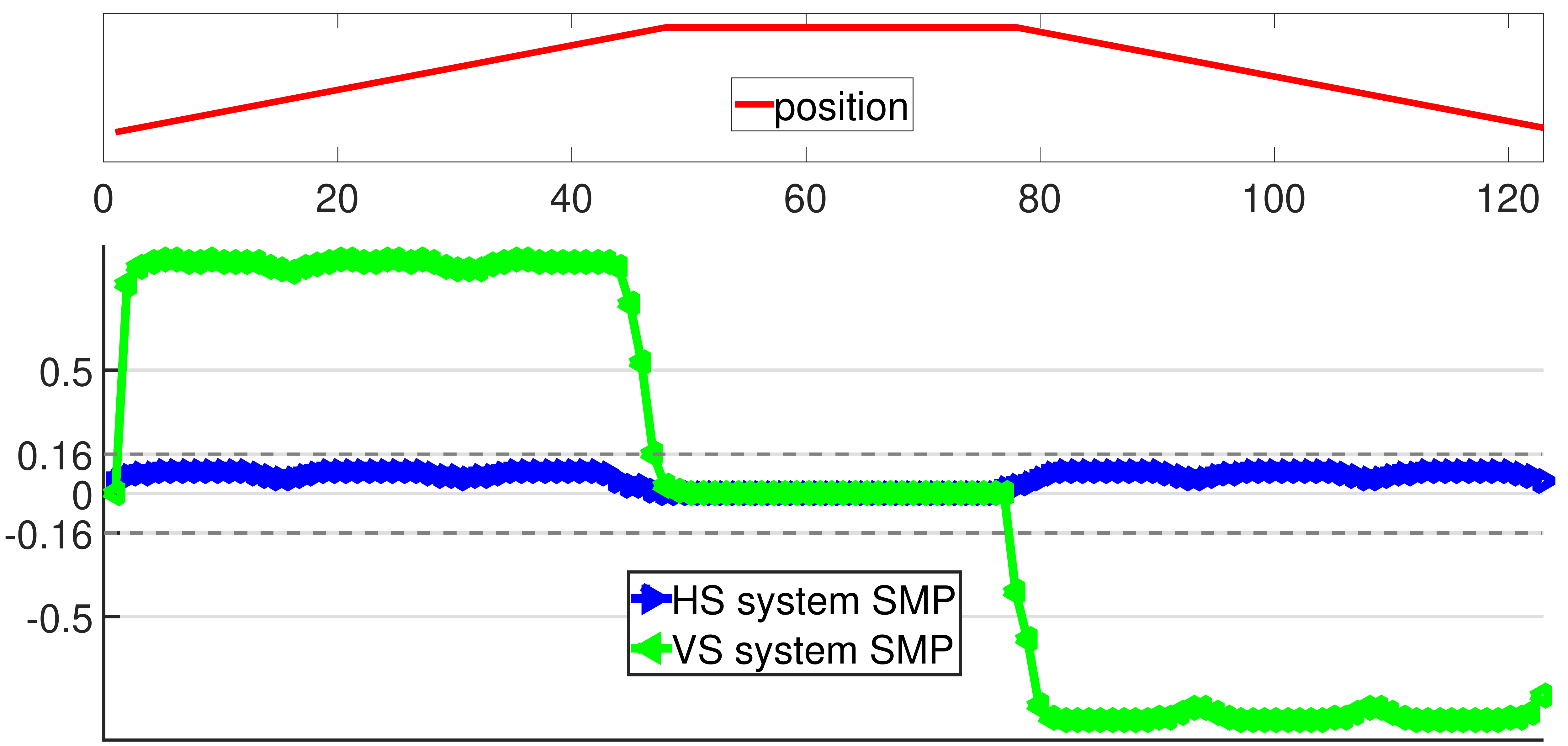}}
		\centerline{\scriptsize(b) a darker object \textbf{translating vertically}}
	\end{minipage}
	\vfill
	\vspace{0.1in}
	\begin{minipage}[t]{0.48\linewidth}
		\centering
		\centerline{\includegraphics[width=1in]{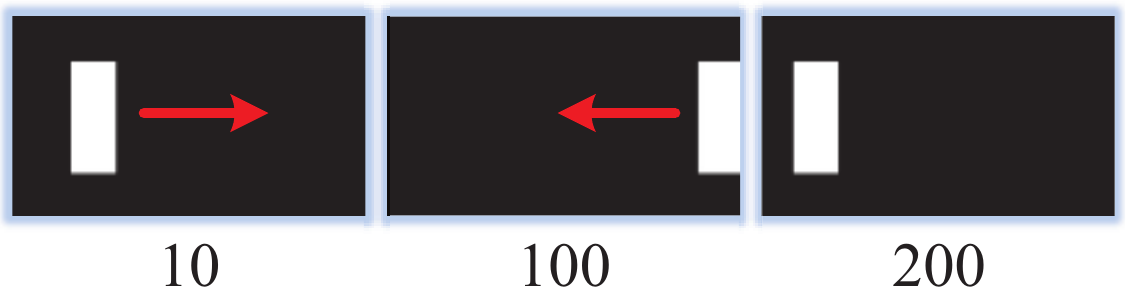}}
	\end{minipage}
	\hfill
	\begin{minipage}[t]{0.48\linewidth}
		\centering
		\centerline{\includegraphics[width=1in]{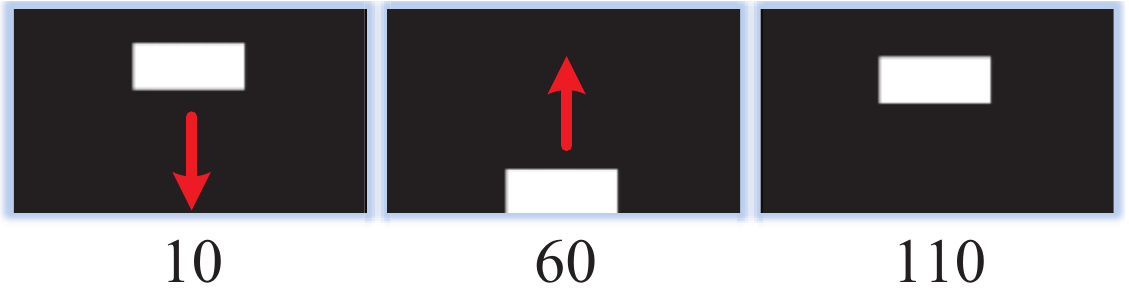}}
	\end{minipage}
	\vfill
	\begin{minipage}[t]{0.48\textwidth}
		\centering
		\centerline{\includegraphics[width=2in]{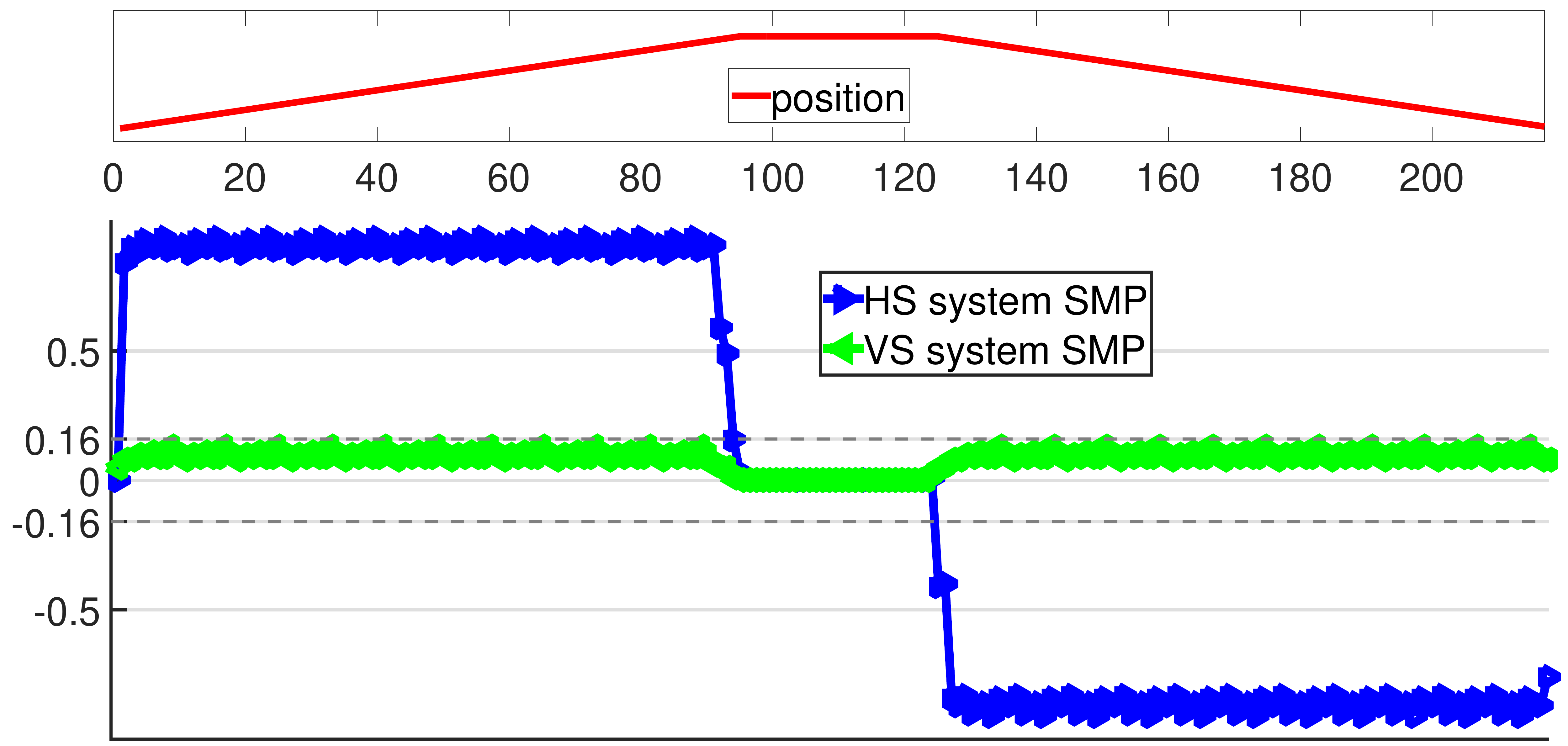}}
		\centerline{\scriptsize(c) a lighter object \textbf{translating horizontally}}
	\end{minipage}
	\hfill
	\begin{minipage}[t]{0.48\textwidth}
		\centering
		\centerline{\includegraphics[width=2in]{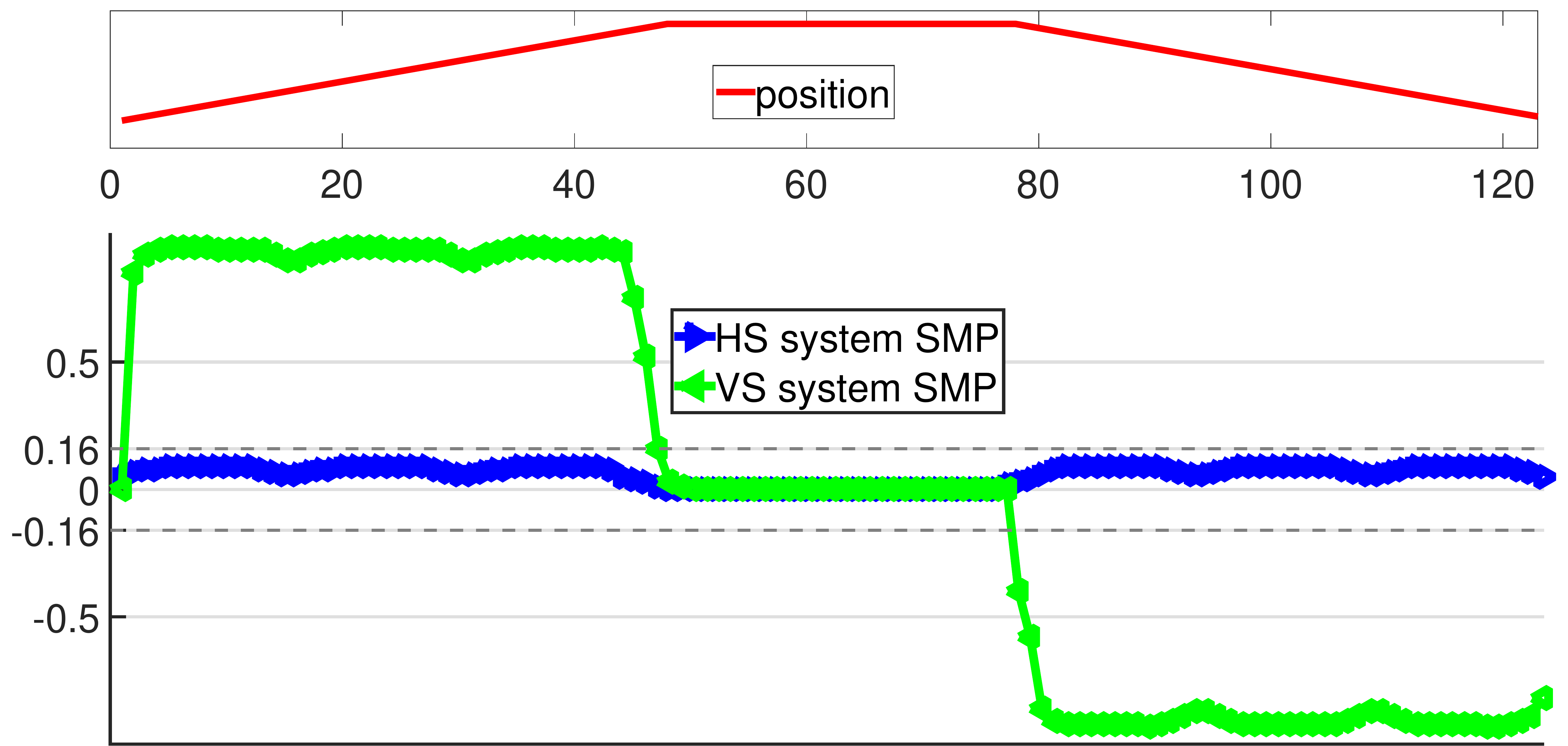}}
		\centerline{\scriptsize(d) a lighter object \textbf{translating vertically}}
	\end{minipage}
	\vfill
	\vspace{0.1in}
	\begin{minipage}[t]{0.48\linewidth}
		\centering
		\centerline{\includegraphics[width=1in]{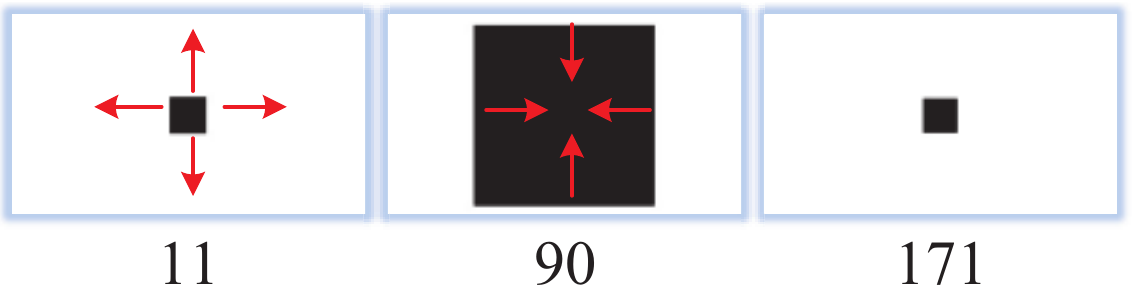}}
	\end{minipage}
	\hfill
	\begin{minipage}[t]{0.48\linewidth}
		\centering
		\centerline{\includegraphics[width=1in]{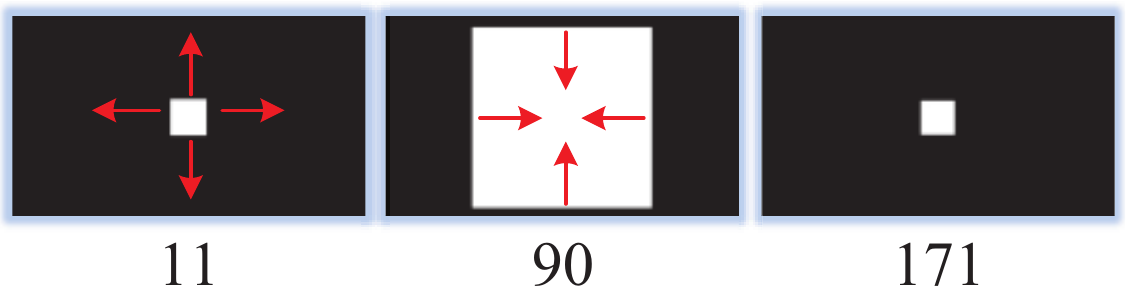}}
	\end{minipage}
	\vfill
	\begin{minipage}[t]{0.48\textwidth}
		\centering
		\centerline{\includegraphics[width=2in]{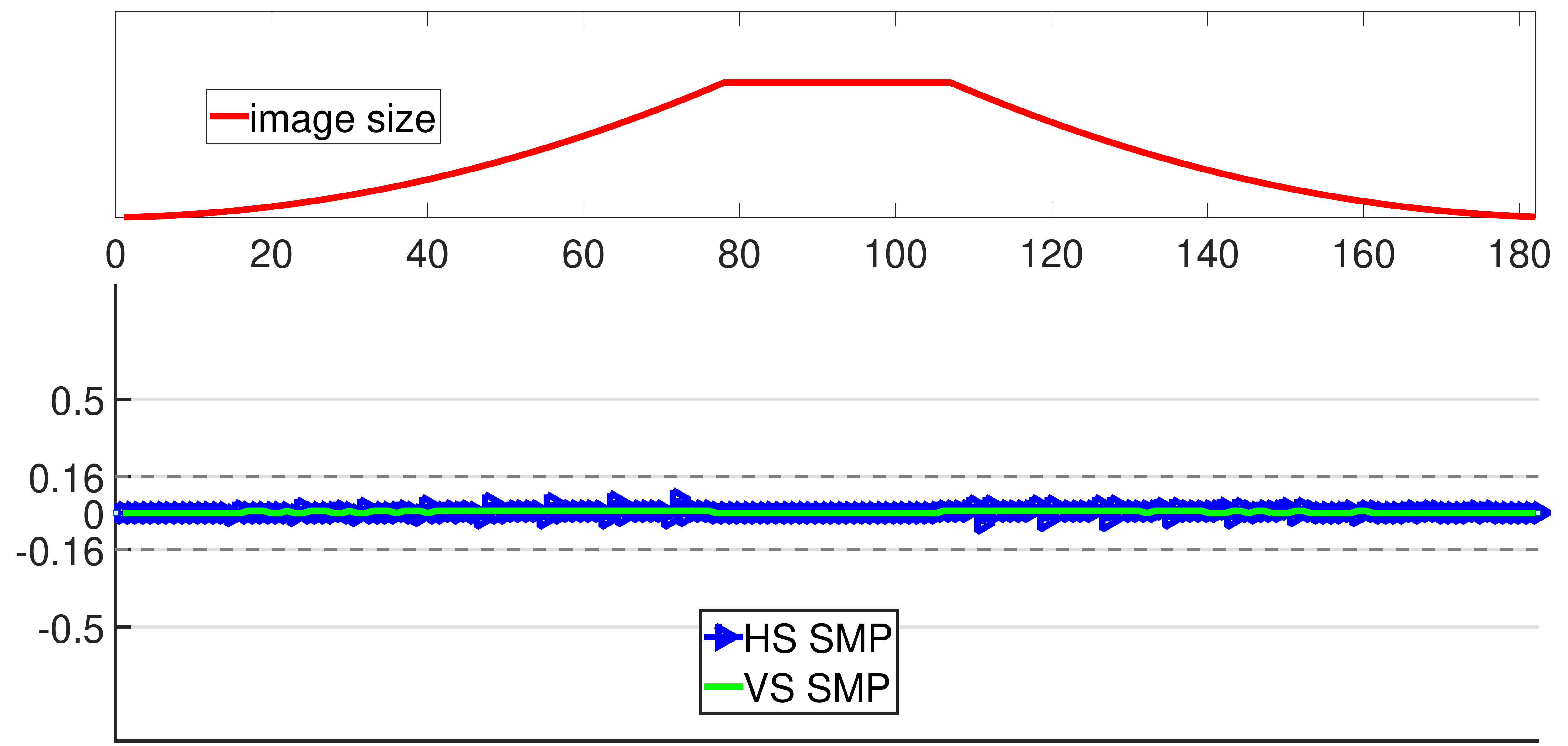}}
		\centerline{\scriptsize(e) a darker object \textbf{looming -- receding}}
	\end{minipage}
	\hfill
	\begin{minipage}[t]{0.48\textwidth}
		\centering
		\centerline{\includegraphics[width=2in]{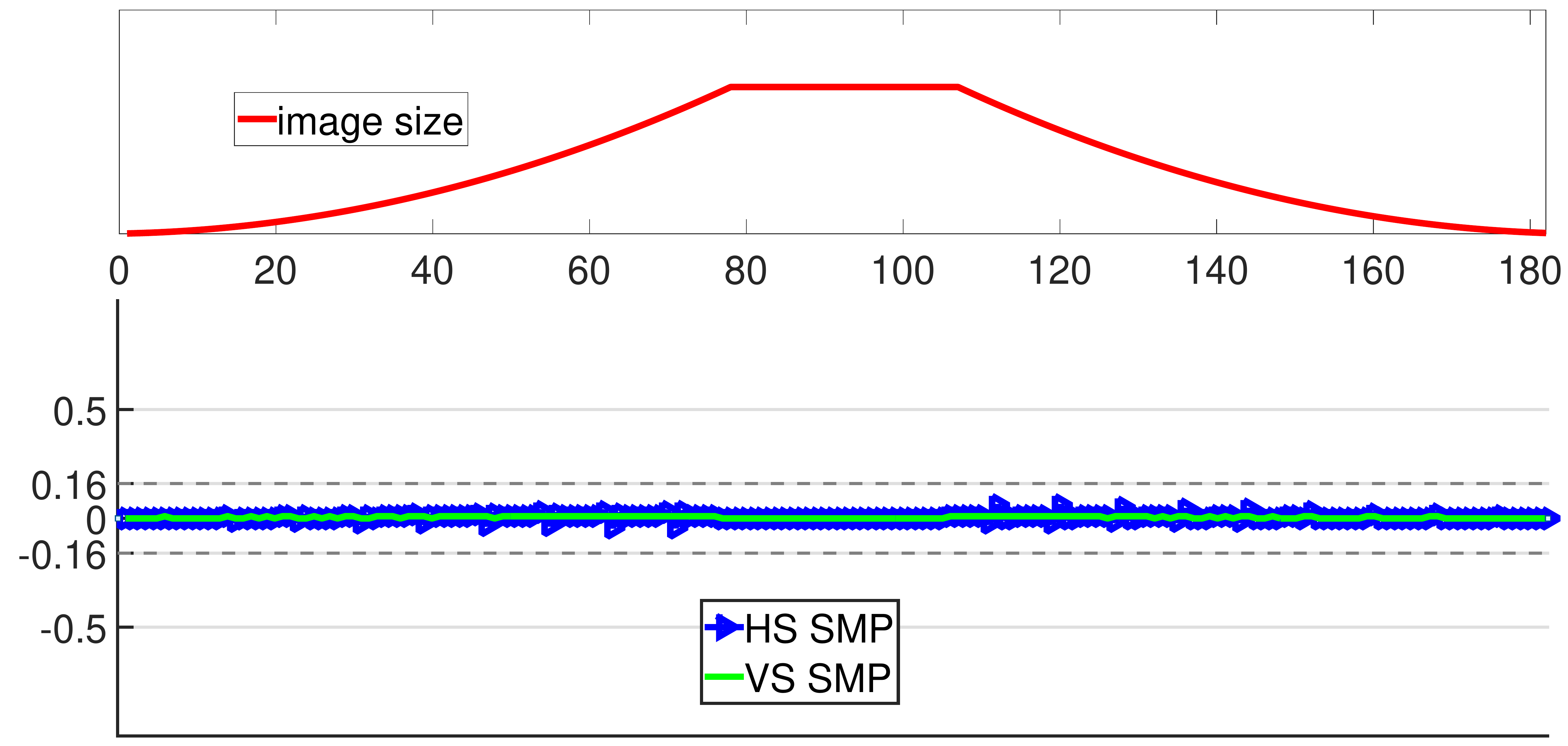}}
		\centerline{\scriptsize(f) a lighter object \textbf{looming -- receding}}
	\end{minipage}
	\caption{The DSNN is challenged by synthetic visual stimuli of dark and light objects approaching, receding and translating against light and dark backgrounds respectively. The example views of input frames are shown at top of each result. The changes of object position or size in the visual field are depicted below the snapshots. The sigmoid membrane potentials (SMP) of HS and VS systems of DSNN are represented separately. The horizontal dashed lines indicate the predefined spiking thresholds ($\pm0.16$). X and Y axes denote the time course in frames and SMP respectively. \textbf{The DSNN responds to translational motion in preferred and non-preferred directions with positive and negative membrane potential, while it is rigorously inhibited during objects approaching and receding}.}
	\label{simu-clean}
\end{figure}
As described above, the first and basic objective of the experiments is to show the basic functionality of the proposed DSNN. First, challenged by translational motion in four cardinal directions (Fig. \ref{simu-clean}(a) -- (d)), the proposed DSNN represents successively positive sigmoid membrane potential (SMP) when challenged by motion in preferred directions (rightward and downward for HS and VS systems respectively), while negative membrane potential against motion in non-preferred directions - leftward for the HS system and upward for the VS system. The motion direction is well tuned by the symmetric structure of Reichardt detectors within the ensembles of ON-ON and OFF-OFF motion detectors in the computational medulla and lobula layers. The results also well match the physiological research outcomes of the fly's visual pathways \cite{Joesch_2013}.

Second, we tested the DSNN with approaching and receding movements of either dark (Fig. \ref{simu-clean}(e)) or light (Fig. \ref{simu-clean}(f)) objects embedded in light and dark backgrounds. The results illustrate that DSNN is rigorously inhibited during each whole course of movements in depth. Interestingly, compared with the looming detectors like LGMD1 \cite{LGMD1-Glayer} and LGMD2 \cite{IROS-LGMDs,LGMD2-BMVC} based neural networks, which rigorously respond to approaching over translating visual stimuli, the DSNN represents totally reverse response. We will further investigate these fundamental characteristics of DSNN in the robot experiments.

\begin{figure}[t]
	\begin{minipage}[t]{0.32\linewidth}
		\centering
		\centerline{\includegraphics[width=1.6in]{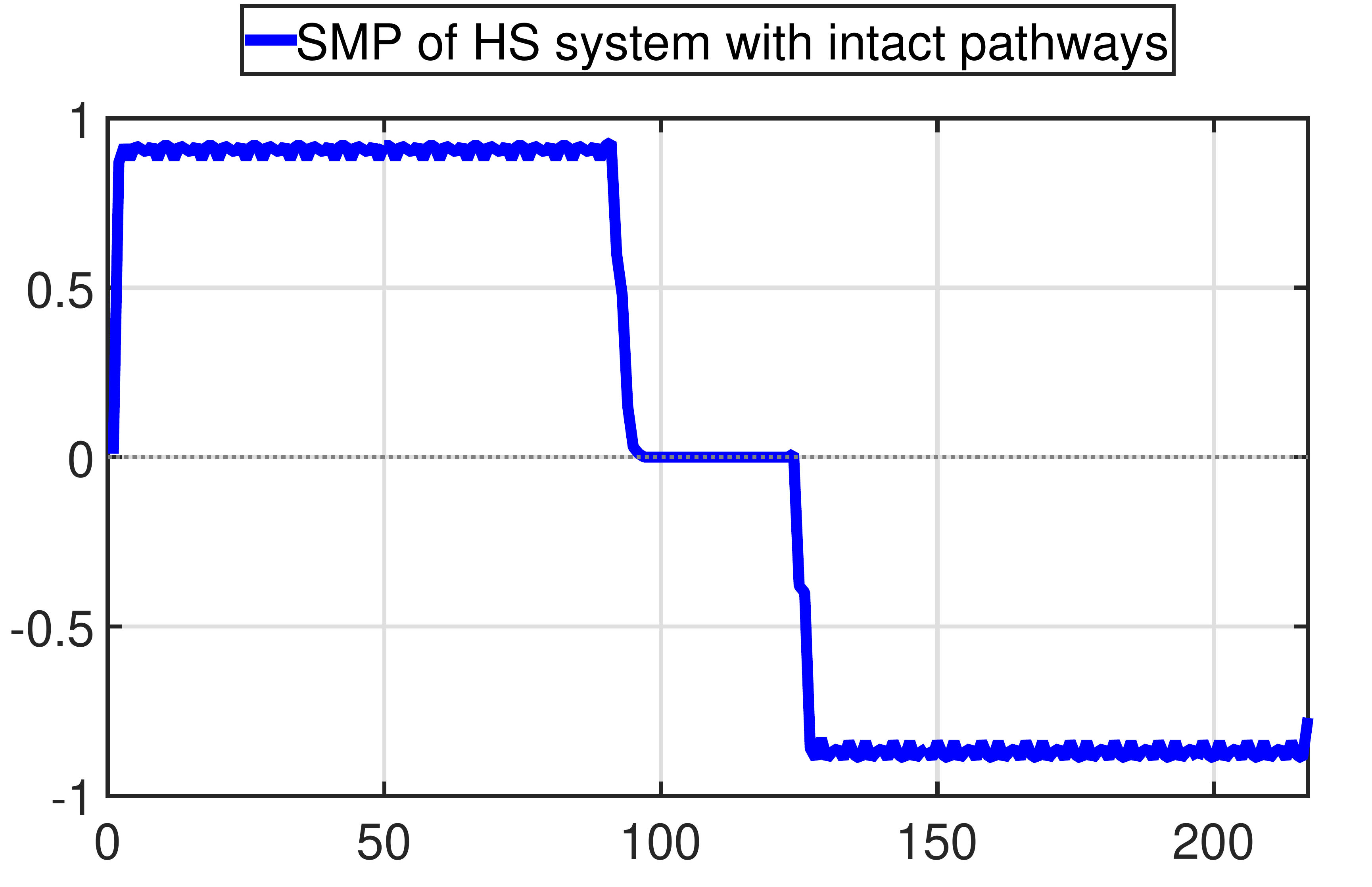}}
		\centerline{\scriptsize(a)}
	\end{minipage}
	\hfill
	\begin{minipage}[t]{0.32\linewidth}
		\centering
		\centerline{\includegraphics[width=1.6in]{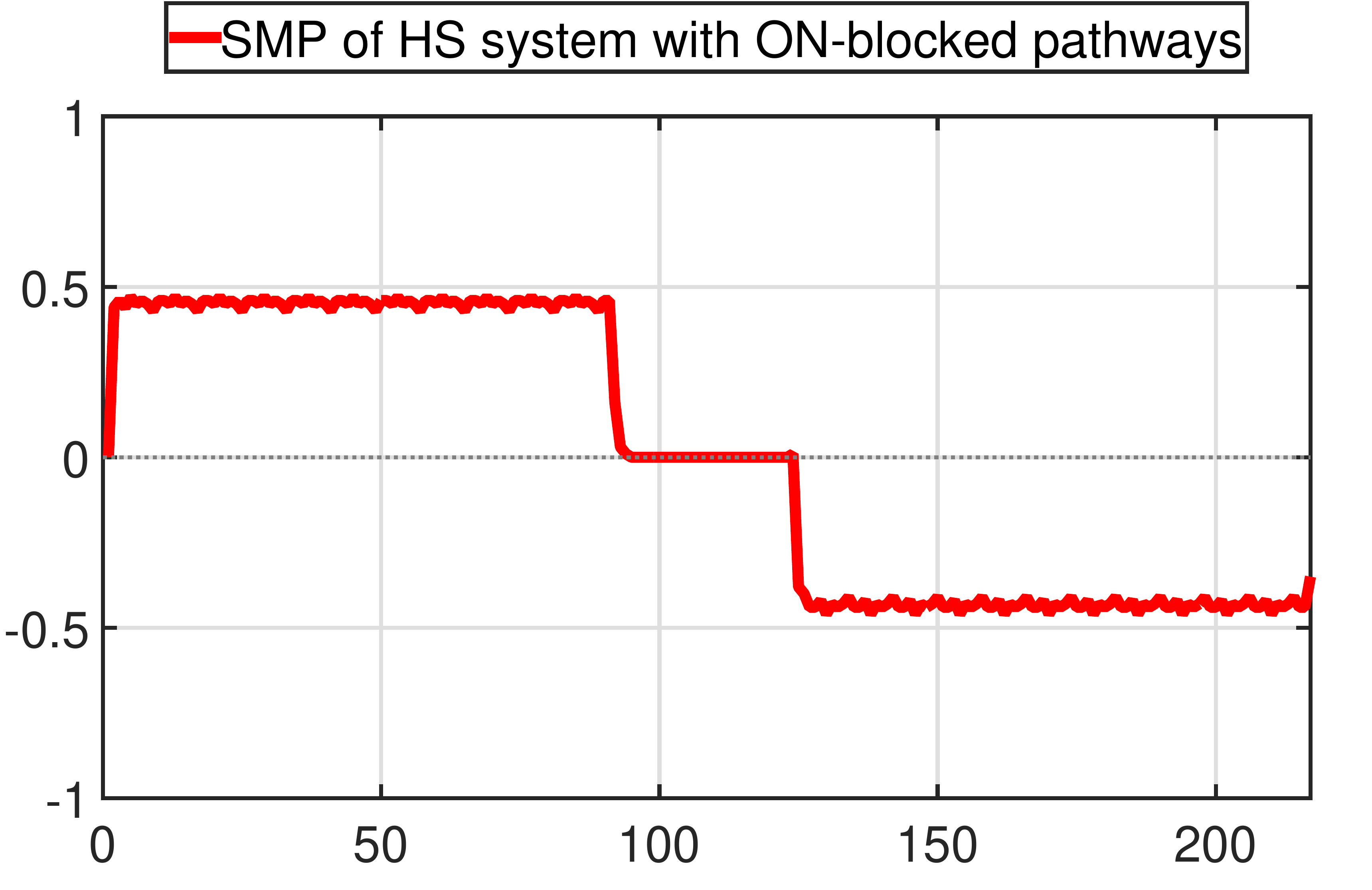}}
		\centerline{\scriptsize(b)}
	\end{minipage}
	\hfill
	\begin{minipage}[t]{0.32\linewidth}
		\centering
		\centerline{\includegraphics[width=1.6in]{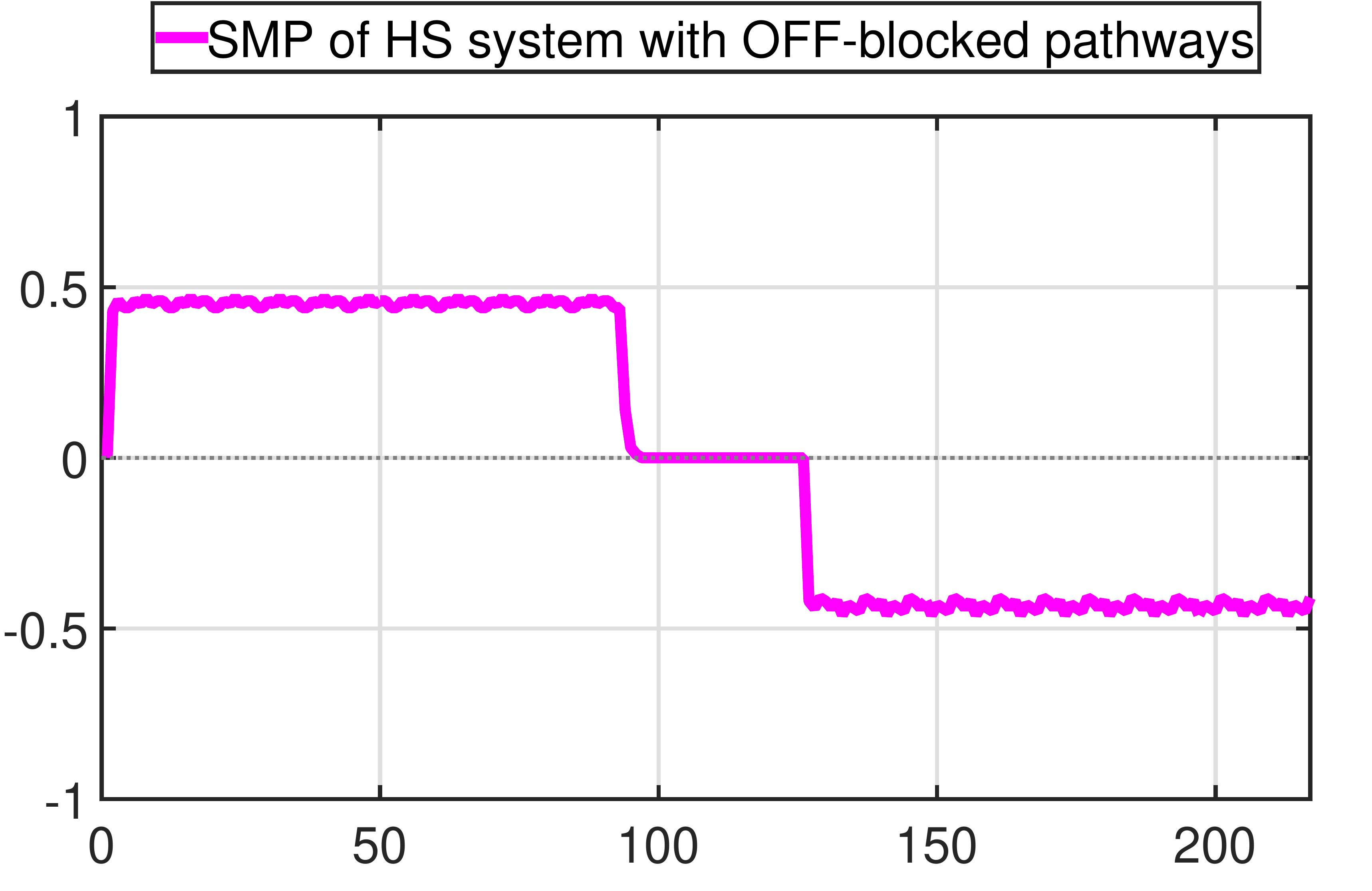}}
		\centerline{\scriptsize(c)}
	\end{minipage}
	\caption{The neural response (SMP) of DSNN with intact ON and OFF pathways (a), ON-blocked pathways (b), and OFF-blocked pathways (c), challenged by a dark object translating \textbf{horizontally} against a light background corresponding to Fig. \ref{simu-clean}(a).}
	\label{simu-clean-dth}
\end{figure}

\begin{figure}[t]
	\begin{minipage}[t]{0.32\linewidth}
		\centering
		\centerline{\includegraphics[width=1.6in]{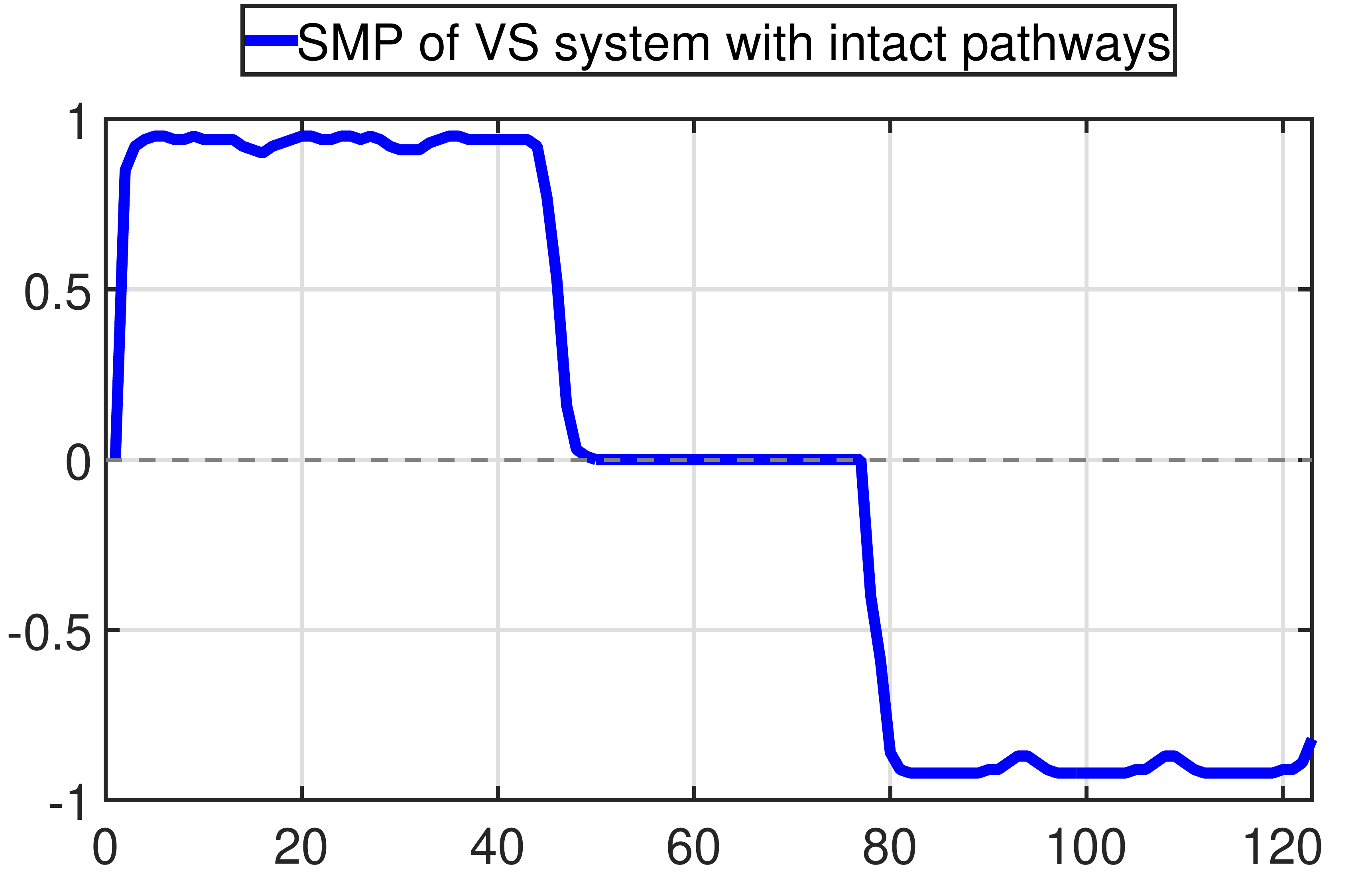}}
		\centerline{\scriptsize(a)}
	\end{minipage}
	\hfill
	\begin{minipage}[t]{0.32\linewidth}
		\centering
		\centerline{\includegraphics[width=1.6in]{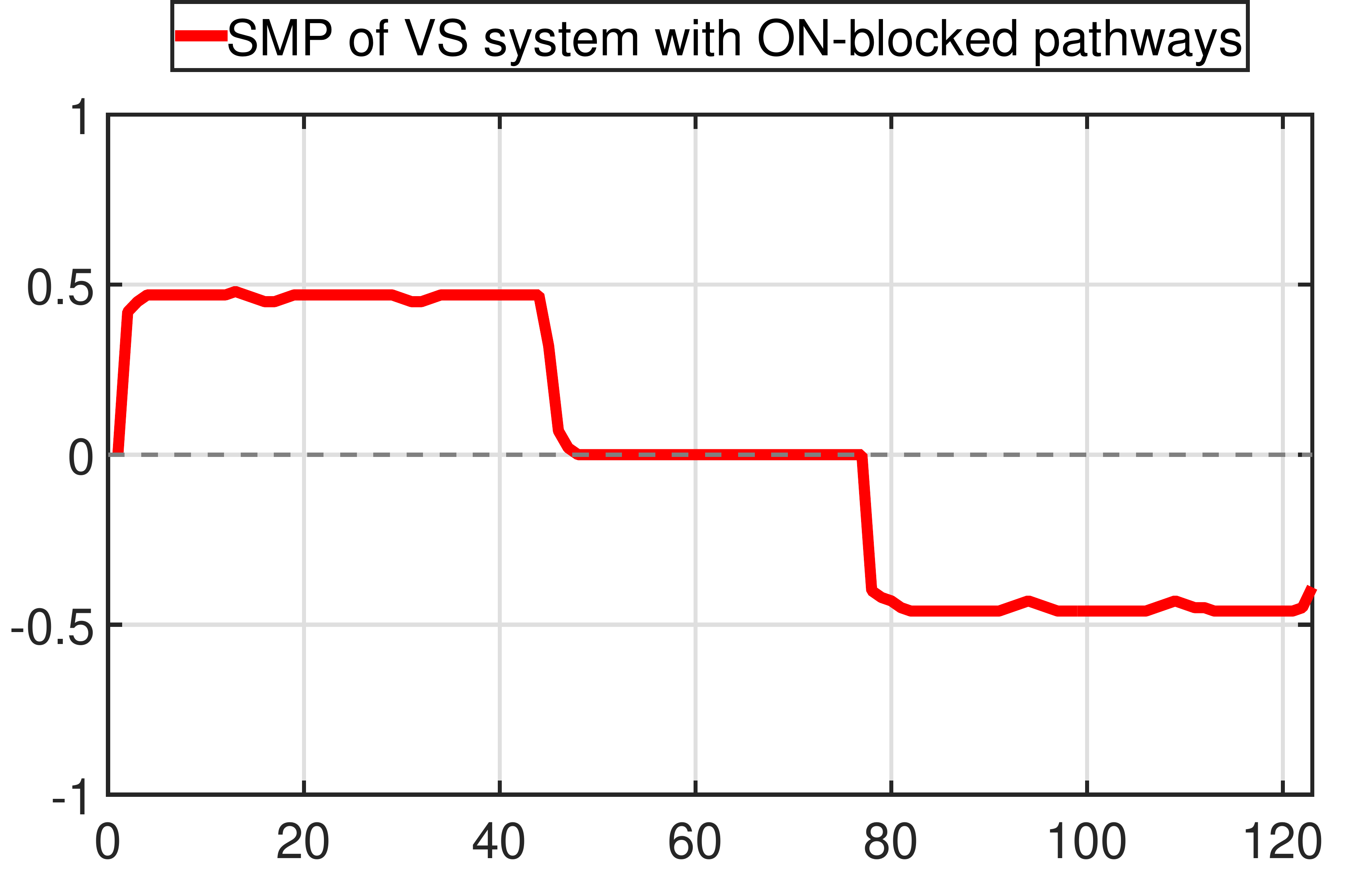}}
		\centerline{\scriptsize(b)}
	\end{minipage}
	\hfill
	\begin{minipage}[t]{0.32\linewidth}
		\centering
		\centerline{\includegraphics[width=1.6in]{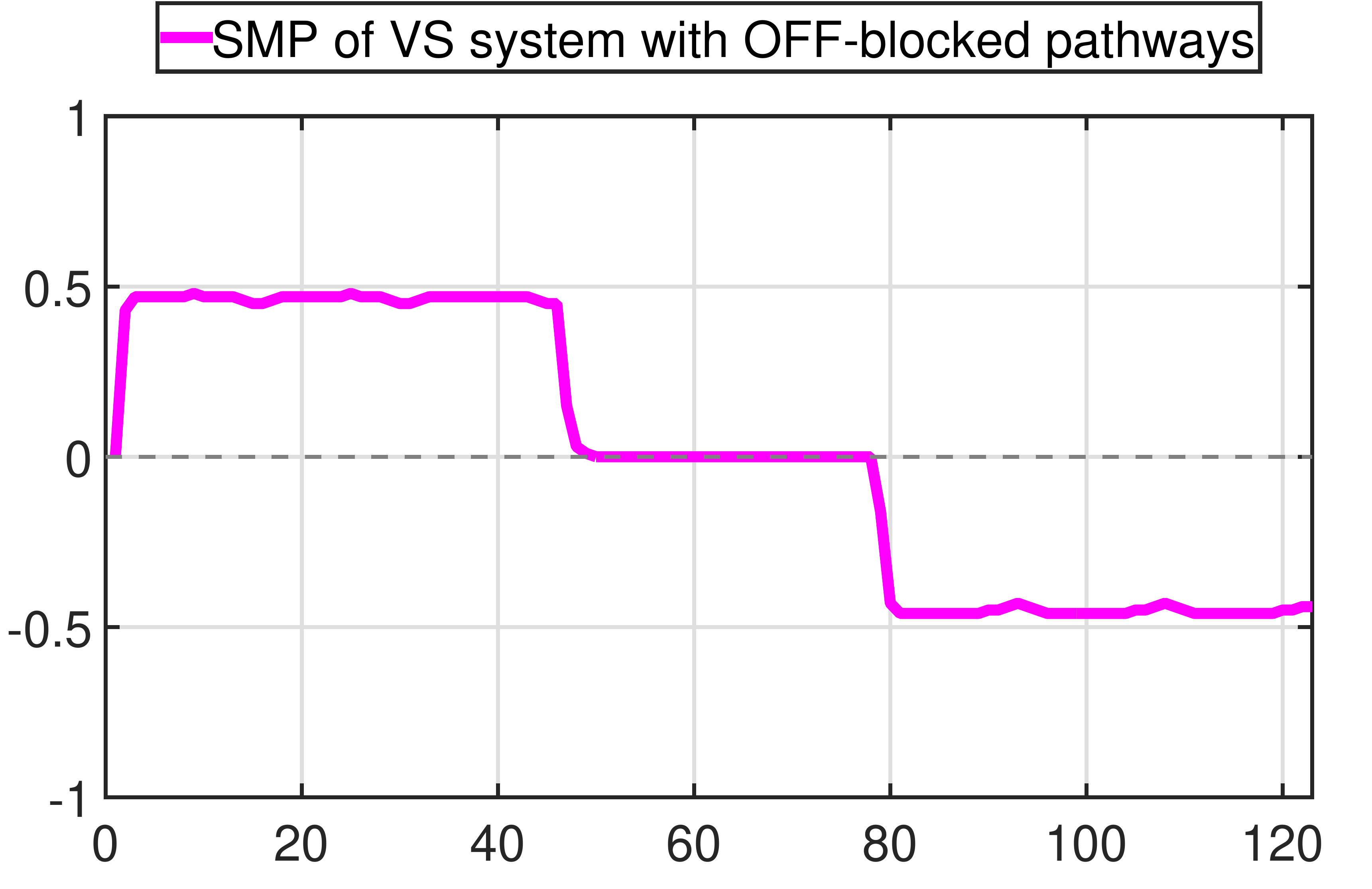}}
		\centerline{\scriptsize(c)}
	\end{minipage}
	\caption{The neural response (SMP) of DSNN with intact ON and OFF pathways (a), ON-blocked pathways (b), and OFF-blocked pathways (c), challenged by a dark object translating \textbf{vertically} against a light background corresponding to Fig. \ref{simu-clean}(b).}
	\label{simu-clean-dtv}
\end{figure}

Moreover, motivated by the systematically physiological experiments demonstrated in \cite{Joesch_2013}, we examined the specialized functionality of ON and OFF pathways in the proposed computational model. With similar ideas, we compared the membrane potential generated by intact ON and OFF pathways with ON-blocked and OFF-blocked systems. Taken the translations of a dark object as an example, the results illustrated in Fig. \ref{simu-clean-dth} -- \ref{simu-clean-dtv} demonstrate that blocking either ON/OFF pathways abolishes the corresponded functions of ON/OFF RTCs respectively, so that cutting down the membrane potential of either HS/VS systems to its half-level produced by the intact pathways. It thus turns out that ON-blocked or OFF-blocked model only possesses the ability of sensing light-off (offset) or light-on (onset) response. To be more specific, for a dark translating object embedded in a light background, the moving leading edge generates an offset response by the light-to-dark luminance change so that rigorously activating the OFF RTCs in the computational lamina layer, whilst the trailing edge leads to an onset response by the dark-to-light luminance change activating the ON RTCs. The opposite happens for a light translating object embedded in a dark background, where the leading and trailing edges rigorously activate ON and OFF RTCs respectively. The results verify that the functionality of separated ON/OFF pathways of DSNN well matches the underlying fly's physiology \cite{Joesch_2013}. 

\paragraph{Visual stimuli embedded in a shifting cluttered background}
After demonstrating the basic functions of the proposed framework, we designed synthetic stimuli in a natural background with global shifting, to further inspect its robustness in translational motion perception in dynamic and complex scenes, and more importantly to compare with two related models - an EMDs-based model \cite{Iida_2000} and a preliminary DSNs model \cite{DSN-IJCNN} from our previous research.
\begin{figure}[!t]
	\begin{minipage}[t]{0.5\linewidth}
		\centering
		\centerline{\includegraphics[width=1.5in]{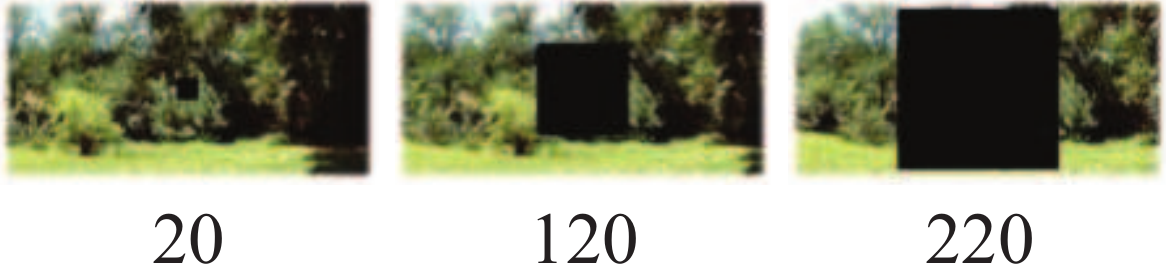}}
	\end{minipage}
	\hfill
	\begin{minipage}[t]{0.5\linewidth}
		\centering
		\centerline{\includegraphics[width=1.5in]{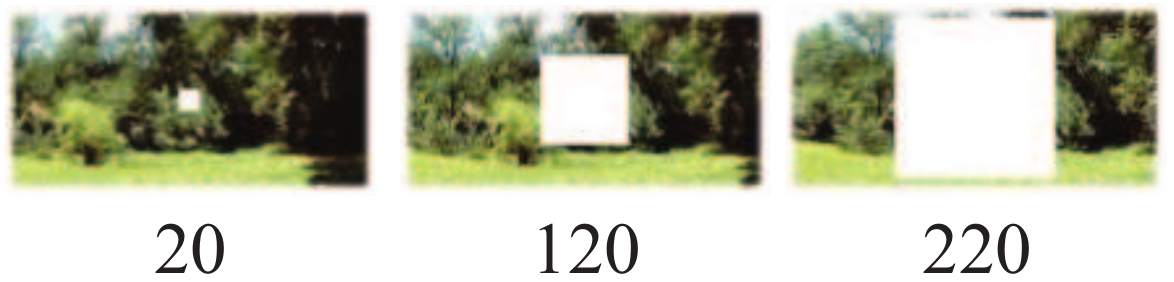}}
	\end{minipage}
	\vfill
	\begin{minipage}[t]{0.5\linewidth}
		\centering
		\centerline{\includegraphics[width=2in]{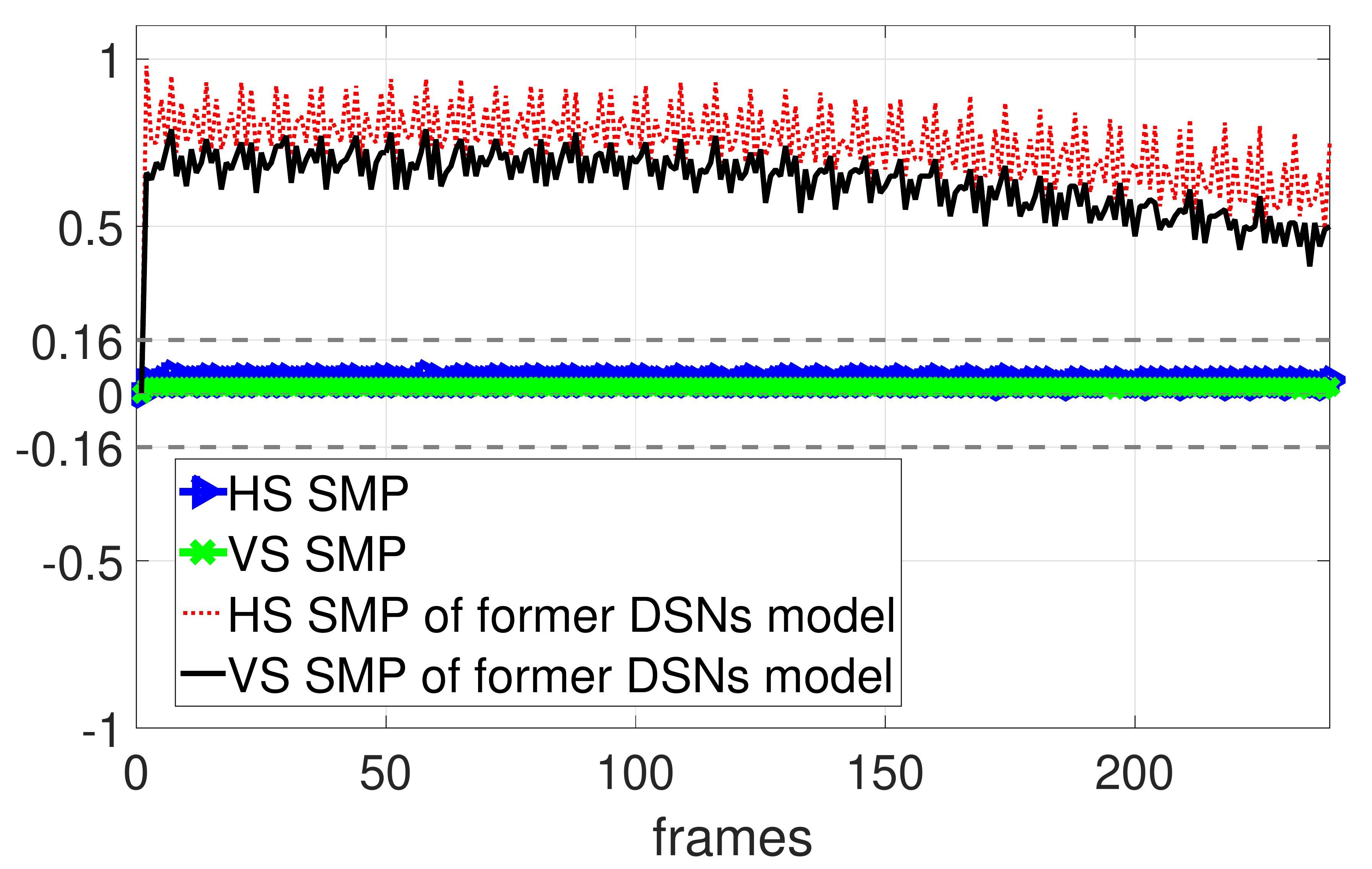}}
		\centerline{\scriptsize(a) a dark object \textbf{approaching}}
	\end{minipage}
	\hfill
	\begin{minipage}[t]{0.5\linewidth}
		\centering
		\centerline{\includegraphics[width=2in]{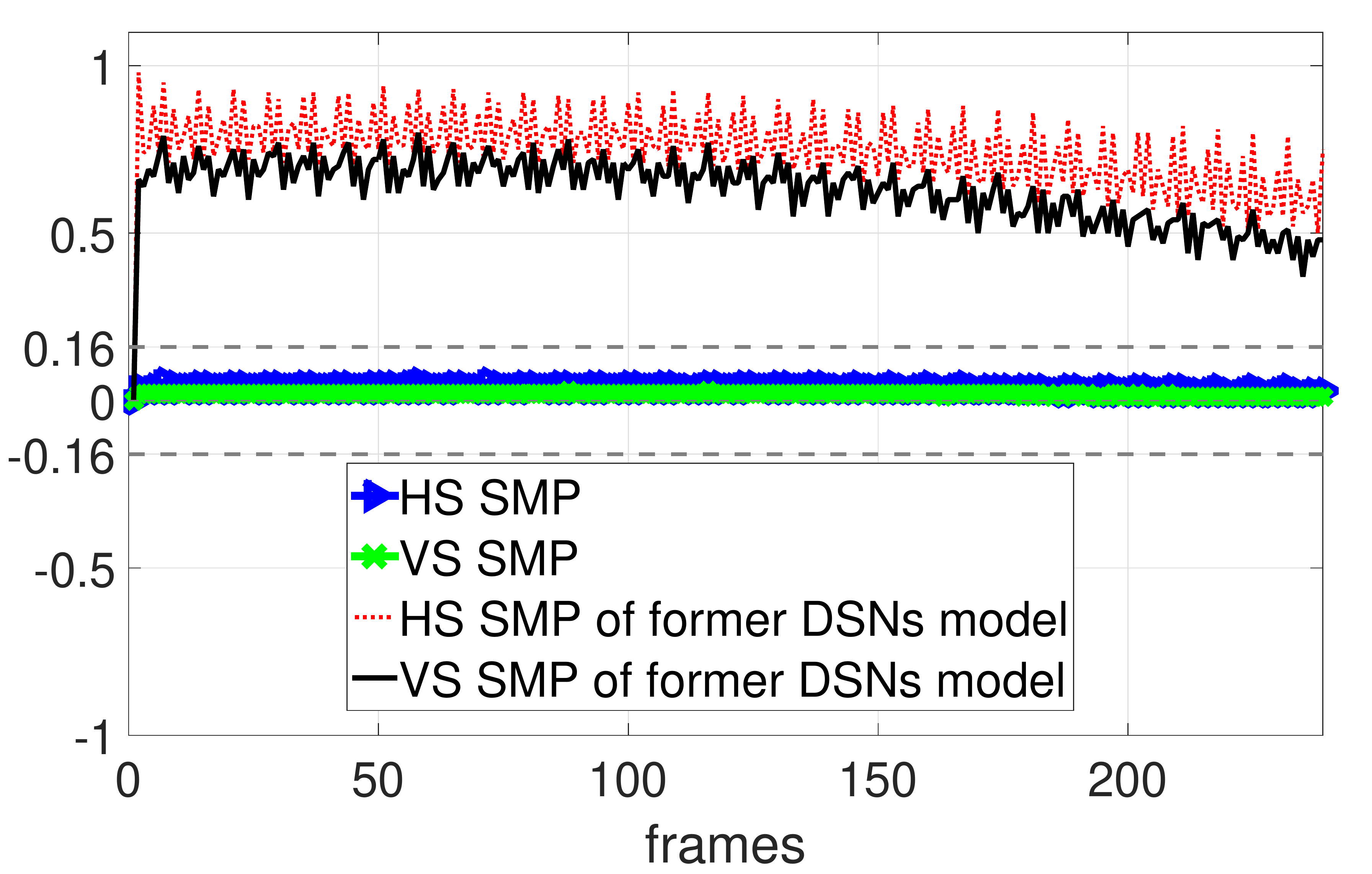}}
		\centerline{\scriptsize(b) a light object \textbf{approaching}}
	\end{minipage}
	\vfill
	\vspace{0.2in}
	\begin{minipage}[t]{0.5\linewidth}
		\centering
		\centerline{\includegraphics[width=1.5in]{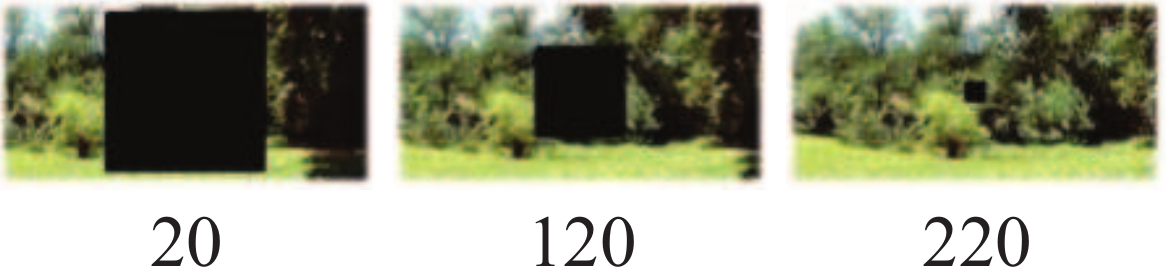}}
	\end{minipage}
	\hfill
	\begin{minipage}[t]{0.5\linewidth}
		\centering
		\centerline{\includegraphics[width=1.5in]{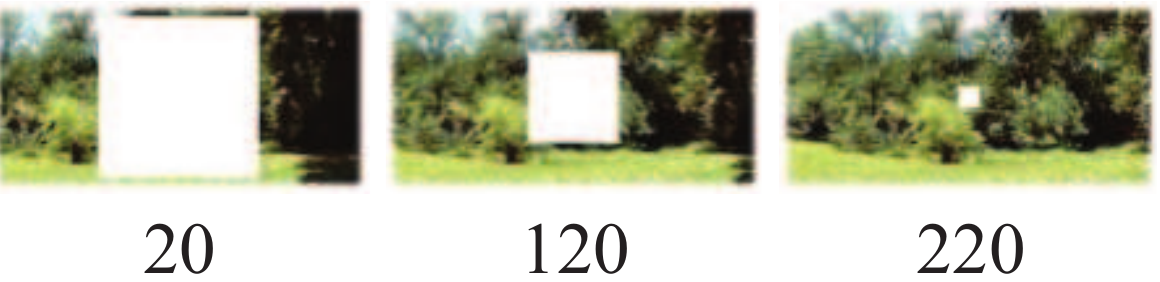}}
	\end{minipage}
	\vfill
	\begin{minipage}[t]{0.5\linewidth}
		\centering
		\centerline{\includegraphics[width=2in]{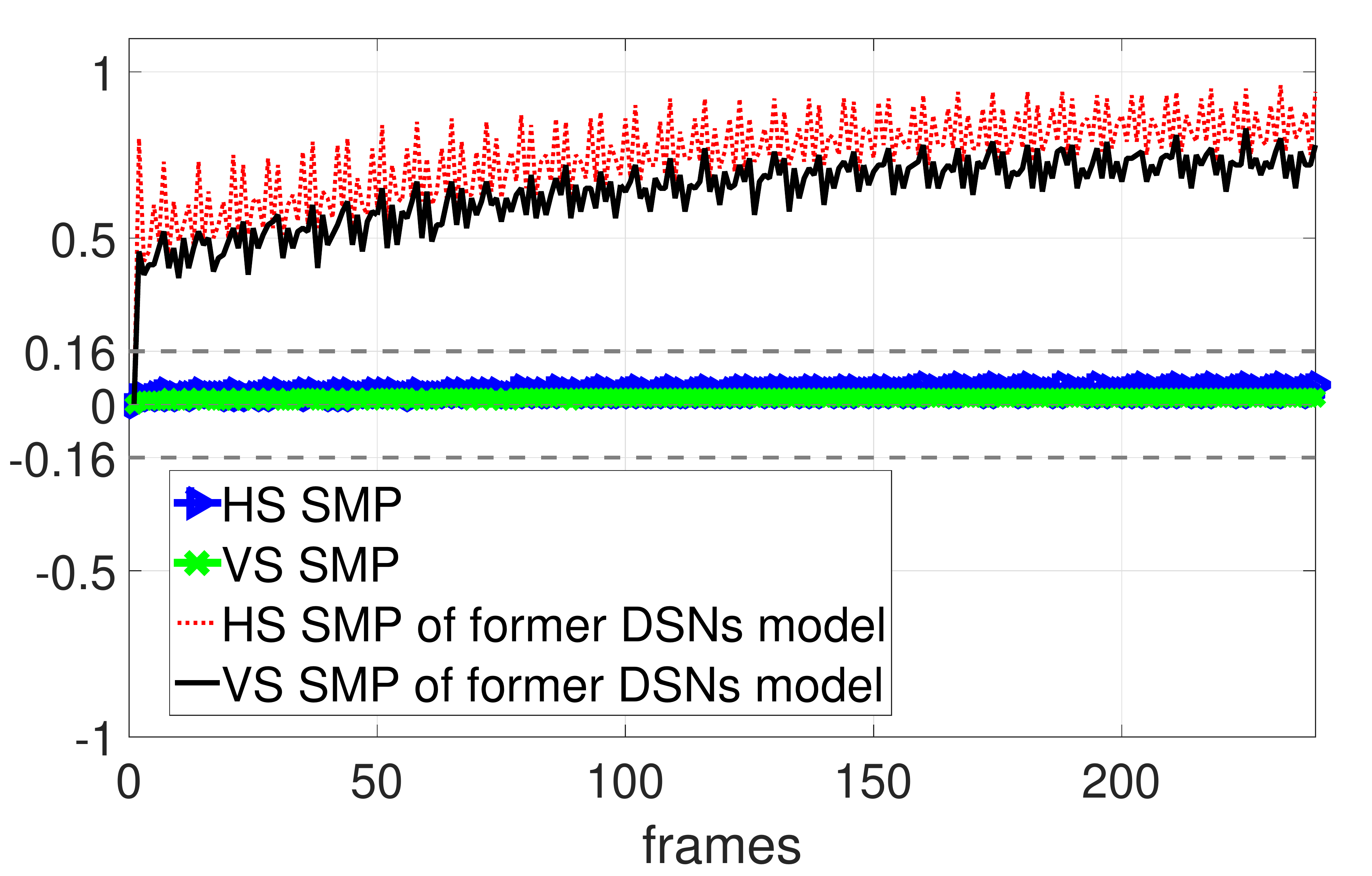}}
		\centerline{\scriptsize(c) a dark object \textbf{receding}}
	\end{minipage}
	\hfill
	\begin{minipage}[t]{0.5\linewidth}
		\centering
		\centerline{\includegraphics[width=2in]{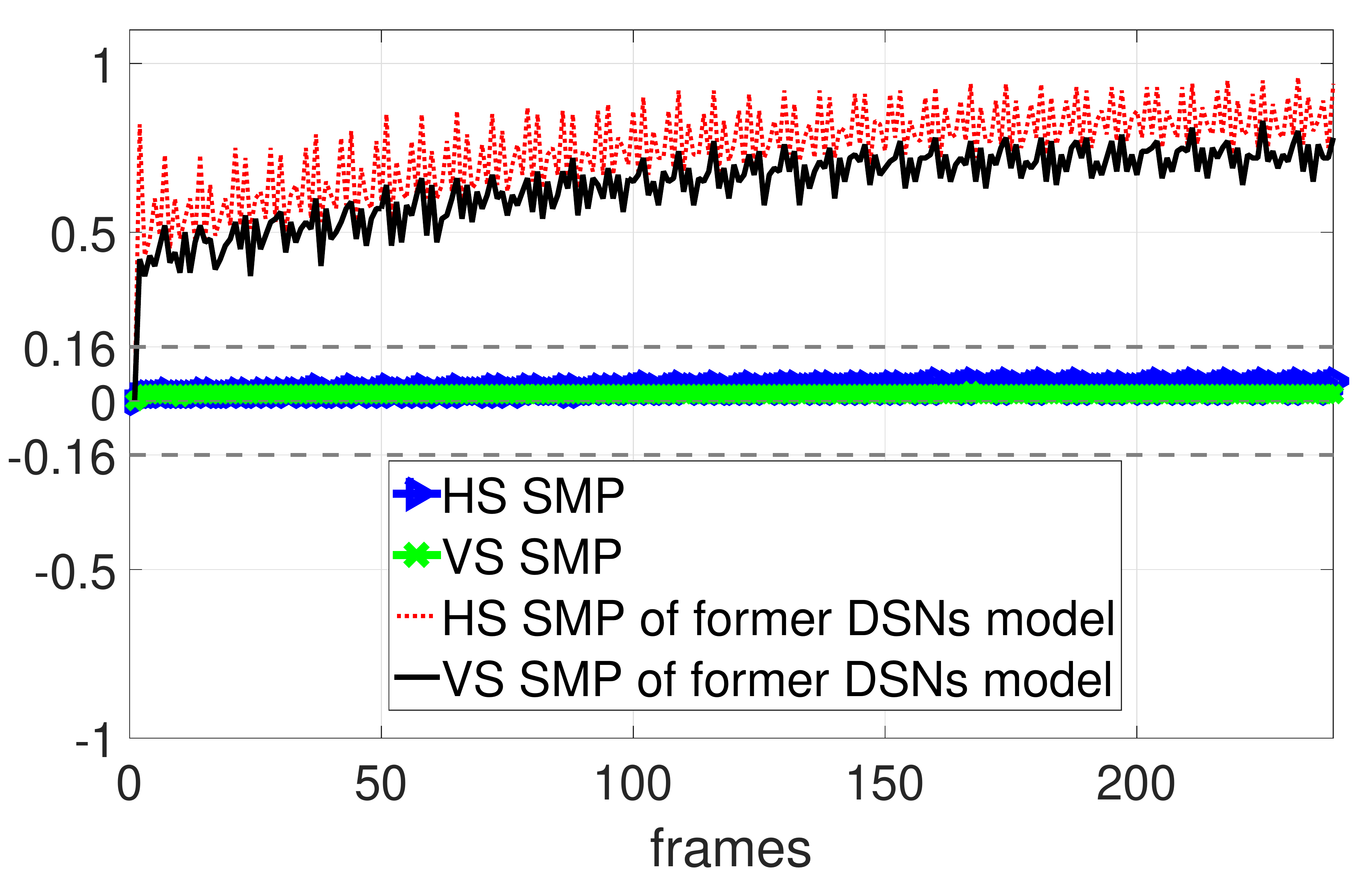}}
		\centerline{\scriptsize(d) a light object \textbf{receding}}
	\end{minipage}
	\caption{The DSNN and comparative DSNs model \cite{DSN-IJCNN} are challenged against dark and light objects approaching, receding embedded in a cluttered background, which is globally shifting rightward at the speed $Vb = 8$ (pixels per frame). The neural response of HS and VS systems of both models are depicted. The horizontal dashed lines indicate the spiking threshold ($\pm0.16$). \textbf{The DSNN remains quiet by looming and receding objects against shifting background; while the comparative model represents high-level response by all visual challenges}.}
	\label{simu-nature-depth}
\end{figure}

First, with similar ideas, we challenge the DSNN and two comparative models by the movements of dark and light objects approaching, receding and translating against the shifting of busy backgrounds separately. As shown in Fig. \ref{simu-nature-depth}, when challenged by the dark/light objects approaching and receding against the shifting of cluttered background, both HS and VS systems of the proposed DSNN remain quiet, the results of which perfectly match those in Fig. \ref{simu-clean}(e) and \ref{simu-clean}(f). On the other hand, both the HS and VS systems of the former model are greatly activated by approaching and receding stimuli. In our previous study \cite{DSN-IJCNN}, the comparative DSNs model demonstrated robust performance in extracting useful translational motion cues from a cluttered but stationary background via the modeling of a spatial pre-filtering mechanism prior to the ON and OFF pathways. However, we found that it was greatly affected by the shifting of cluttered backgrounds, the situation of which may never happen in the fly's visual system. Therefore, this research provides an important implication that a robust artificial motion detector requires a spatiotemporal process to filter out irrelevant motion from relevant motion. In the proposed framework, we demonstrate a bio-plausible solution by modeling an `FDSR' temporal mechanism in the motion-detecting pathways to enhance the ability of extracting useful motion cues from a visually cluttered environment.
\begin{figure}[t]
	\begin{minipage}[t]{0.5\linewidth}
		\centering
		\centerline{\includegraphics[width=1.5in]{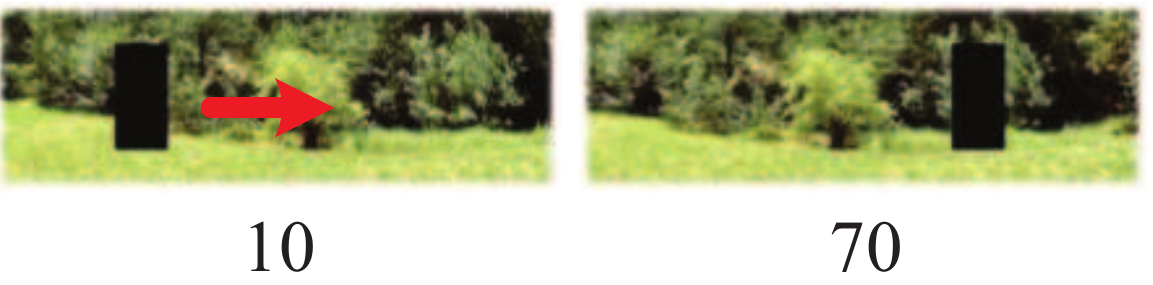}}
	\end{minipage}
	\hfill
	\begin{minipage}[t]{0.5\linewidth}
		\centering
		\centerline{\includegraphics[width=1.5in]{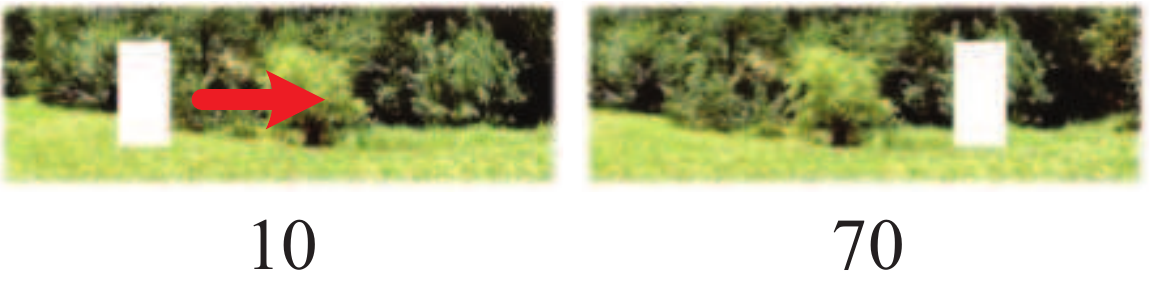}}
	\end{minipage}
	\vfill
	\begin{minipage}[t]{0.5\linewidth}
		\centering
		\centerline{\includegraphics[width=2in]{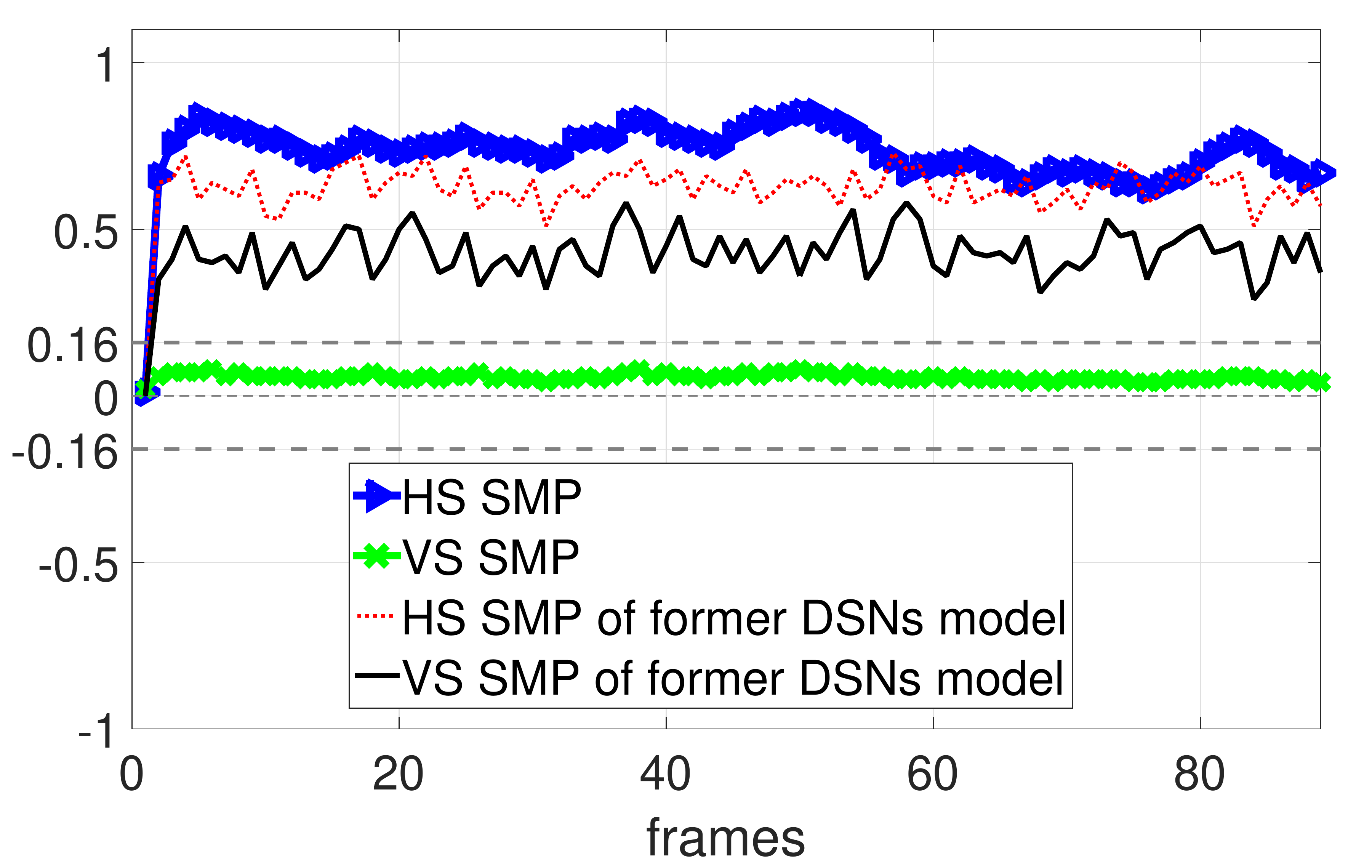}}
		\centerline{\scriptsize(a) a dark object \textbf{translating rightward}}
	\end{minipage}
	\hfill
	\begin{minipage}[t]{0.5\linewidth}
		\centering
		\centerline{\includegraphics[width=2in]{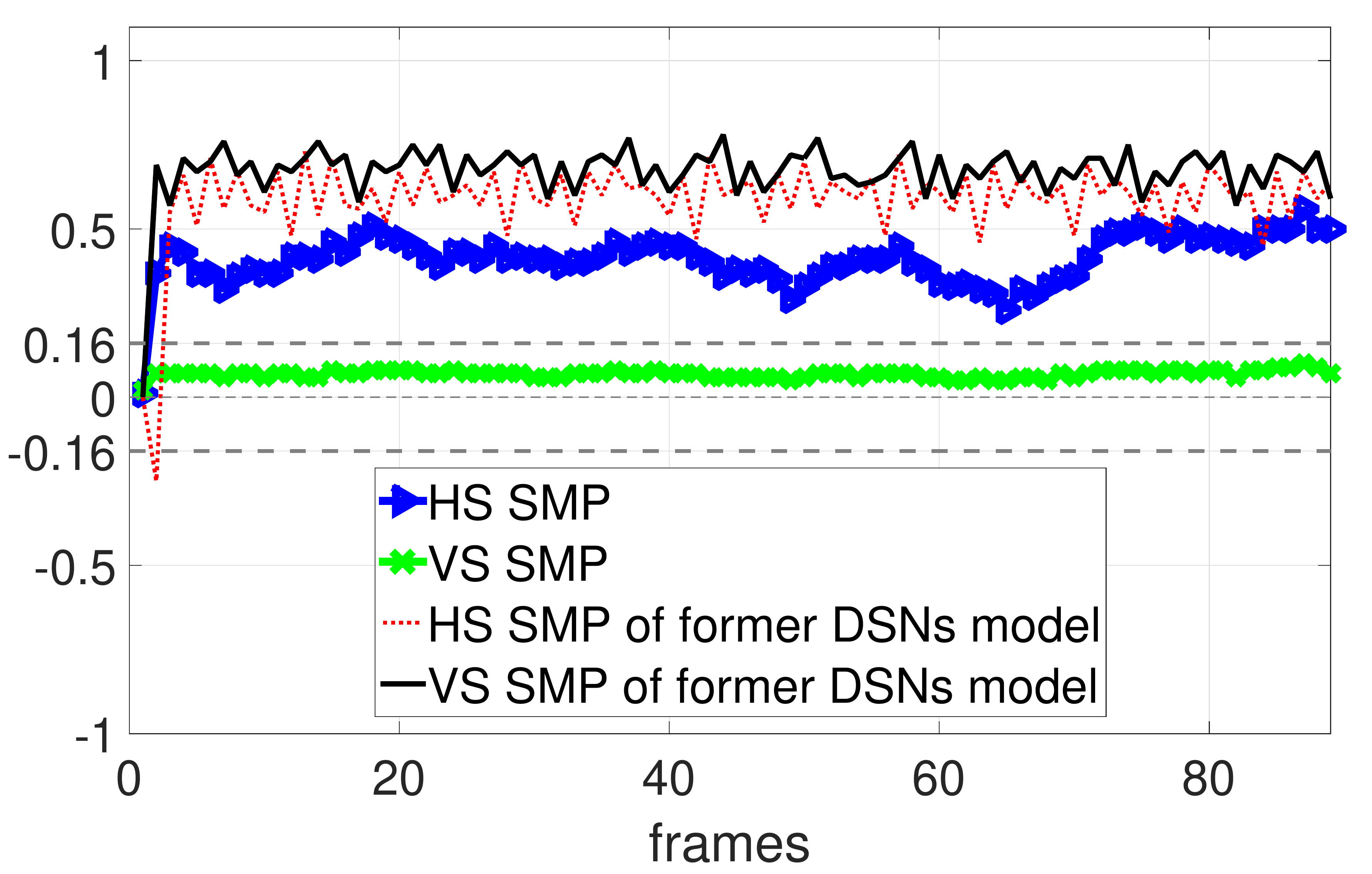}}
		\centerline{\scriptsize(b) a light object \textbf{translating rightward}}
	\end{minipage}
	\caption{The DSNN and comparative DSNs model \cite{DSN-IJCNN} are challenged against dark and light objects translating rightward embedded in a shifting cluttered background, which is moving in the opposite direction $Vb = -8$ (pixels per frame). \textbf{Only the HS system of DSNN is rigorously activated by horizontal translations against shifting background; while both HS and VS systems of the comparative model are highly activated}.}
	\label{simu-nature-trans}
\end{figure}

\begin{figure}[t]
	\begin{minipage}[t]{0.23\linewidth}
		\centering
		\centerline{\includegraphics[width=1.22in]{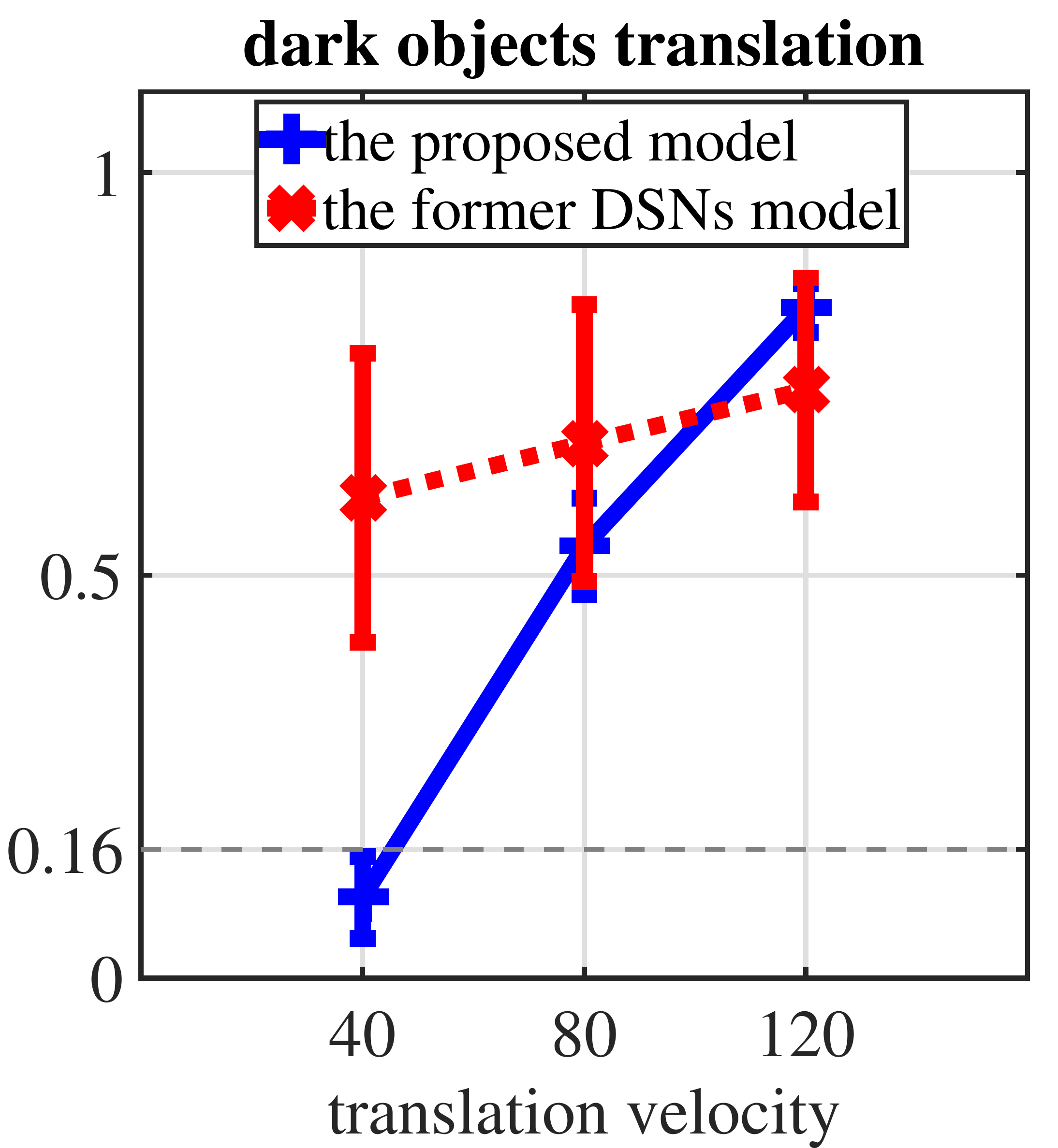}}
		\centerline{\scriptsize(a) \textbf{HS} peak-response}
	\end{minipage}
	\hfill
	\begin{minipage}[t]{0.23\linewidth}
		\centering
		\centerline{\includegraphics[width=1.2in]{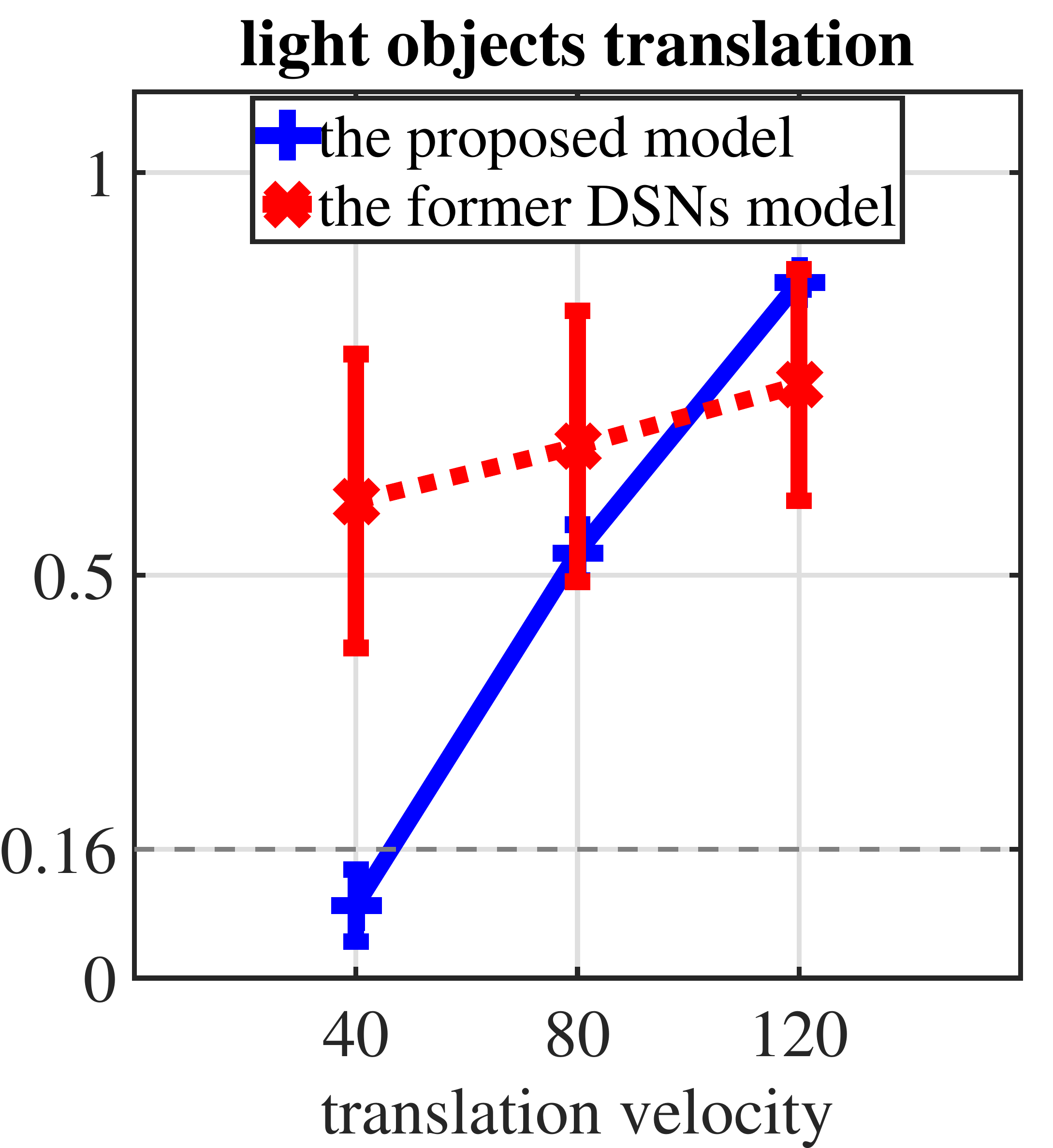}}
		\centerline{\scriptsize(b) \textbf{HS} peak-response}
	\end{minipage}
	\hfill
	\begin{minipage}[t]{0.23\linewidth}
		\centering
		\centerline{\includegraphics[width=1.2in]{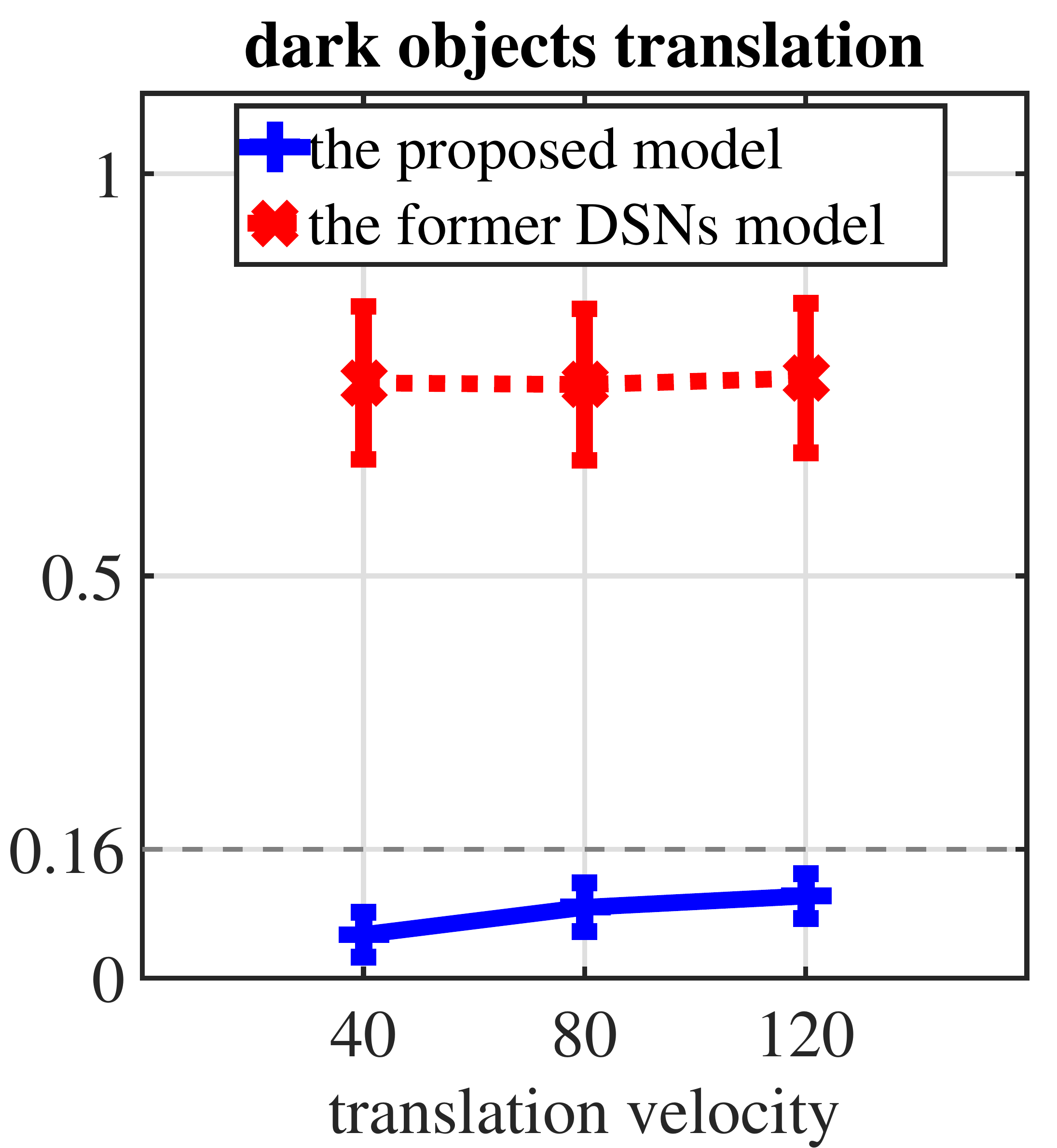}}
		\centerline{\scriptsize(c) \textbf{VS} peak-response}
	\end{minipage}
	\hfill
	\begin{minipage}[t]{0.23\linewidth}
		\centering
		\centerline{\includegraphics[width=1.19in]{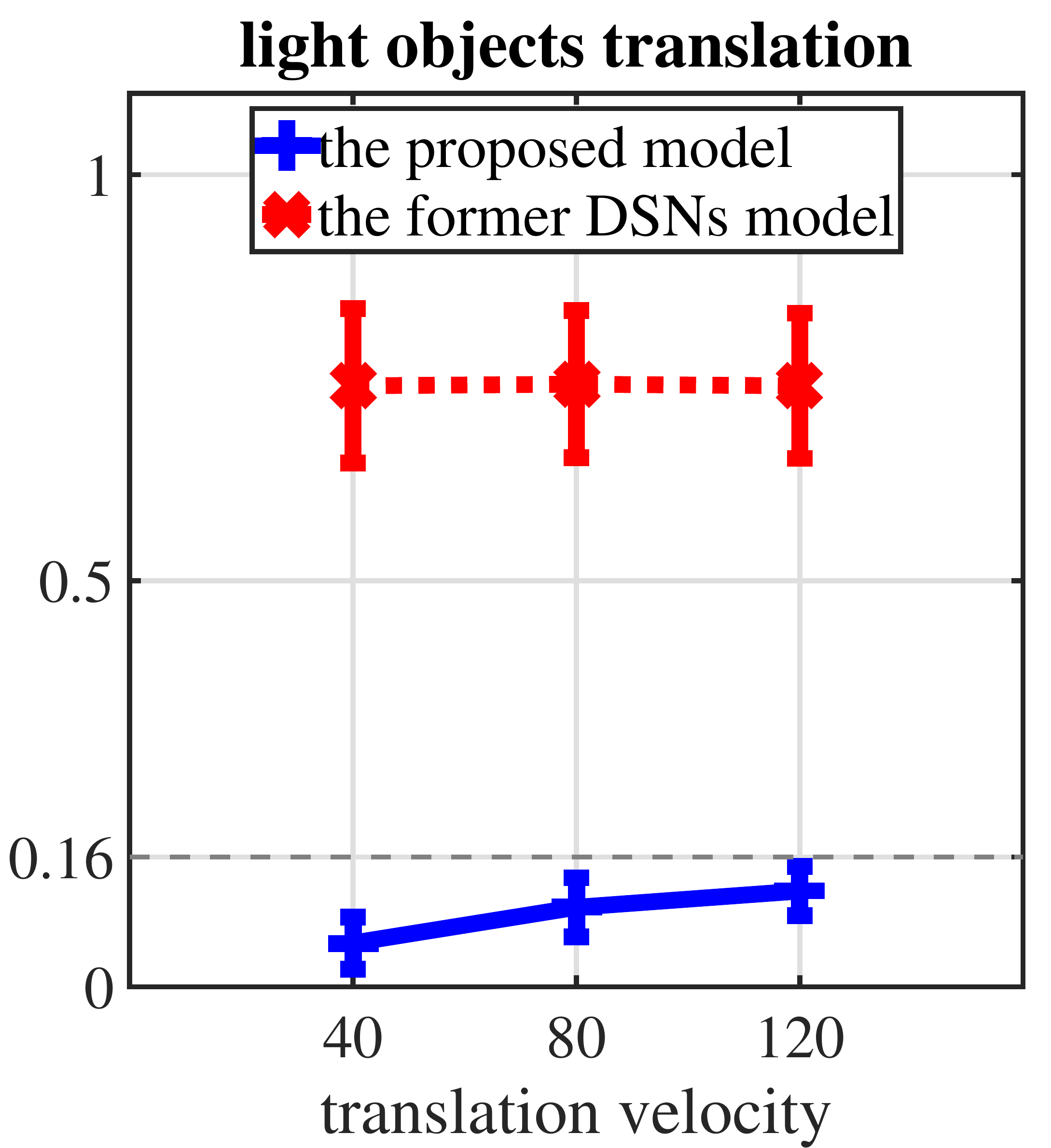}}
		\centerline{\scriptsize(d) \textbf{VS} peak-response}
	\end{minipage}
	\caption{The statistical results of peak SMP of the comparative models, tested by the dark and light objects translating rightward at three individual velocities: $40$, $80$, $120$ (pixels per frame), each against the leftward shifting background, at five velocities: $-2$, $-4$, $-6$, $-8$, $-10$ (p/f) respectively. The horizontal dashed line indicates the spiking threshold ($0.16$). \textbf{The DSNN represents better speed response and the VS system completely remains quiet indicating robust performance against shifting cluttered backgrounds}.}
	\label{simu-sys-speed}
\end{figure}

On the aspect of being challenged by rightward translation movements along with the shifting of cluttered background in an opposite direction, as illustrated in Fig. \ref{simu-nature-trans}, both the HS systems of comparative neural networks produce successively positive membrane potential. The VS system of DSNN remains inactive, whilst the VS system of the comparative model is highly activated by the moving background as well. For deepening our understanding of the advantages of DSNN, we systematically test both neural networks with visual stimuli of both dark and light objects translating rightward, at three speed levels, all embedded in the natural background, shifting leftward at five speed levels respectively. The statistics illustrated in Fig. \ref{simu-sys-speed} allow the following conclusions to be drawn: both comparative models show speed response to translating stimuli at varied velocities; the HS system of DSNN represents a more significantly rising with larger gradient and smaller invariance of peak response, implying more stable performance when tested by the shifting of cluttered background at varied velocities (Fig. \ref{simu-sys-speed}(a), \ref{simu-sys-speed}(b)). More importantly, Fig. \ref{simu-sys-speed}(c), \ref{simu-sys-speed}(d) demonstrate that the peak responses of the VS system of DSNN are all below the predefine spiking threshold, whilst the VS system of the former DSNs model is highly activated, the results of which match Fig. \ref{simu-nature-trans}. Informative results prove that the proposed DSNN outperforms the former DSNs model from our previous work when challenged by shifting cluttered background with higher degree of complexity, a situation which is similar to animals' self-motion in navigation.

\begin{figure}[t]
	\begin{minipage}[t]{0.5\linewidth}
		\centering
		\centerline{\includegraphics[width=2.2in]{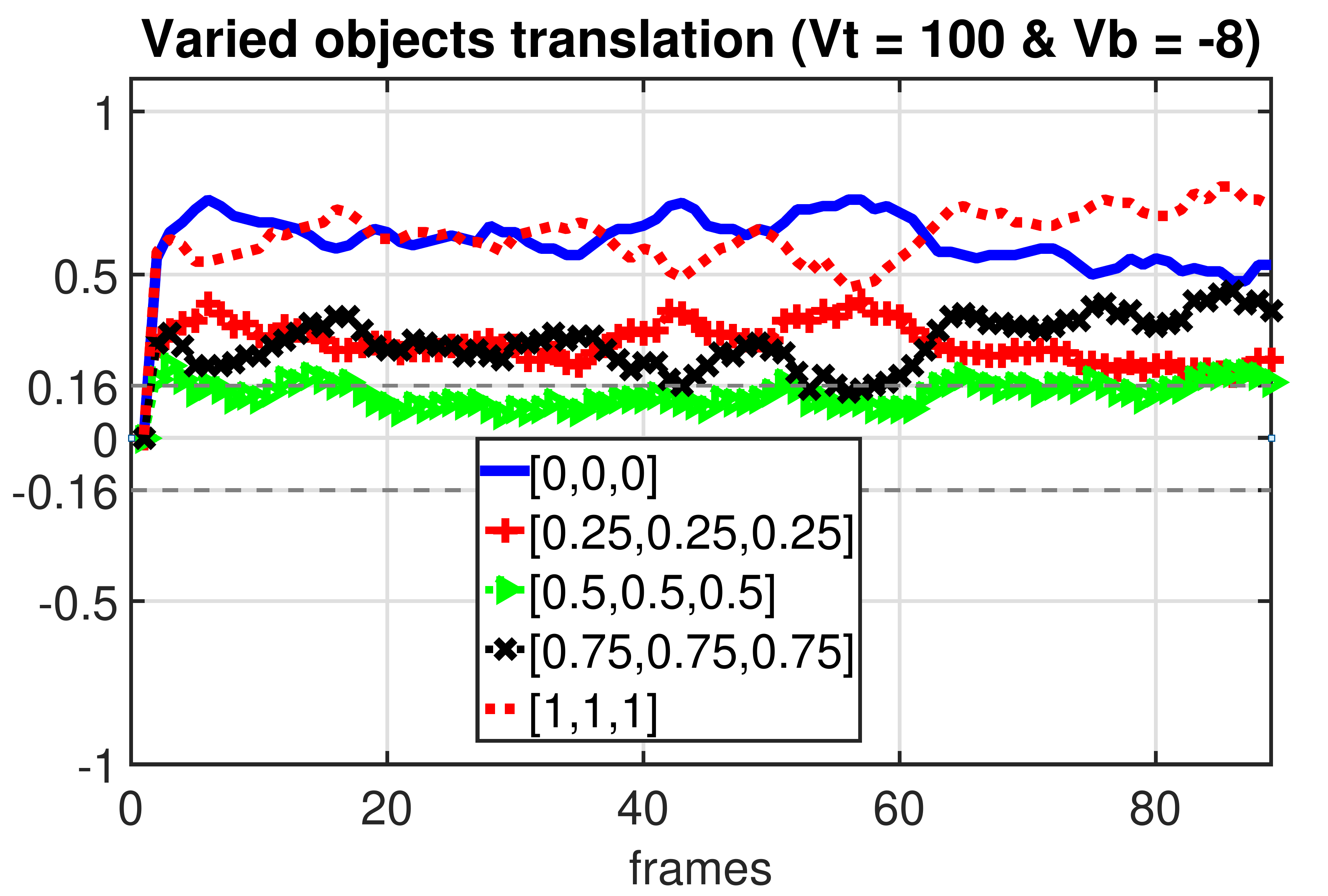}}
		\centerline{\scriptsize(a)}
	\end{minipage}
	\hfill
	\begin{minipage}[t]{0.5\linewidth}
		\centering
		\centerline{\includegraphics[width=2.2in]{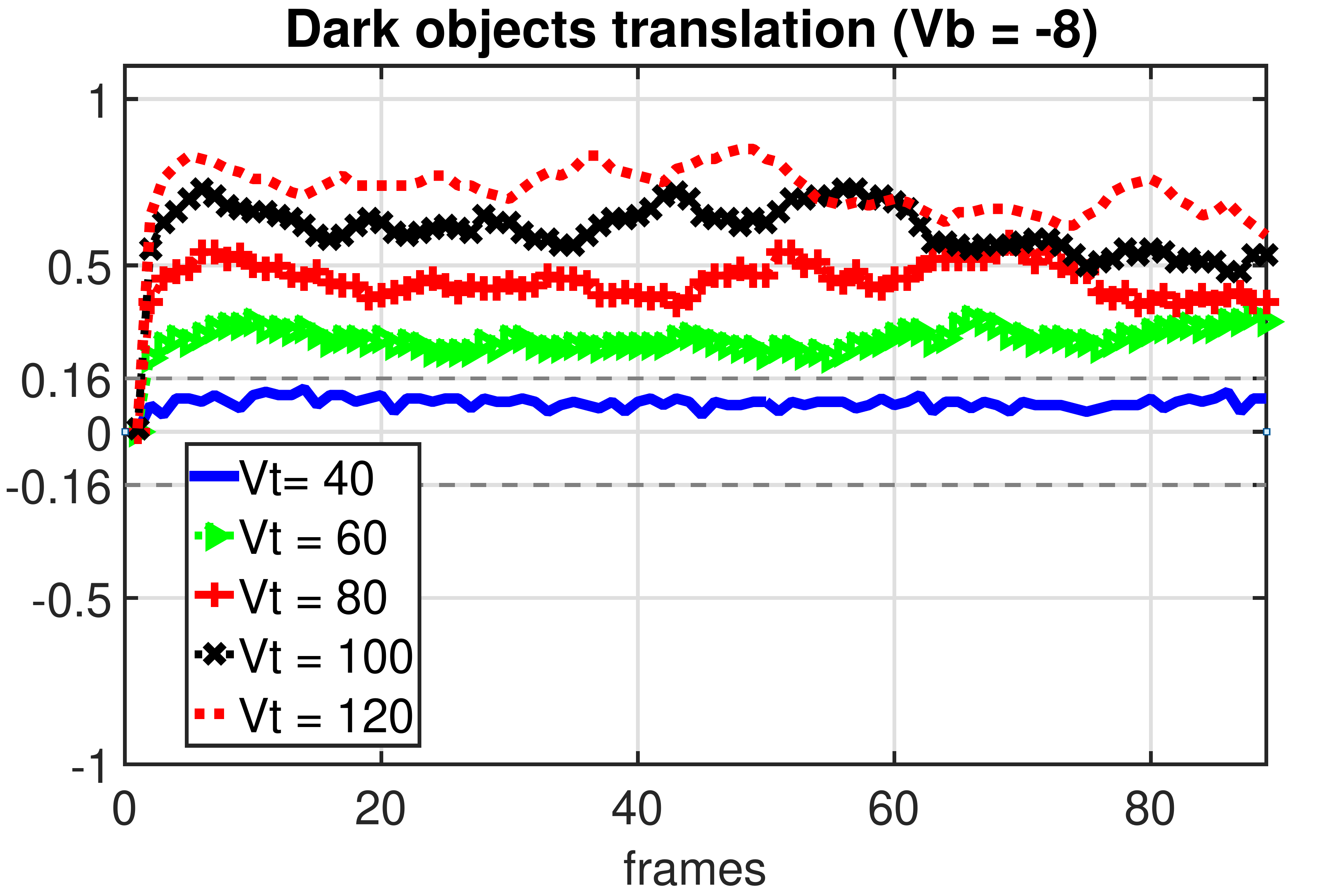}}
		\centerline{\scriptsize(b)}
	\end{minipage}
	\caption{The neural response of the HS system of DSNN, challenged by five gray-scaled objects translating rightward respectively, at an identical speed (a); and by a same gray-scaled object translating at five velocities respectively (b). All movements were embedded in a cluttered background shifting leftward $Vb=-8$ (pixels per frame). The horizontal dashed lines indicate the spiking thresholds ($\pm0.16$). \textbf{The DSNN demonstrates speed response and contrast sensitivity to translations}.}
	\label{simu-sys-grayscale-mp}
\end{figure}

\begin{figure}[t]
	\begin{minipage}[t]{0.5\linewidth}
		\centering
		\centerline{\includegraphics[width=2.2in]{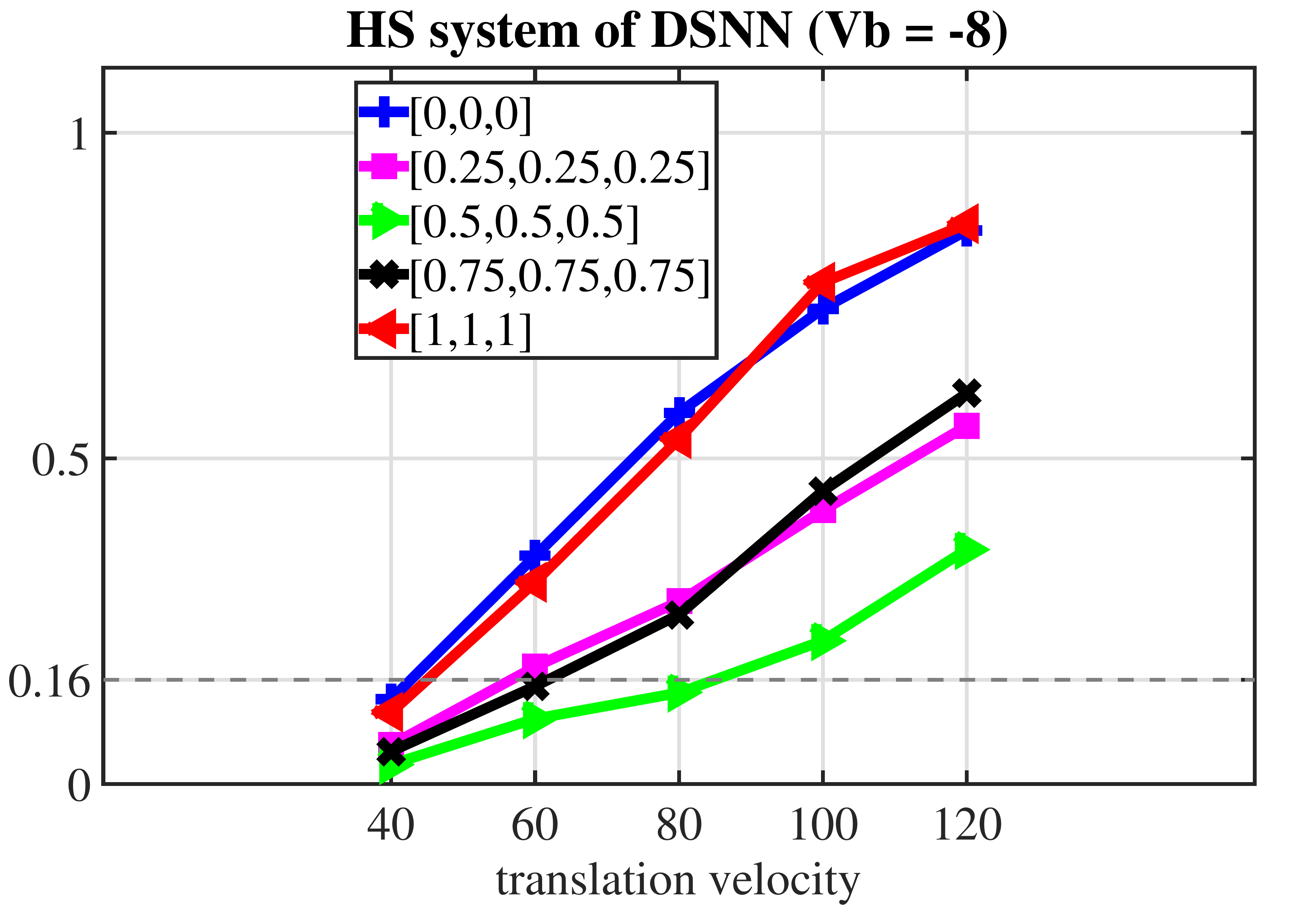}}
		\centerline{\scriptsize(a)}
	\end{minipage}
	\hfill
	\begin{minipage}[t]{0.5\textwidth}
		\centering
		\centerline{\includegraphics[width=2.2in]{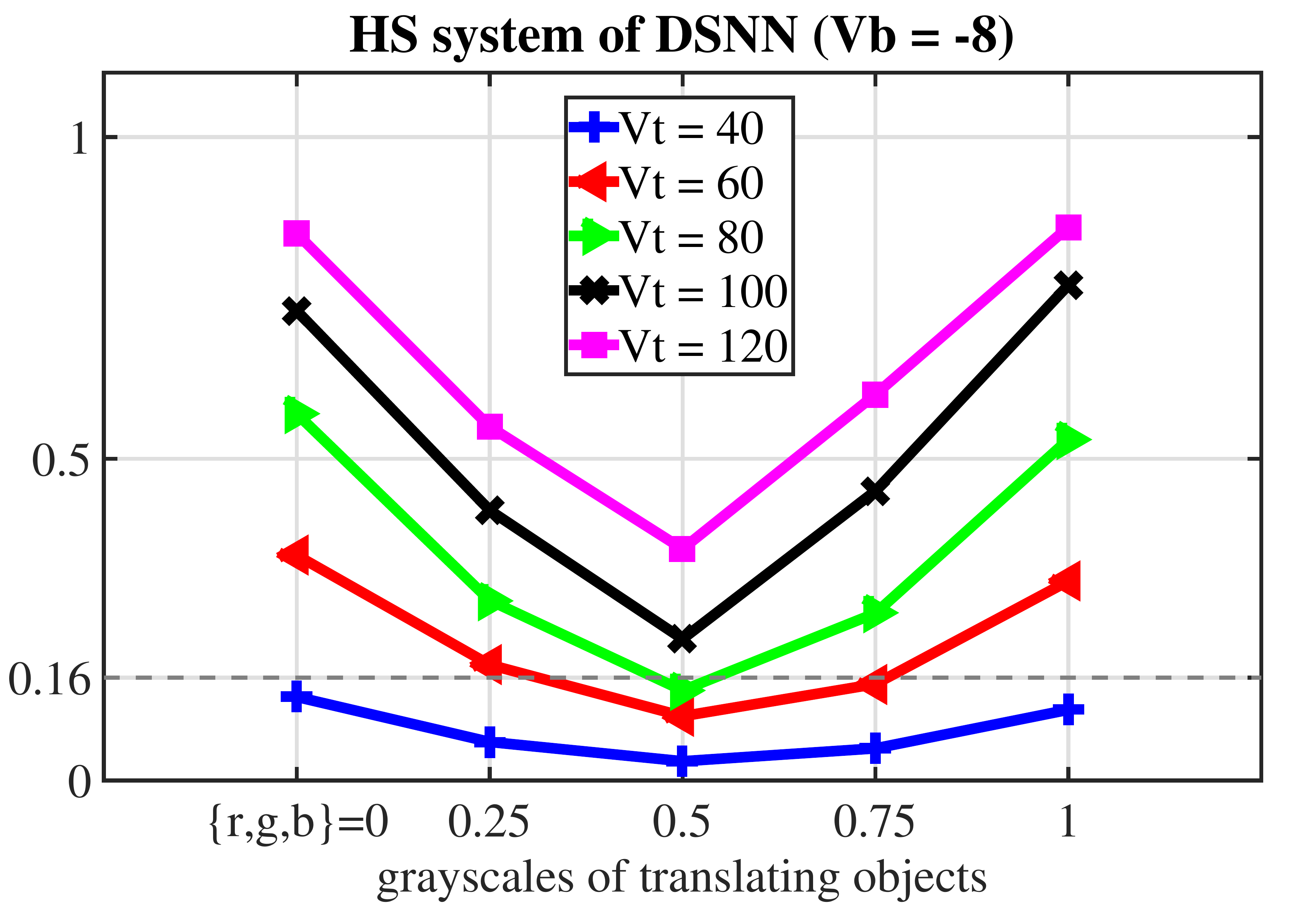}}
		\centerline{\scriptsize(b)}
	\end{minipage}
	\vfill
	\vspace{0.05in}
	\begin{minipage}[t]{0.5\textwidth}
		\centering
		\centerline{\includegraphics[width=2.2in]{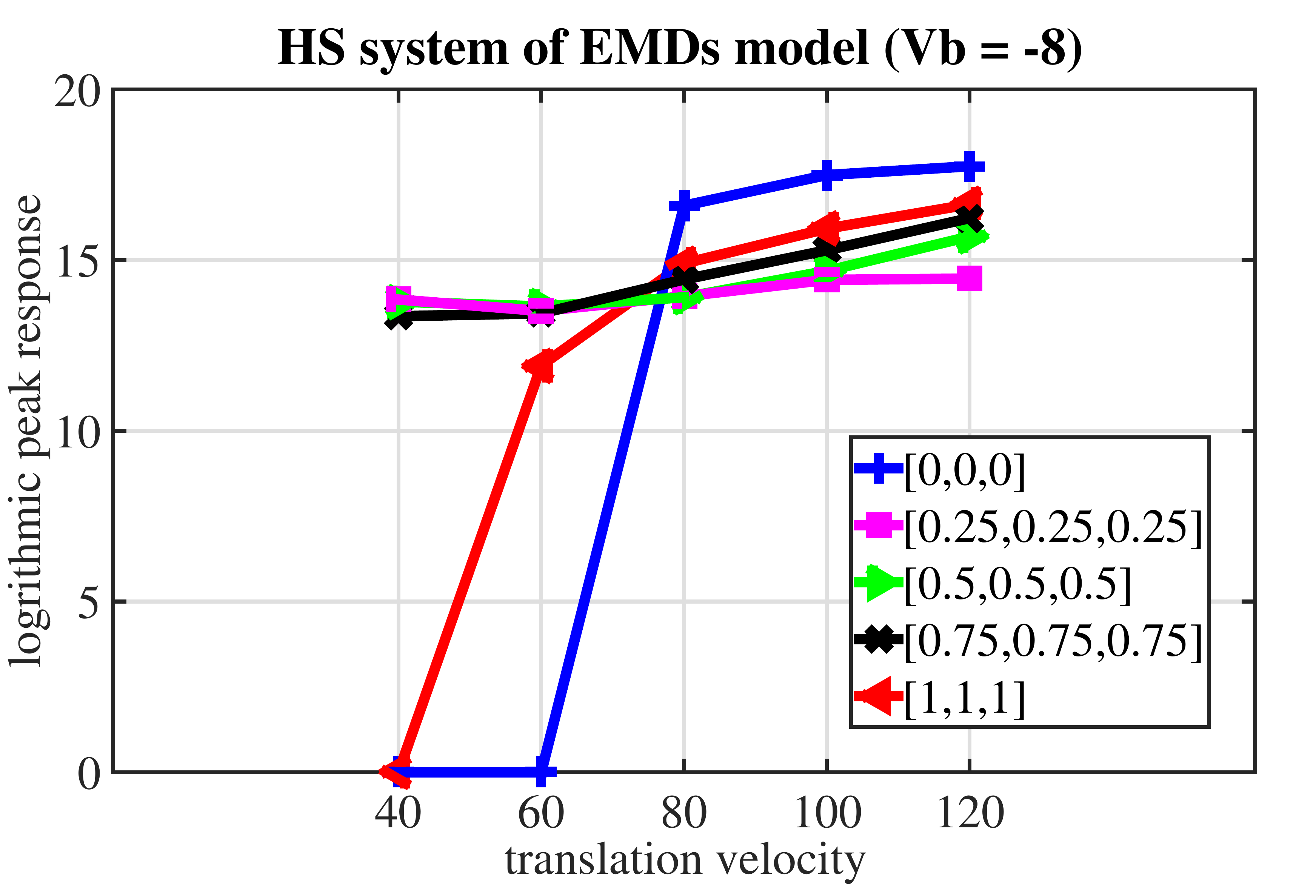}}
		\centerline{\scriptsize(c)}
	\end{minipage}
	\hfill
	\begin{minipage}[t]{0.5\linewidth}
		\centering
		\centerline{\includegraphics[width=2.2in]{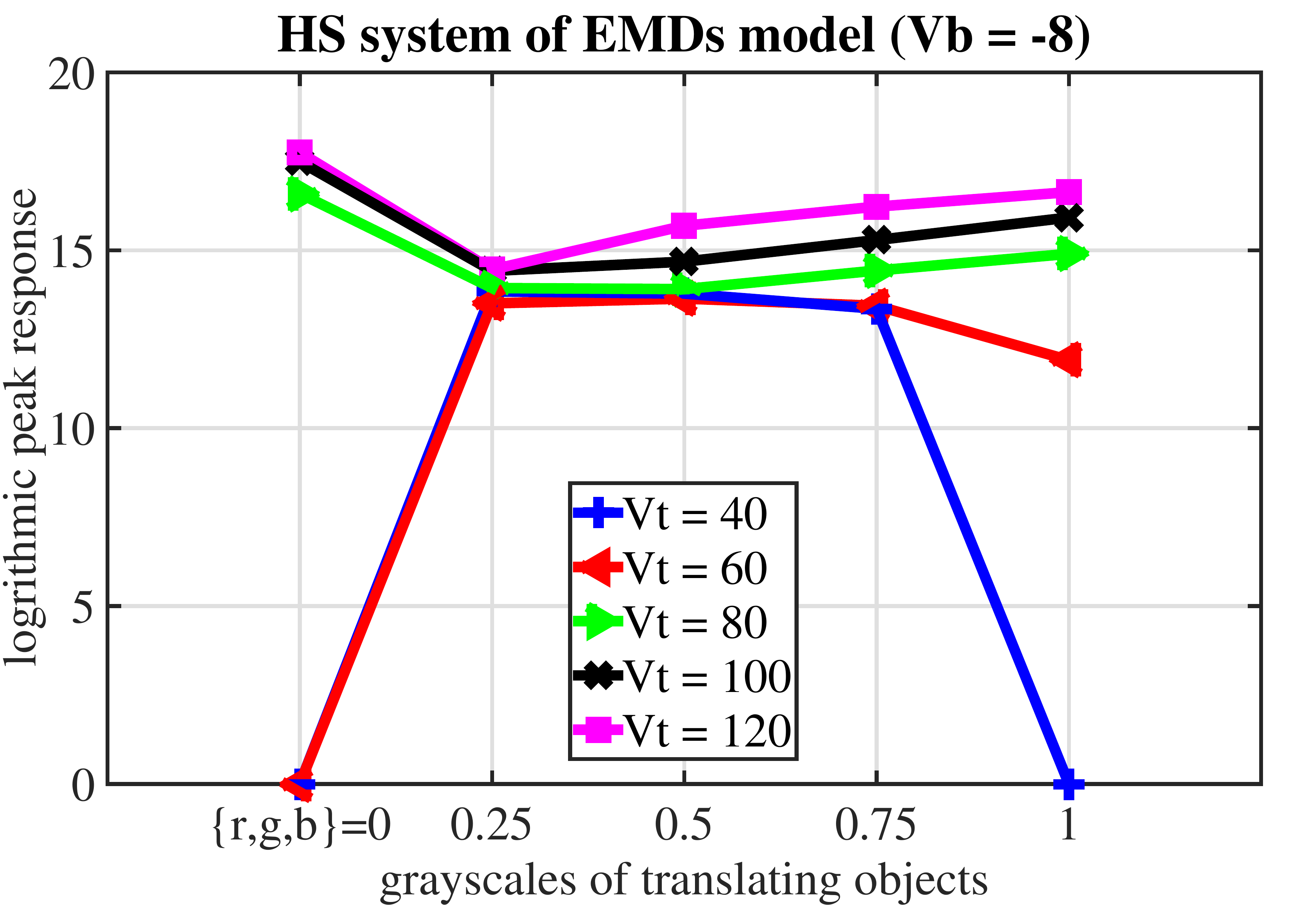}}
		\centerline{\scriptsize(d)}
	\end{minipage}
	\caption{The statistical results of peak-response generated by HS systems of two comparative models -- the proposed DSNN and an EMDs model \cite{Iida_2000}, challenged by five gray-scaled objects translating rightward, at five velocities ($Vt$) respectively. All movements are embedded in a cluttered background shifting leftward $Vb=-8$. (a), (b) The peak-SMPs of the HS system of DSNN; (c), (d) The logarithmic peak-response of the HS system of EMDs. \textbf{The proposed DSNN, with stable speed response and contrast sensitivity at all tested translations, outperforms the former EMDs model.}}
	\label{simu-sys-grayscale}
\end{figure}

Furthermore, we also compared the DSNN with an EMDs model \cite{Iida_2000}, for inspecting the effects of translating speed and contrast on peak neural responses, which are represented by the SMPs of DSNN and the logarithmic output of EMDs. First, the results in Fig. \ref{simu-sys-grayscale-mp} demonstrate satisfactory speed response and contrast sensitivity of the proposed DSNN, i.e., it produces stronger response to the translating stimuli at higher level of velocity, and is more sensitive to either darker or lighter moving features with relatively larger contrasts to the background. Second, contrary to the comparative EMDs model, the statistics in Fig. \ref{simu-sys-grayscale} clearly demonstrate that the proposed DSNN performs more robustly against the shifting of visually cluttered background, with better speed response to the translating of all gray-scaled (contrasts) objects: the SMPs of DSNN smoothly peak at higher level along with the increasing of translating speed (Fig. \ref{simu-sys-grayscale}(a)), while the comparative EMDs model only shows good speed response to translations of both the darkest and lightest objects (Fig. \ref{simu-sys-grayscale}(c)).

Fig. \ref{simu-sys-grayscale}(b), \ref{simu-sys-grayscale}(d) also demonstrate the proposed DSNN performs robustly on all tested gray-scaled objects at various translating velocities, i.e., it can perceive and retrieve useful motion cues of the foreground translating objects, even at the lowest velocity or with the smallest contrast, from the shifting of cluttered background. Intuitively, the peak-SMPs of DSNN all reach the valley in the translation of medium gray-scaled object with relatively smaller contrast to the cluttered background (Fig. \ref{simu-sys-grayscale}(b)). On the contrary, the EMDs model is not able to detect all gray-scaled translating features at the lower velocities of $40$ and $60$ pixels per frame, against the shifting of natural background (Fig. \ref{simu-sys-grayscale}(d)). The DSNN better represents speed response and contrast sensitivity to translational motion cues, especially in complex and dynamic scenes.

To briefly summarize all the off-line synthetic stimuli tests, first we have shown the proposed DSNN possesses similar abilities to DSNs in the fly's visual system for perceiving translational motion cues rather than other kinds of stimuli, like the movements of approaching and receding. More importantly, through systematic tests, we compared the DSNN's performance with two related models. We found that the DSNN outperforms the comparative models and may provide useful solutions for solving the shortcomings of previous translational motion sensitive systems mentioned in Section \ref{introduction}.

\subsection{Real world stimuli tests}
\begin{figure}[!t]
	\begin{minipage}[t]{0.5\linewidth}
		\centering
		\centerline{\includegraphics[width=2in]{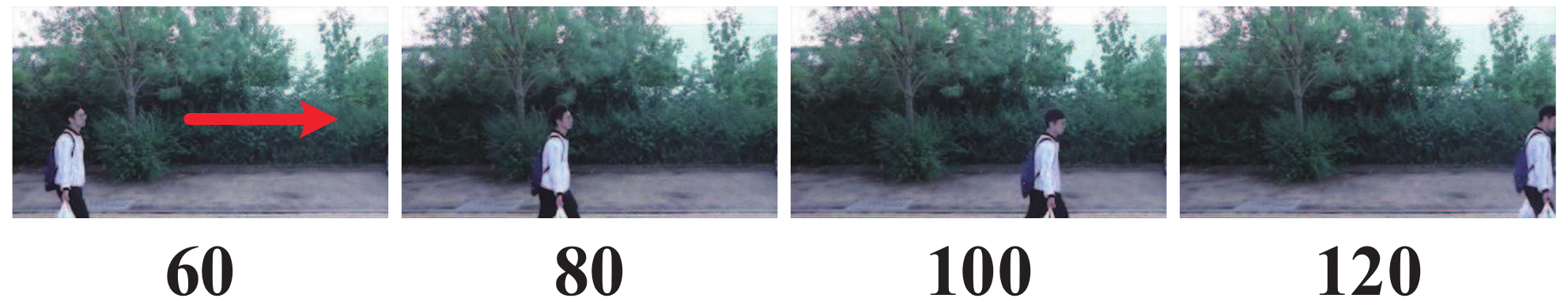}}
	\end{minipage}
	\hfill
	\begin{minipage}[t]{0.5\linewidth}
		\centering
		\centerline{\includegraphics[width=2in]{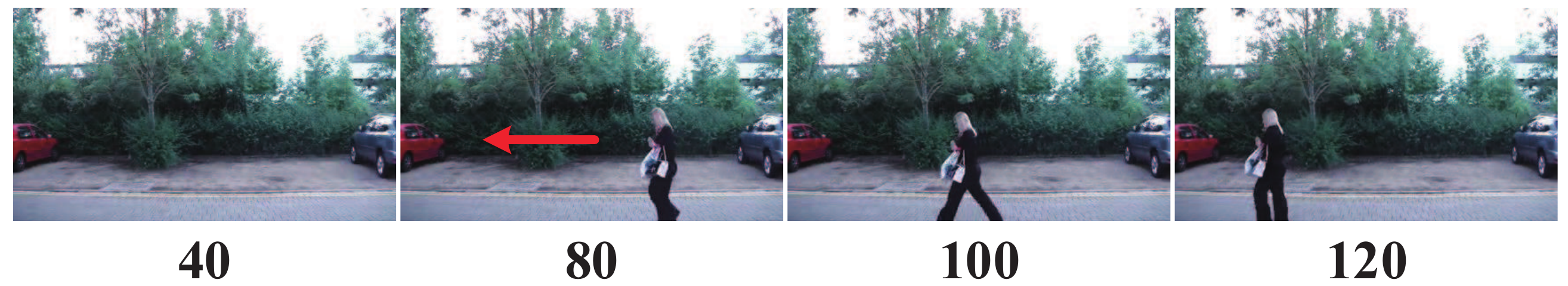}}
	\end{minipage}
	\vfill
	\begin{minipage}[t]{0.5\textwidth}
		\centering
		\centerline{\includegraphics[width=2.2in]{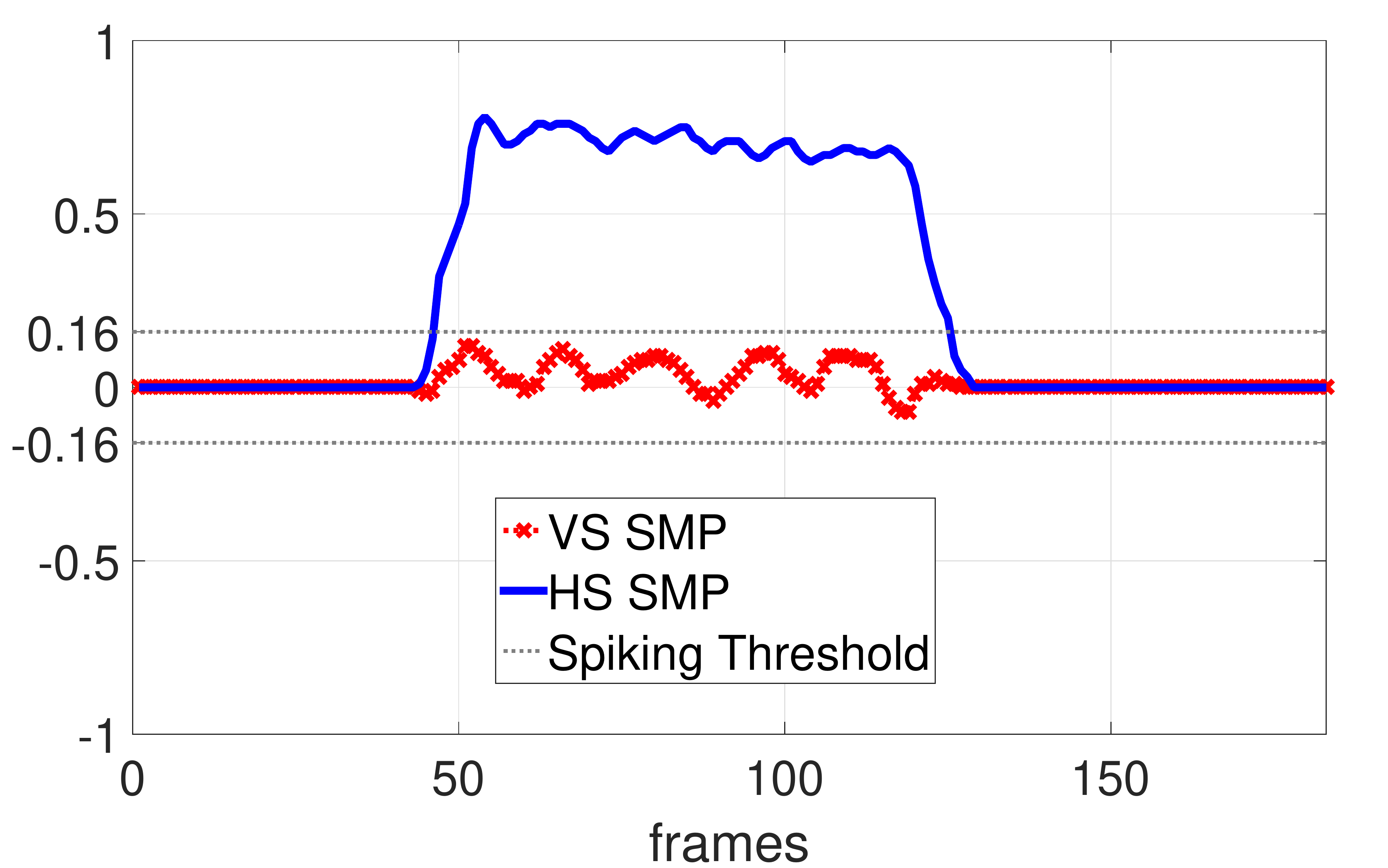}}
		\centerline{\scriptsize(a) a pedestrian translating rightward}
	\end{minipage}
	\hfill
	\begin{minipage}[t]{0.5\textwidth}
		\centering
		\centerline{\includegraphics[width=2.2in]{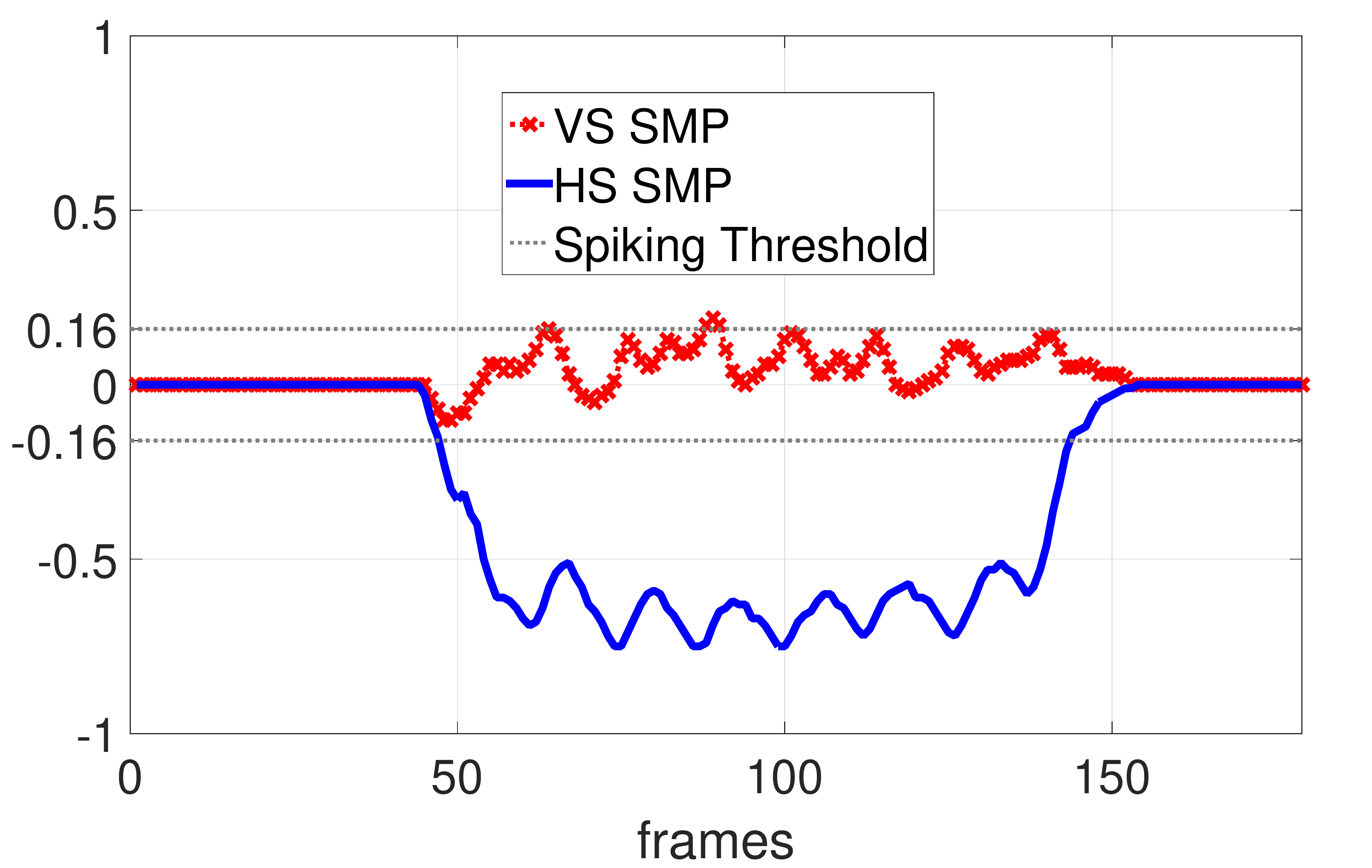}}
		\centerline{\scriptsize(b) a pedestrian translating leftward}
	\end{minipage}
	\vfill
	\vspace{0.05in}
	\begin{minipage}[t]{0.5\linewidth}
		\centering
		\centerline{\includegraphics[width=2in]{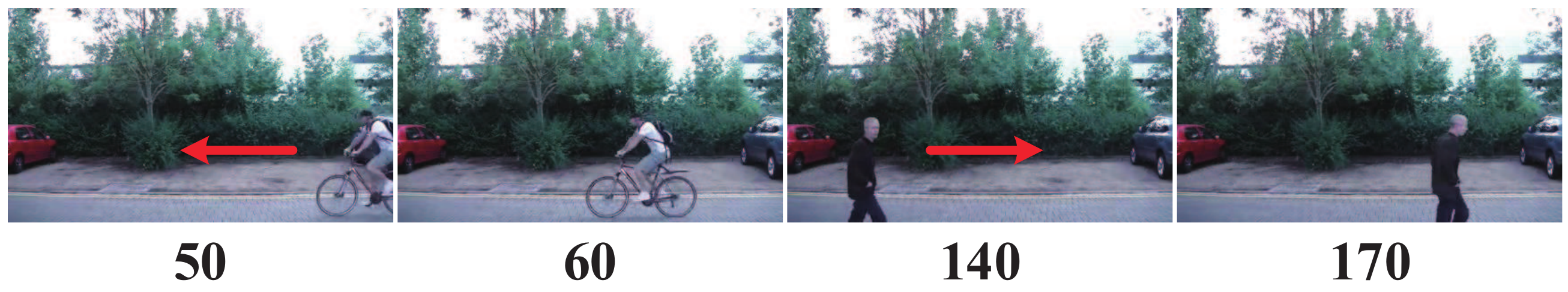}}
	\end{minipage}
	\hfill
	\begin{minipage}[t]{0.5\linewidth}
		\centering
		\centerline{\includegraphics[width=2in]{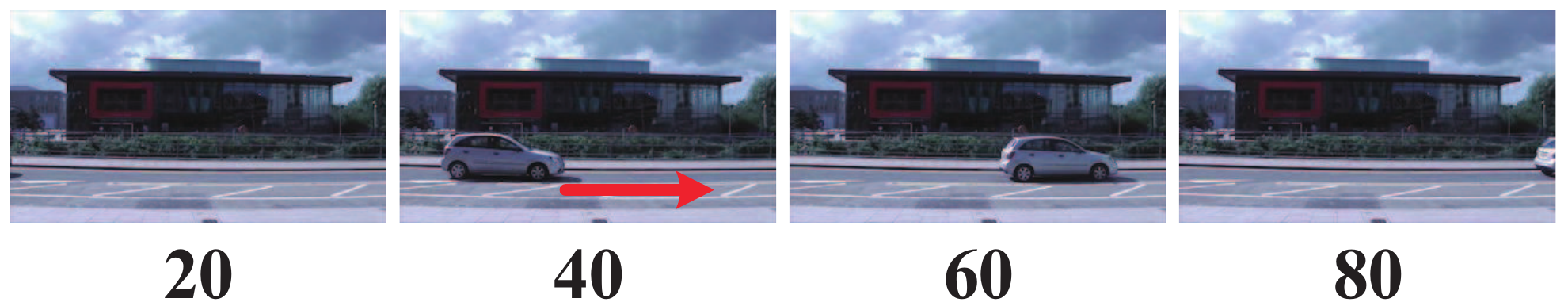}}
	\end{minipage}
	\vfill
	\begin{minipage}[t]{0.5\textwidth}
		\centering
		\centerline{\includegraphics[width=2.2in]{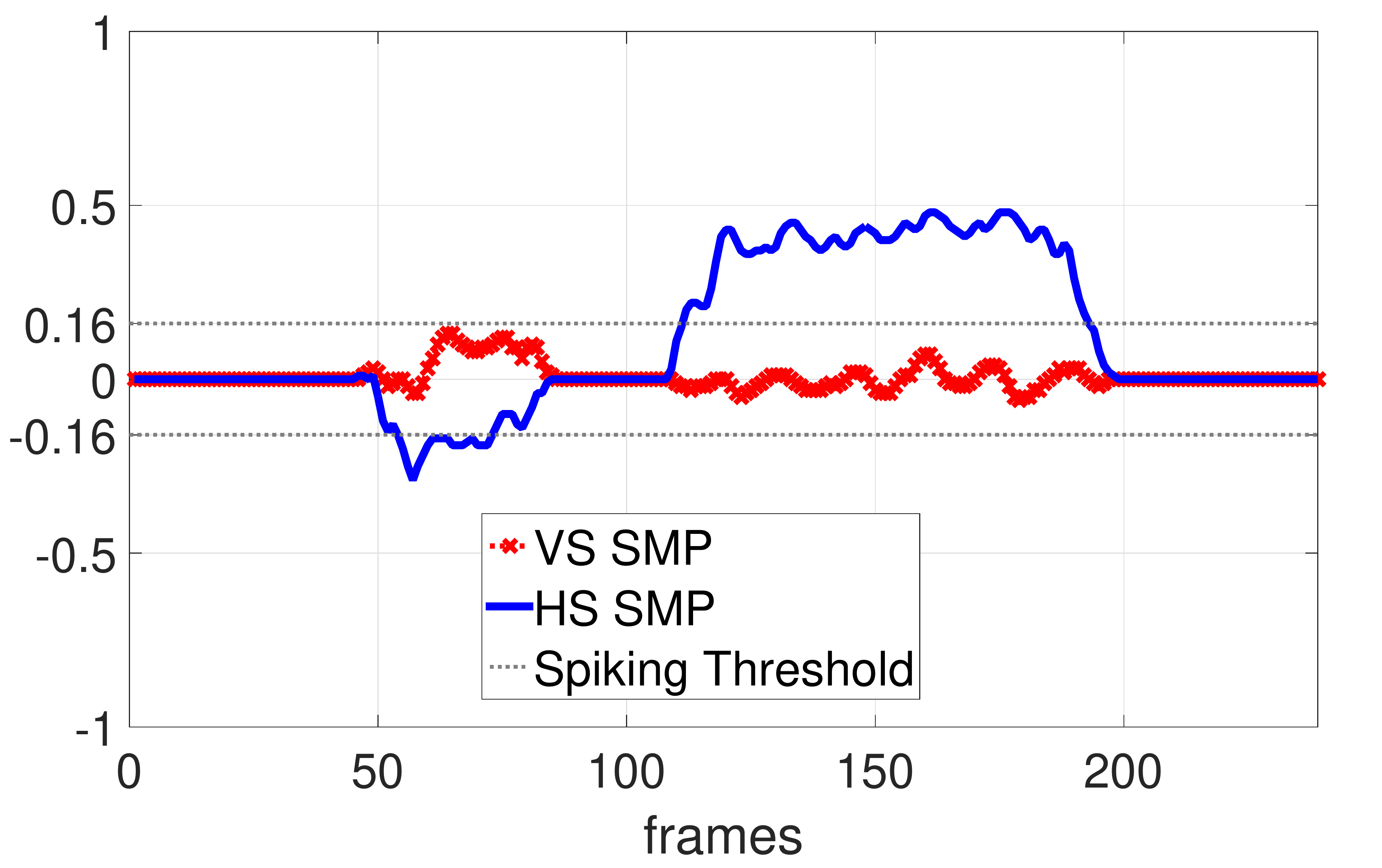}}
		\centerline{\scriptsize(c) translations in both horizontal directions}
	\end{minipage}
	\hfill
	\begin{minipage}[t]{0.5\textwidth}
		\centering
		\centerline{\includegraphics[width=2.2in]{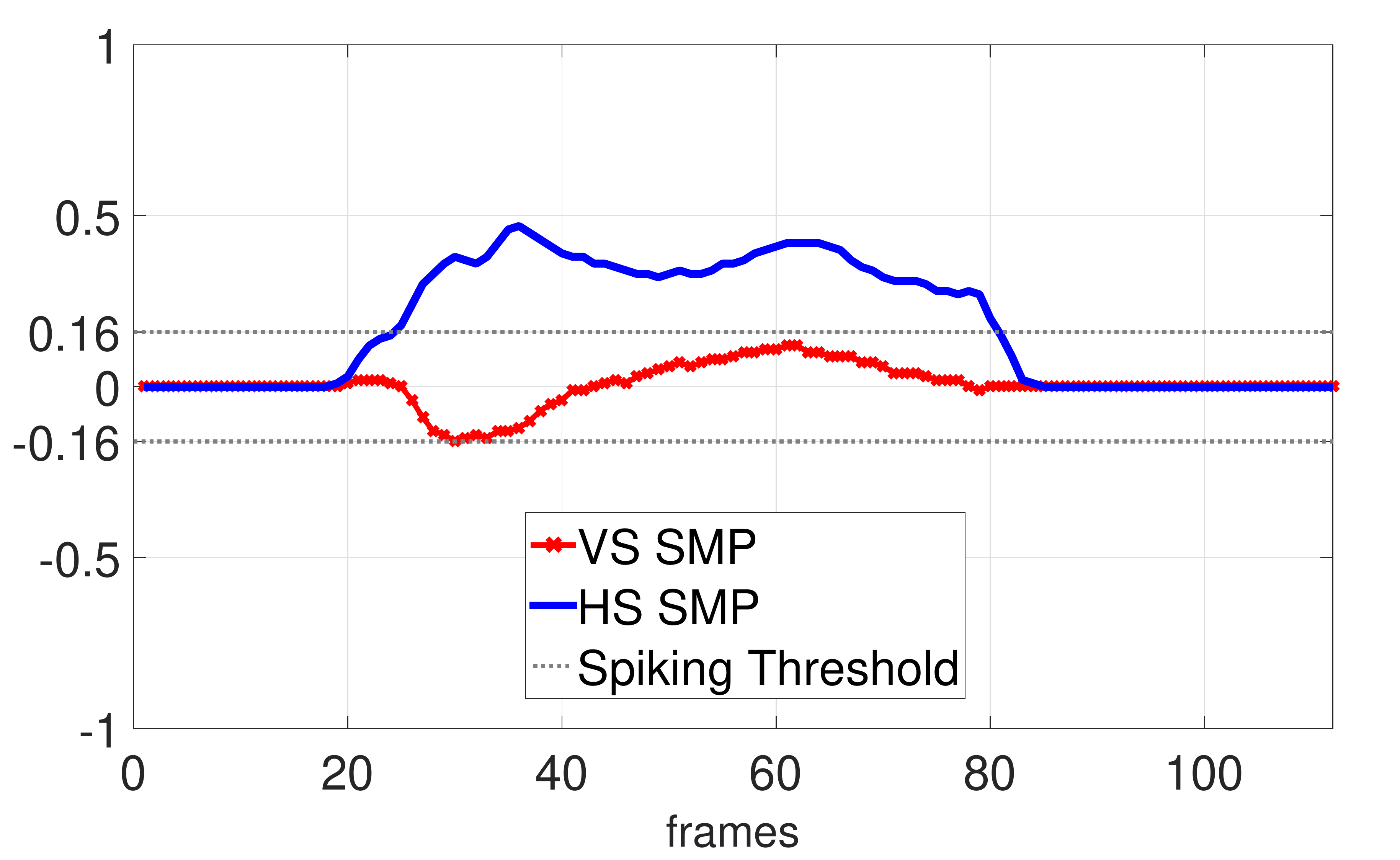}}
		\centerline{\scriptsize(d) a car translating rightward}
	\end{minipage}
	\vfill
	\vspace{0.05in}
	\begin{minipage}[t]{0.5\linewidth}
		\centering
		\centerline{\includegraphics[width=2in]{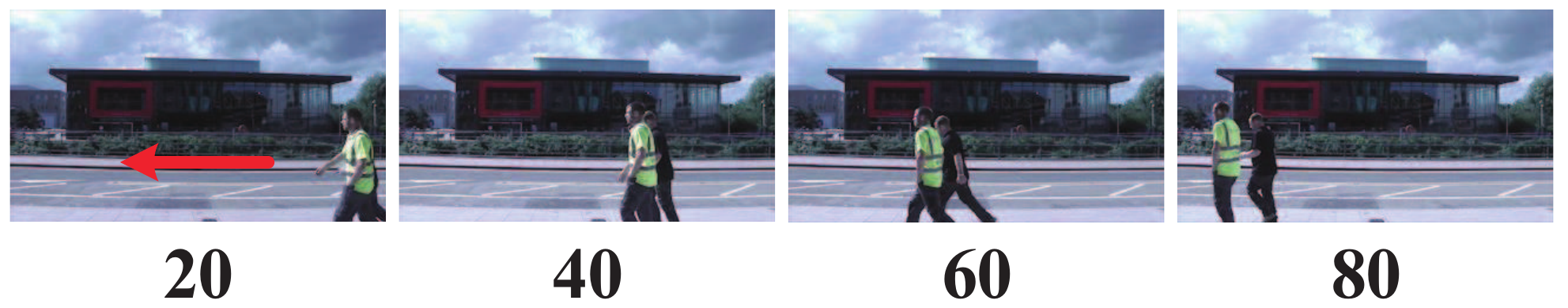}}
	\end{minipage}
	\hfill
	\begin{minipage}[t]{0.5\linewidth}
		\centering
		\centerline{\includegraphics[width=2in]{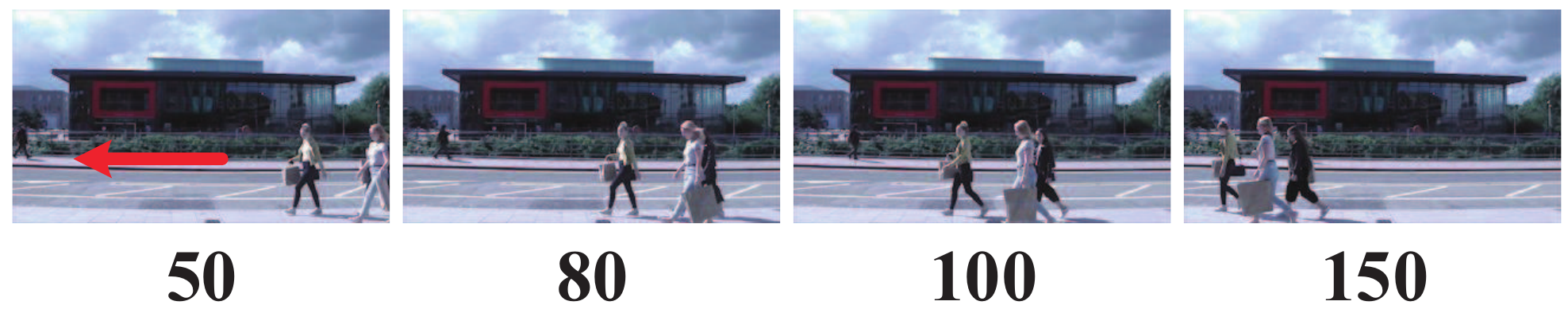}}
	\end{minipage}
	\vfill
	\begin{minipage}[t]{0.5\textwidth}
		\centering
		\centerline{\includegraphics[width=2.2in]{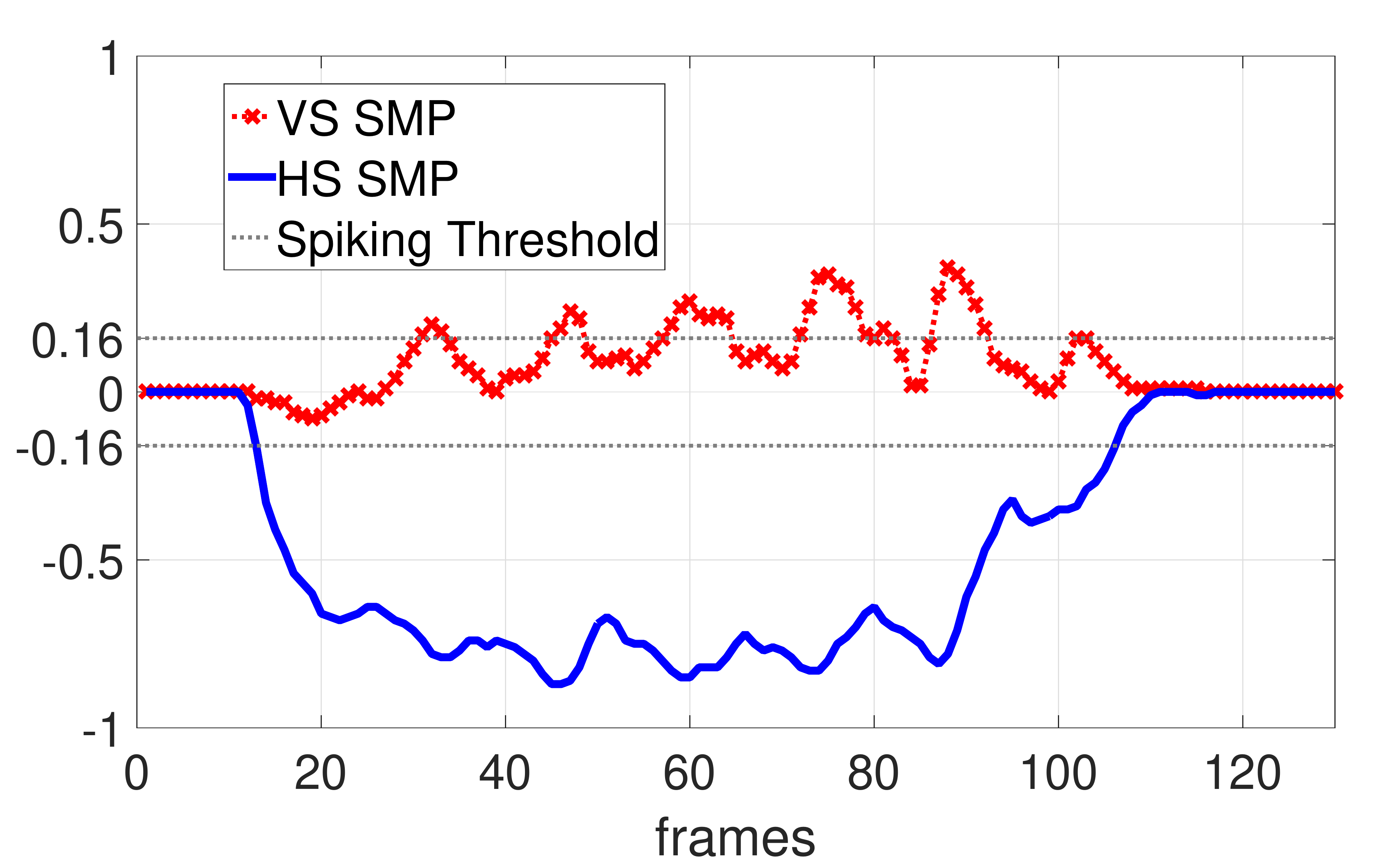}}
		\centerline{\scriptsize(e) a grouped people translating leftward}
	\end{minipage}
	\hfill
	\begin{minipage}[t]{0.5\textwidth}
		\centering
		\centerline{\includegraphics[width=2.2in]{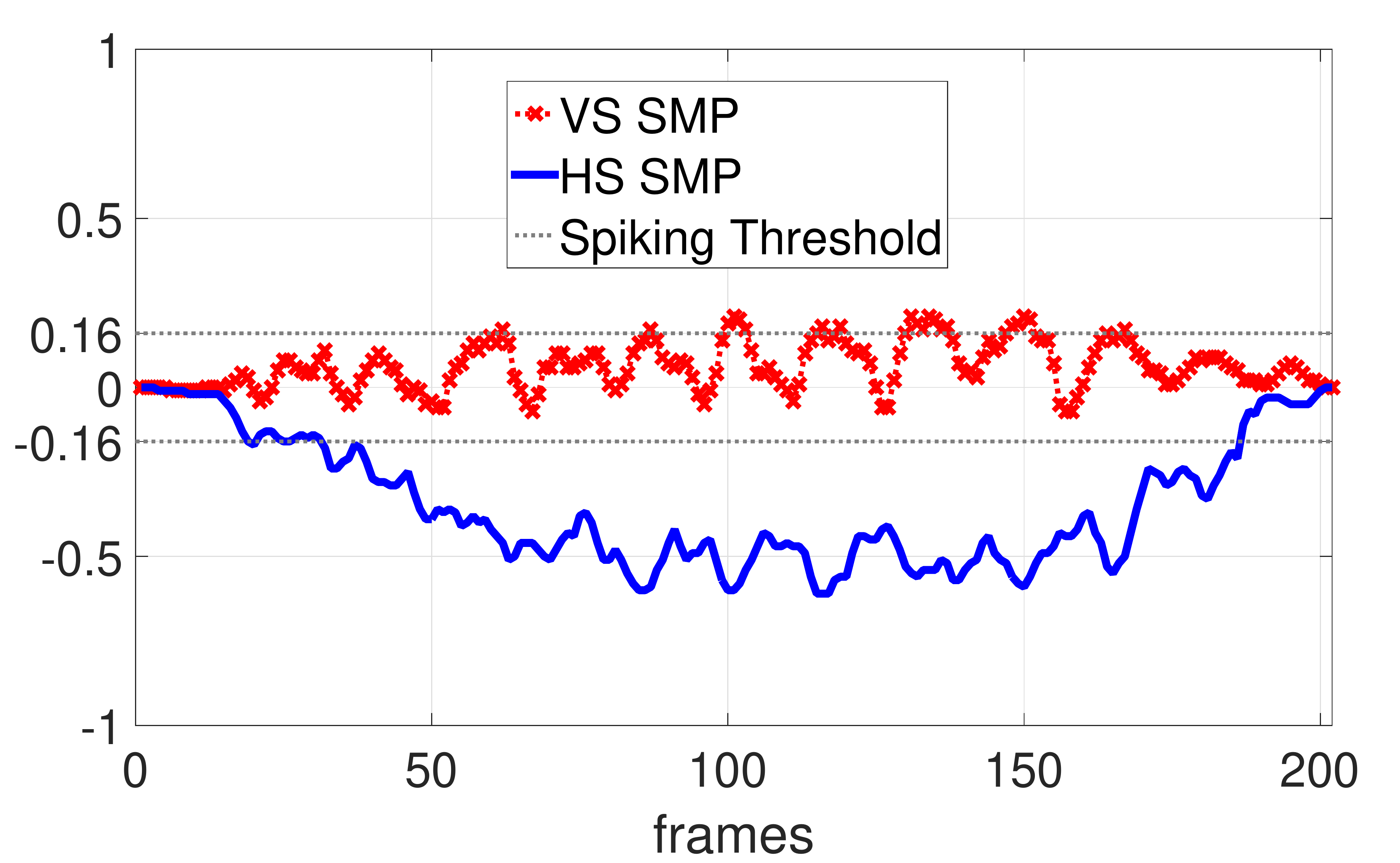}}
		\centerline{\scriptsize(f) a grouped people translating leftward}
	\end{minipage}
	\caption{The DSNN is challenged by translating stimuli in real physical scenes all with a visually cluttered and dynamic environment. All the translations are in horizontal directions. \textbf{The DSNN responds to preferred and null directional wide-field translations with positive and negative responses respectively; the HS system is highly activated}.}
	\label{real-world}
\end{figure}
In this subsection, we present the off-line experiments in real world scenarios. Compared with the synthetic stimuli tests, the degree of complexity of real physical scenes is relatively higher, including more environmental noise or irrelevant motion like windblown vegetation. We tested the proposed framework by horizontally translating movements embedded in two scenes: a campus avenue and a street view  -- as visually cluttered backgrounds, shown in Fig. \ref{real-world}.

In general, the results in Fig. \ref{real-world} demonstrate that the DSNN is able to detect all the wide-field translational motion in visually cluttered environments, which fulfills the requirements of a robust motion detector for real-world visual tasks. To be more specific, the useful motion cues, including direction and magnitude information of horizontally translating objects, are extracted from the busy backgrounds by the HS system of the DSNN, which are rigorously mapped by the positive and negative neural response for translations along preferred (rightward) and null (leftward) directions respectively. On the contrary, the neural response of VS system of the DSNN mainly remains at much lower level, below the spiking threshold. Fig. \ref{real-world}(a), \ref{real-world}(b), \ref{real-world}(c) and \ref{real-world}(d) demonstrate the DSNN well perceives the translating movements mixed with the background motion of windblown vegetation. Fig. \ref{real-world}(e) and \ref{real-world}(f) indicate that it can also detect the same directional translation of a group of objects. However, it is also very important to state that, since the proposed framework only detects translational motion across a wide-field of visual field mimicking the DSNs in the flies' visual system, it is not able to provide translational motion information locally for each individual translating agent without the segmentation and/or visual attention-based functions.

To conclude, the results of the off-line tests verify the usefulness and robustness of the DSNN framework for translational motion perception against either simple or complex backgrounds. The underlined functionalities explain the characteristics of DSNs in the fly's visual brain revealed by biologists. And importantly, the model represents both the speed response and contrast sensitivity to translating objects. In addition, the comparative experiments with two related translational motion detectors prove two advantages of the DSNN, i.e., its enhanced speed response to translating objects and more robust ability of filtering out irrelevant motion from relevant motion. In the next subsection, we will present the on-line robot experiments to investigate its potential in robotic vision applications.

\subsection{Robot tests}
\begin{figure}[t]
	\begin{minipage}[t]{0.32\linewidth}
		\centering
		\centerline{\frame{\includegraphics[width=1.3in]{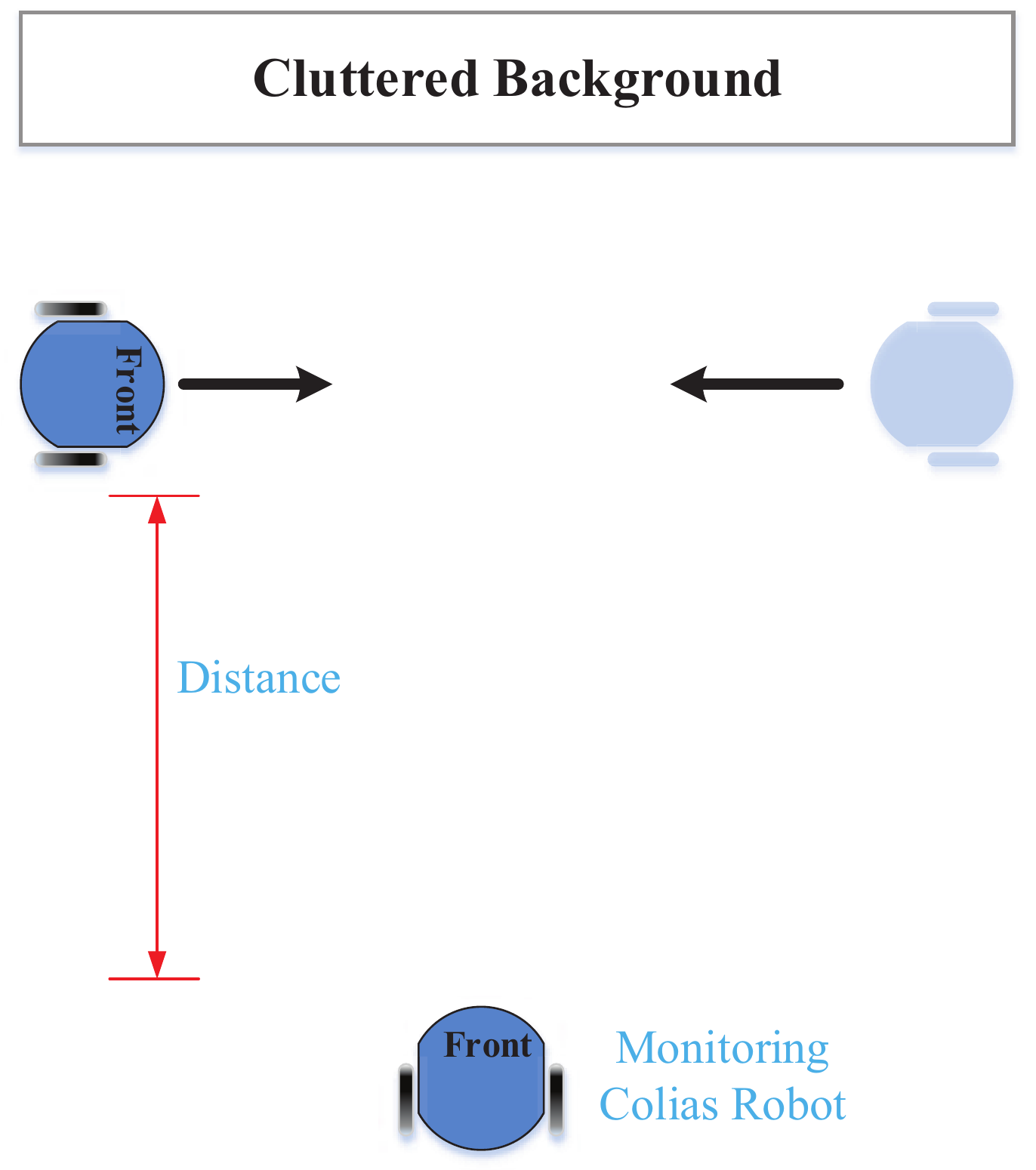}}}
		\centerline{\scriptsize(a) \textbf{translation} tests}
	\end{minipage}
	\hfill
	\begin{minipage}[t]{0.32\linewidth}
		\centering
		\centerline{\frame{\includegraphics[width=1.33in]{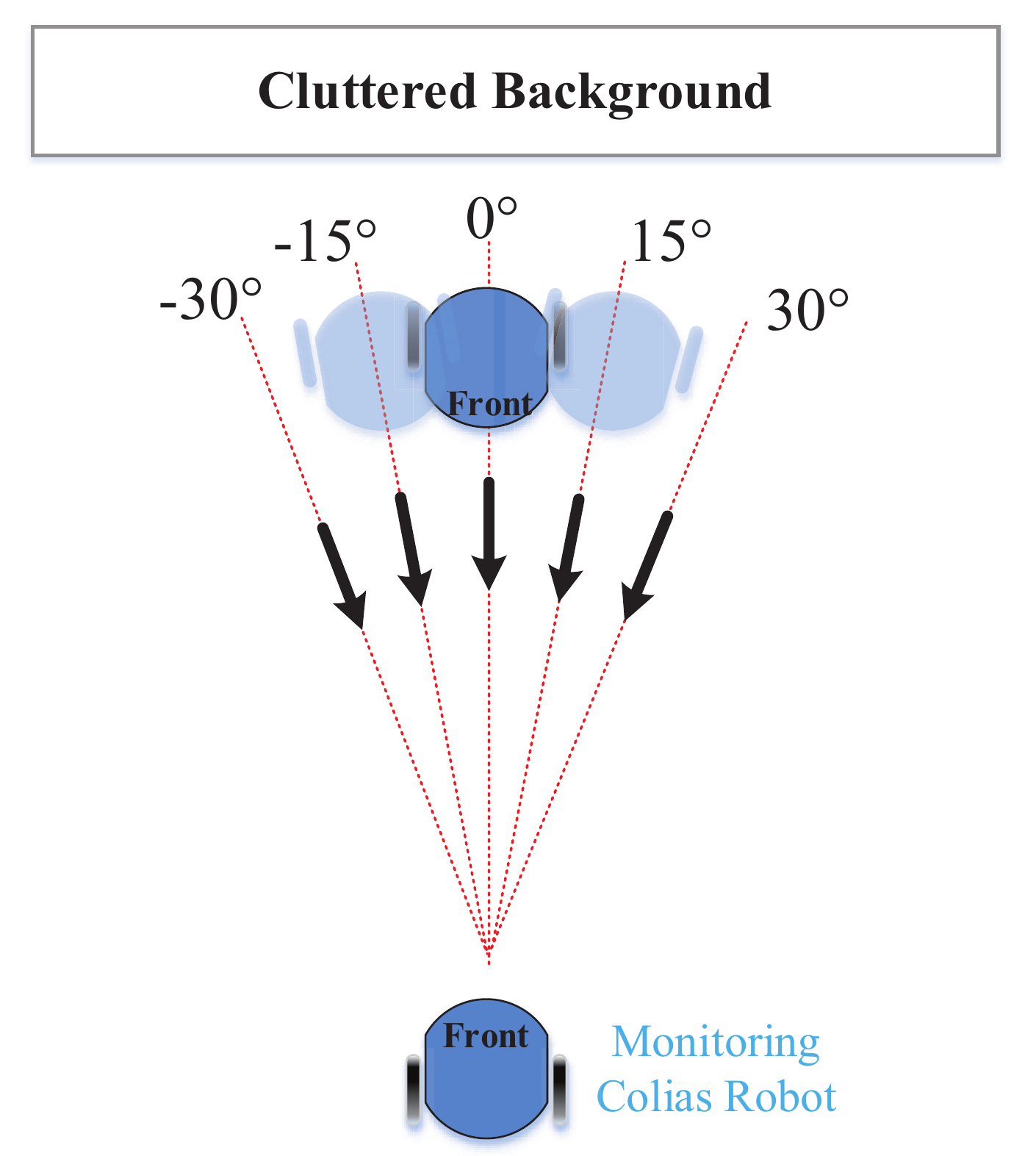}}}
		\centerline{\scriptsize(b) \textbf{angular-approach} tests}
	\end{minipage}
	\hfill
	\begin{minipage}[t]{0.32\textwidth}
		\centering
		\centerline{\frame{\includegraphics[width=1.28in]{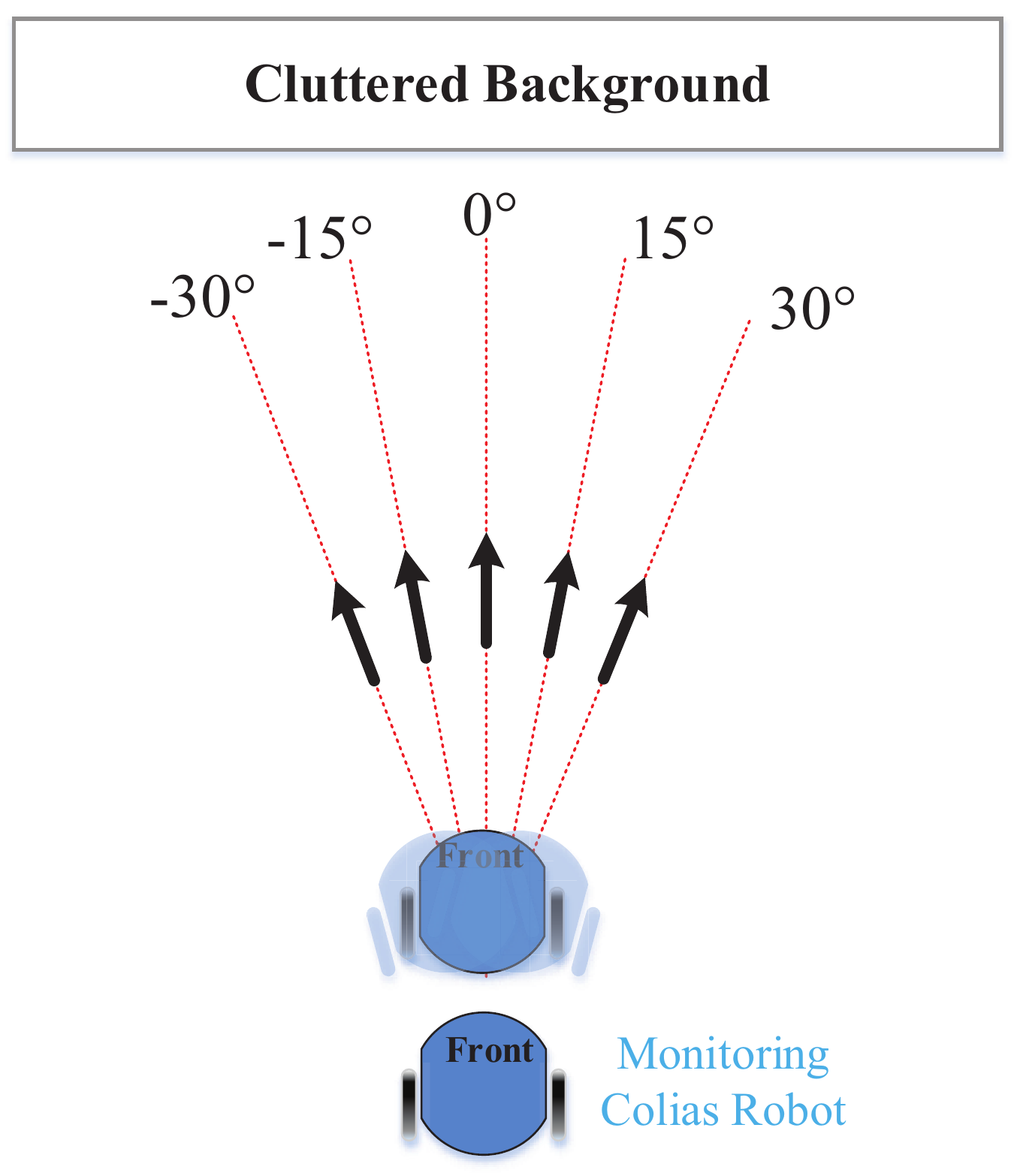}}}
		\centerline{\scriptsize(c) \textbf{angular-recession} tests}
	\end{minipage}
	\caption{The real-time robot experiments set-ups -- in all the tests, we collected the outputs of embedded DSNN from the monitoring Colias robot. Another Colias robot illustrated in Fig. \ref{colias}(c) was used as the visually moving stimuli. Dark arrows indicate motion directions.}
	\label{robot-exps-setting}
\end{figure}

In the last type of experiments, the DSNN was implemented in the Colias robot and tested in real time, for the purpose of evaluating its effectiveness and potential in robotic vision applications, along with deepening the understanding of its internal characteristics through systematic real-time trials. We designed two kinds of tests: the first was similar to the off-line tests to inspect its fundamental motion-detecting ability using general stimuli of approaching, receding and translating objects; the second sort involved systematic translation, angular-approach and angular-recession tests. The experimental settings are illustrated in Fig. \ref{robot-exps-setting}. It is necessary to state the small robot is only able to run on a 2D surface, so that we only investigated its HS system.

\begin{figure}[t]
	\begin{minipage}[t]{0.5\linewidth}
		\centering
		\centerline{\includegraphics[width=1.6in]{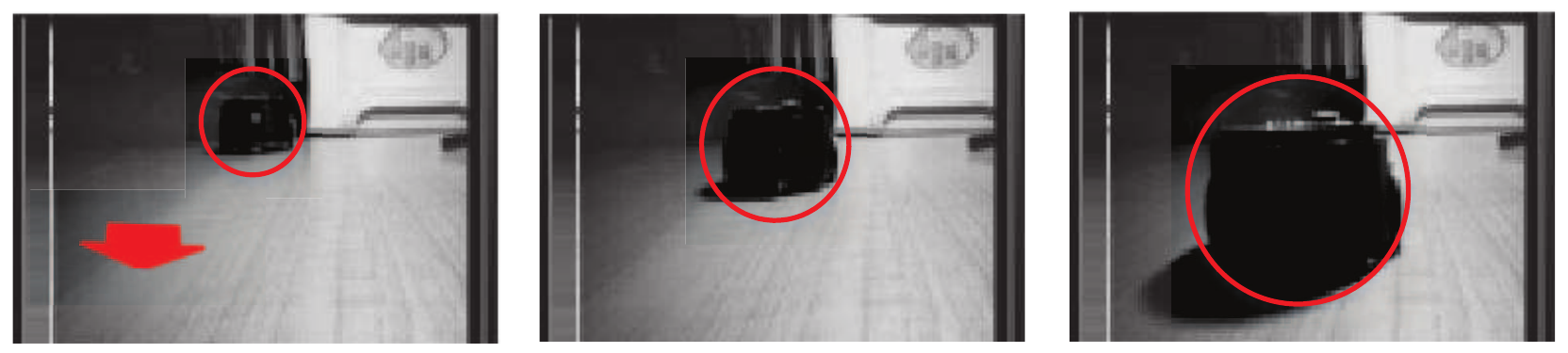}}
	\end{minipage}
	\hfill
	\begin{minipage}[t]{0.5\linewidth}
		\centering
		\centerline{\includegraphics[width=1.6in]{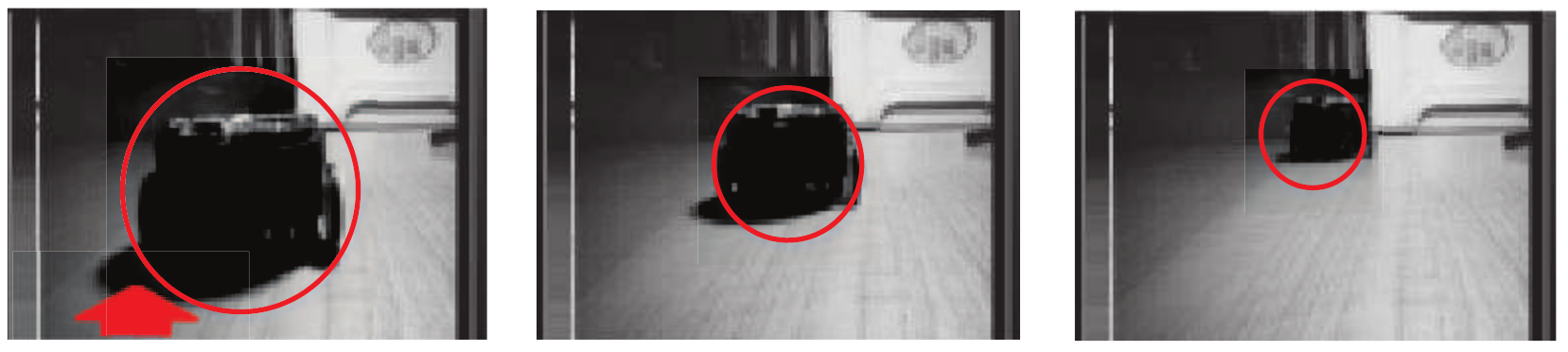}}
	\end{minipage}
	\vfill
	\begin{minipage}[t]{0.5\linewidth}
		\centering
		\centerline{\includegraphics[width=2.2in]{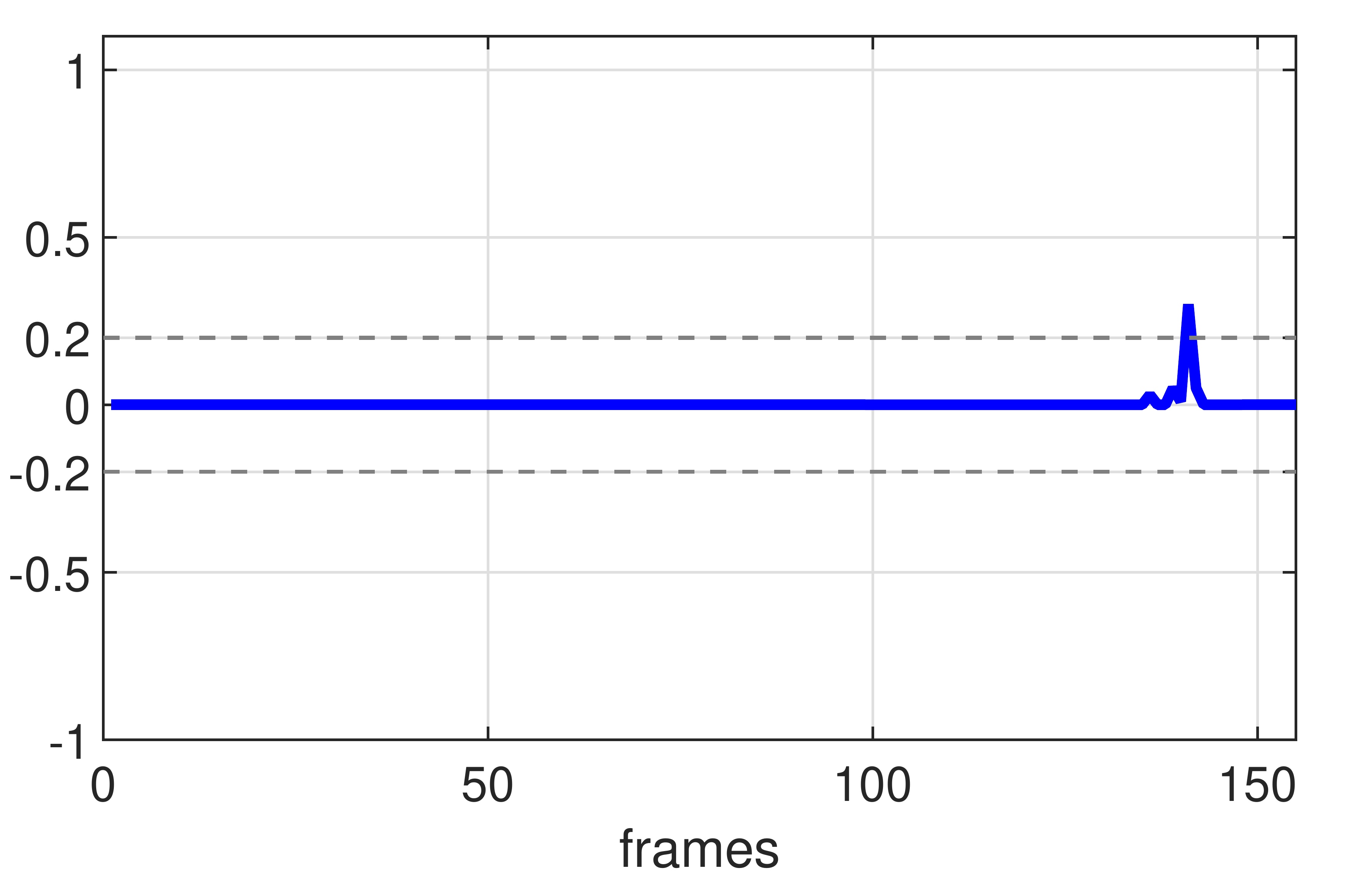}}
		\centerline{\scriptsize(a) a Colias robot \textbf{approaching}}
	\end{minipage}
	\hfill
	\begin{minipage}[t]{0.5\linewidth}
		\centering
		\centerline{\includegraphics[width=2.2in]{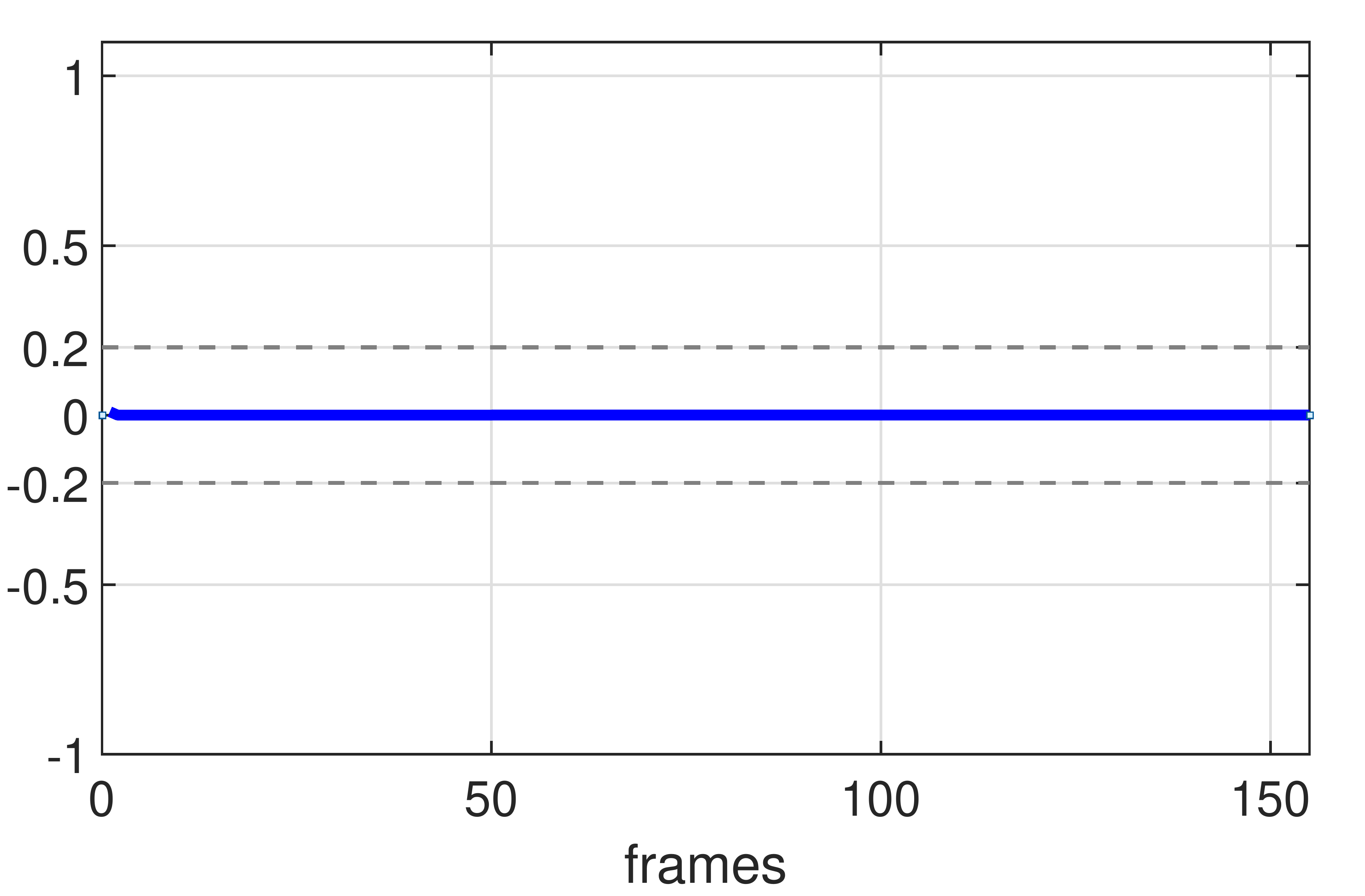}}
		\centerline{\scriptsize(b) a Colias robot \textbf{receding}}
	\end{minipage}
	\vfill
	\vspace{0.05in}
	\begin{minipage}[t]{0.5\textwidth}
		\centering
		\centerline{\includegraphics[width=1.6in]{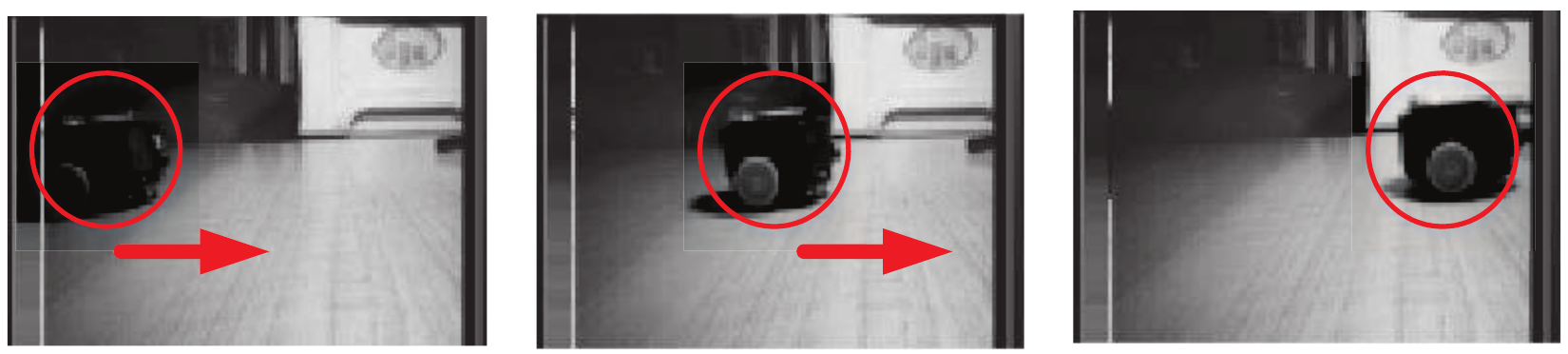}}
	\end{minipage}
	\hfill
	\begin{minipage}[t]{0.5\textwidth}
		\centering
		\centerline{\includegraphics[width=1.6in]{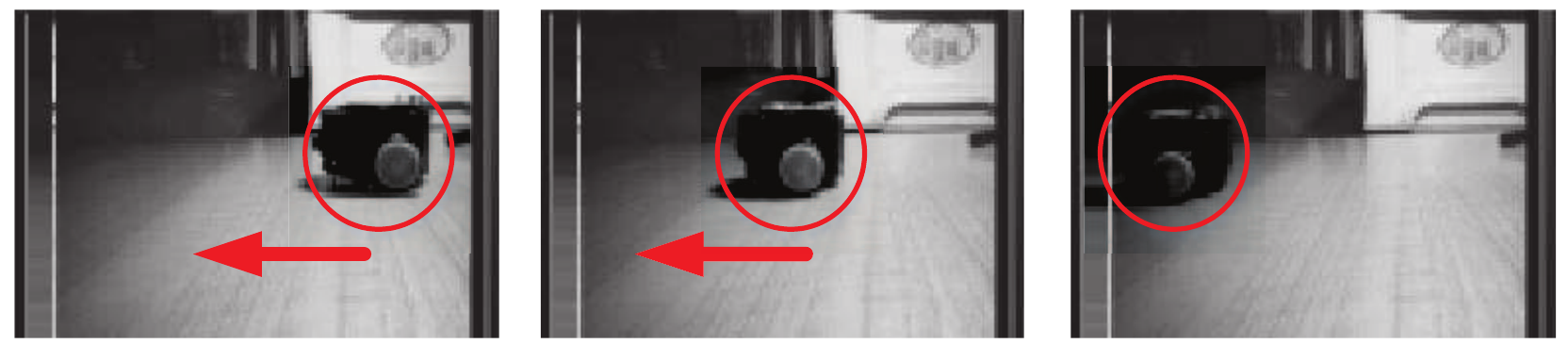}}
	\end{minipage}
	\vfill
	\begin{minipage}[t]{0.5\textwidth}
		\centering
		\centerline{\includegraphics[width=2.2in]{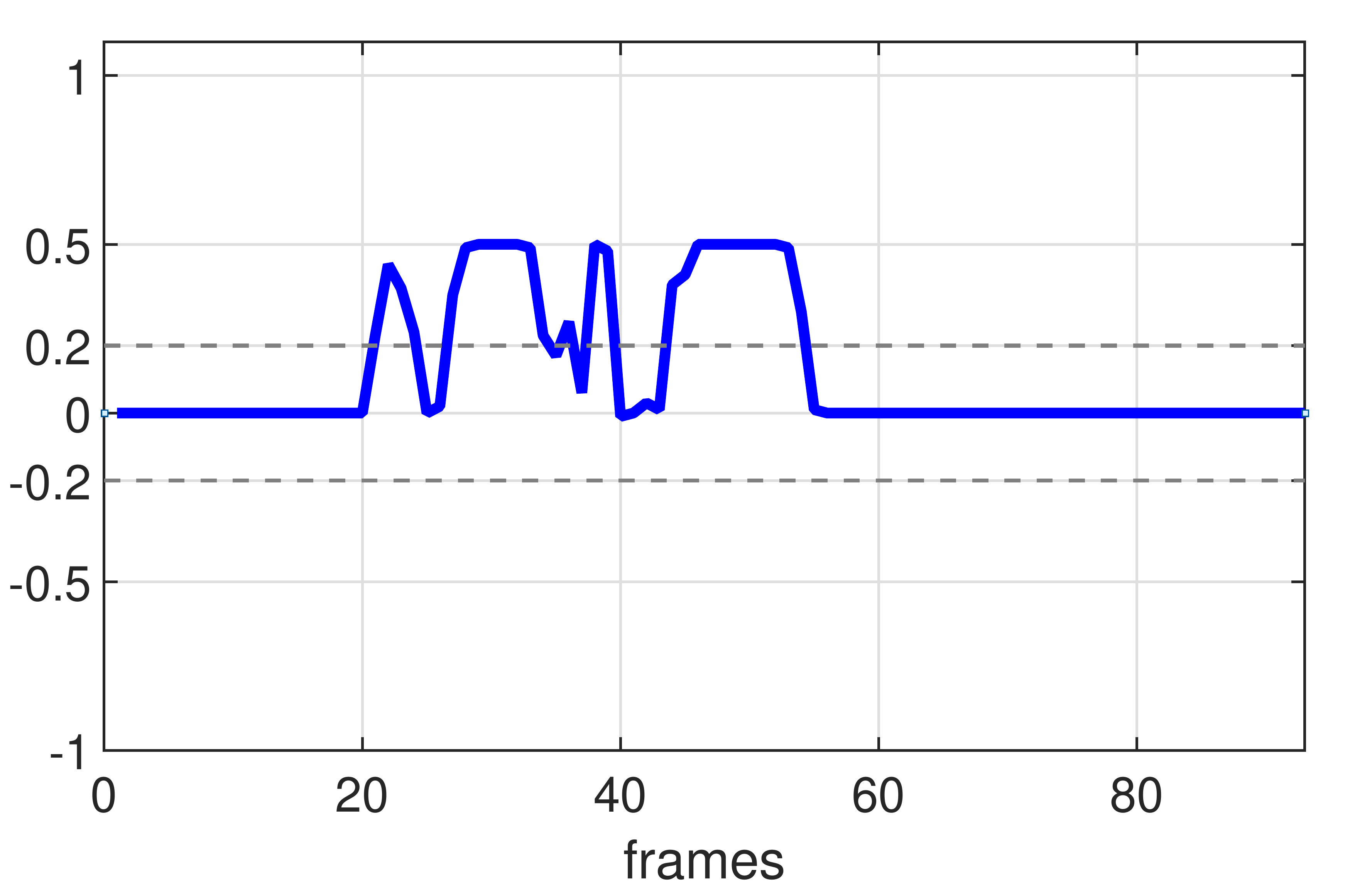}}
		\centerline{\scriptsize(c) a Colias robot \textbf{translating rightward}}
	\end{minipage}
	\hfill
	\begin{minipage}[t]{0.5\textwidth}
		\centering
		\centerline{\includegraphics[width=2.2in]{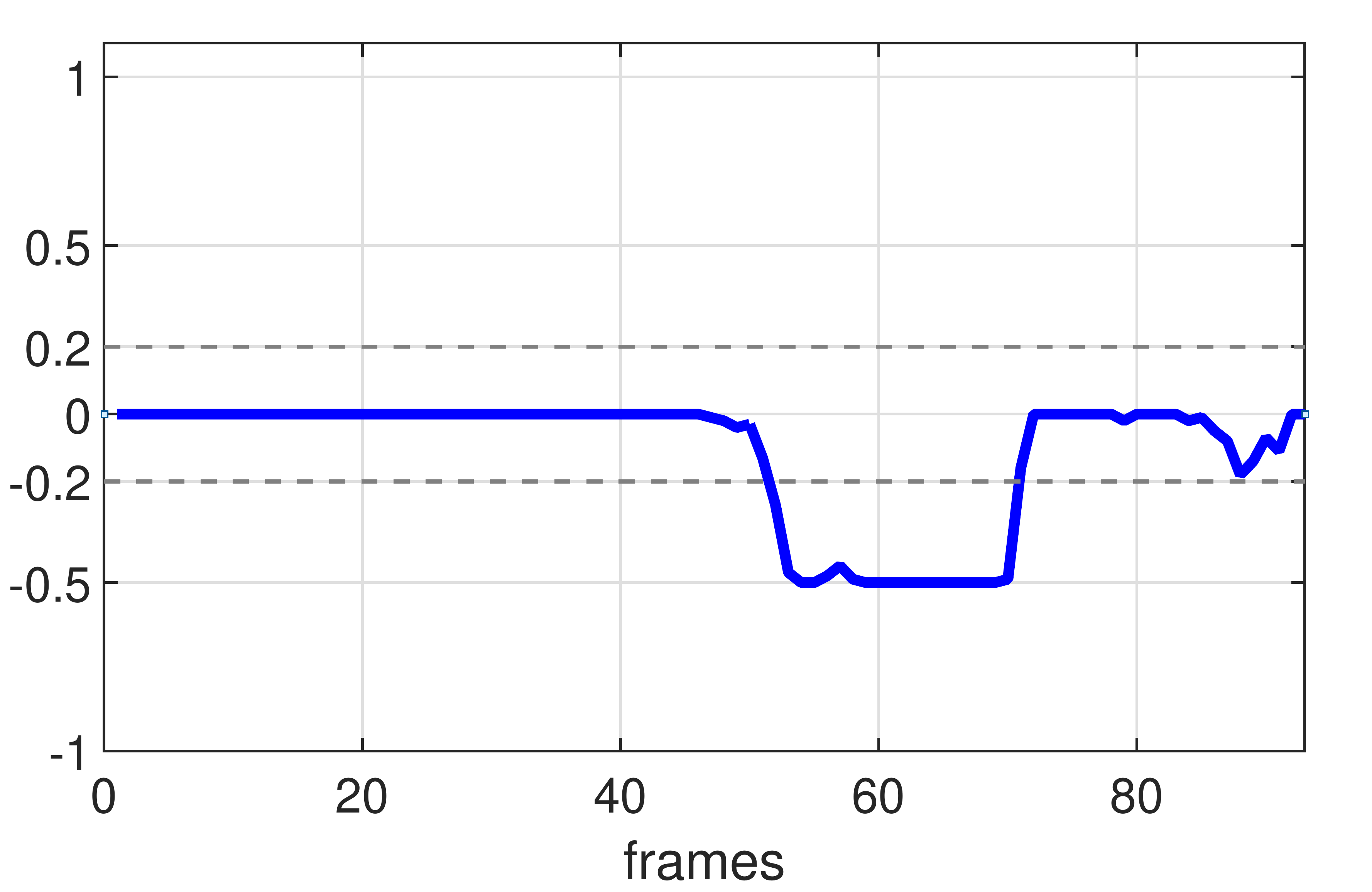}}
		\centerline{\scriptsize(d) a Colias robot \textbf{translating leftward}}
	\end{minipage}
	\caption{The embedded DSNN is challenged by an approaching, receding and translating Colias robot respectively. The examples of frontal-views captured by the monitoring Colias robot are shown at the top of neural response of the HS system. The spiking thresholds are set at $\pm0.2$. \textbf{The DSNN responds to translations rather than proximity and recession}.}
	\label{robot-general-tests}
\end{figure}

\paragraph{Tested by general visual stimuli}
In the first round of on-line robot tests, the DSNN implemented in the Colias robot was challenged by individual approaching, receding and translating objects, which are also very frequent visual stimuli for robots. Fig. \ref{robot-general-tests} illustrates the example of first-views from the monitoring Colias robot and the neural responses of the embedded DSNN, including SMPs and spikes of the HS system. Similarly, the embedded DSNN remains quiet during the whole course of either proximity or recession stimuli, i.e., movements in depth. On the other hand, it is rigorously activated by translating movements - the membrane potential is tuned to be positive for the rightward translation, and negative for the leftward translation. Satisfactory results with the on-line robot tests well match the outcomes of above off-line tests (Fig. \ref{simu-clean}, \ref{simu-nature-depth}, \ref{simu-nature-trans} and \ref{real-world}), which demonstrate the proposed framework can provide neuromorphic solutions to guide translational motion perception in autonomous robots.

\paragraph{Systematic translation tests}
In the second round of real time robot experiments, we looked deeper into its intrinsic properties of motion detection as an embedded vision system. The Colias robot with the on-board DSNN was challenged against systematic translating movements in visual clutter. More specifically, as illustrated in Fig. \ref{robot-exps-setting} (a), another Colias robot translated rightward across the visual field of the monitoring Colias robot, from different distances or at various linear-speed levels.
\begin{figure}[t]
	\begin{minipage}[t]{0.5\linewidth}
		\centering
		\centerline{\includegraphics[width=2.6in]{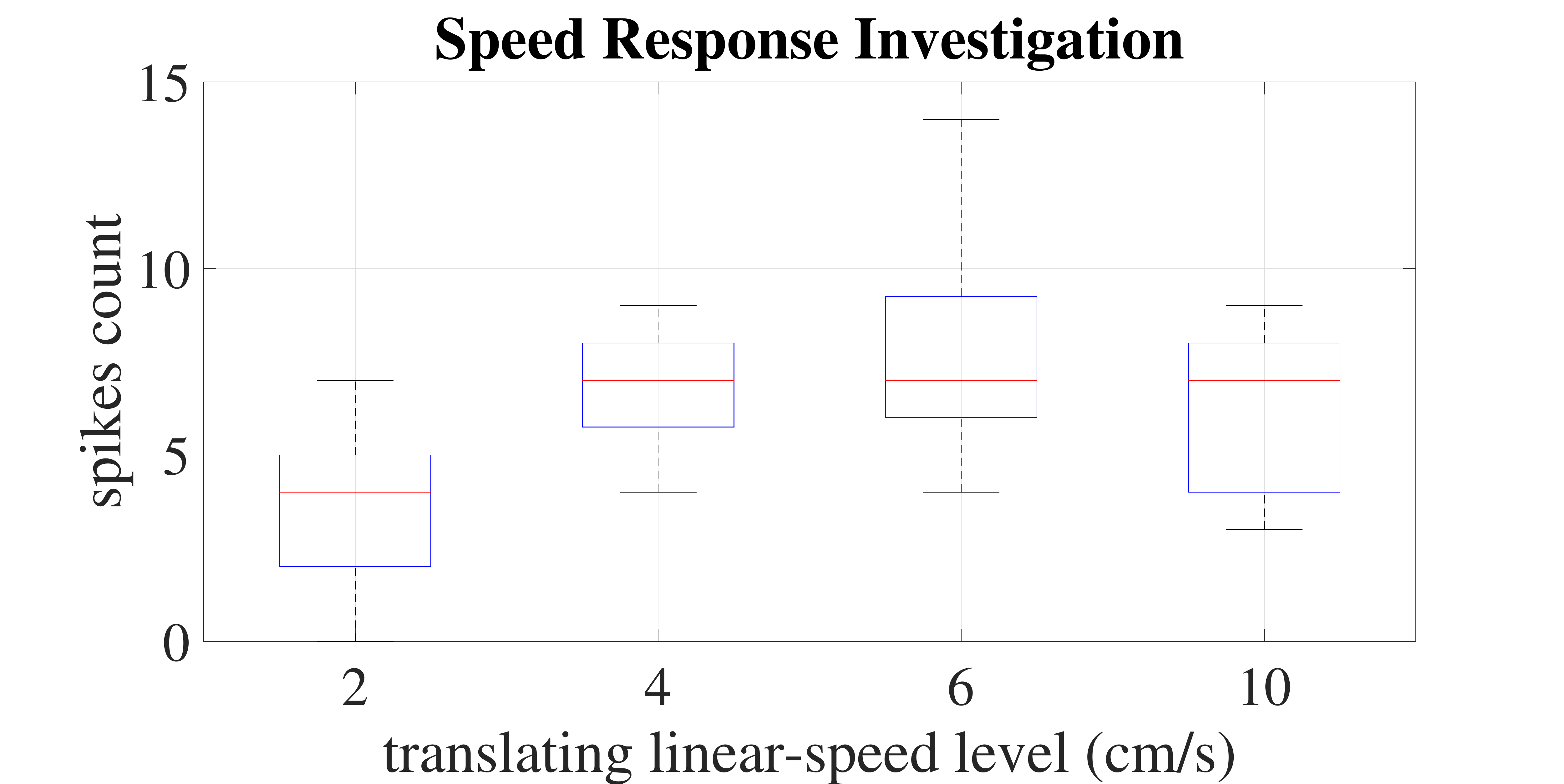}}
		\centerline{\scriptsize(a)}
	\end{minipage}
	\hfil
	\begin{minipage}[t]{0.5\linewidth}
		\centering
		\centerline{\includegraphics[width=2.6in]{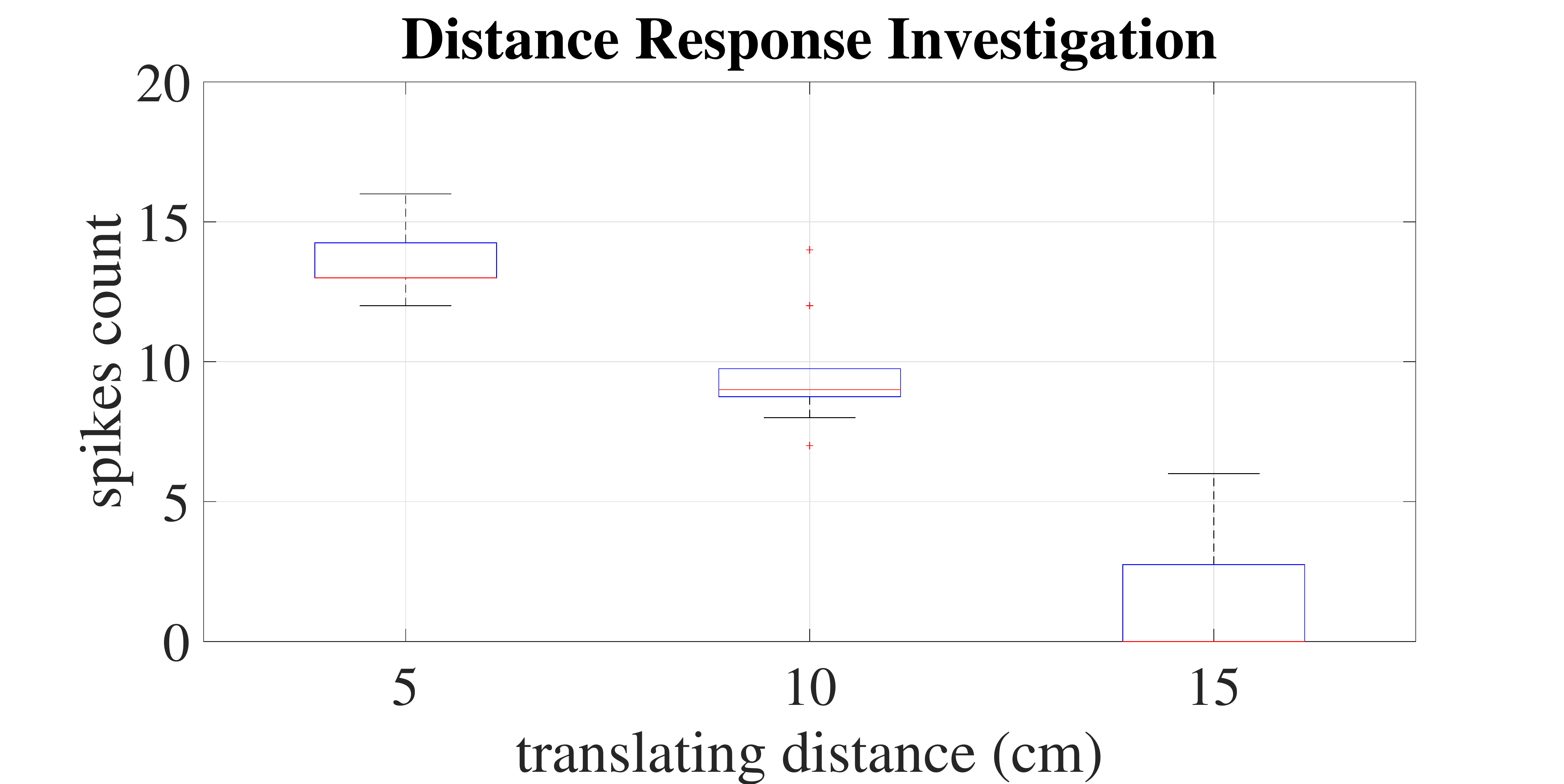}}
		\centerline{\scriptsize(b)}
	\end{minipage}
	\caption{The statistical results of real-time translation tests: (a) The distance between the stimulated Colias robot and the translating stimuli was fixed at 8 cm, whilst the translating linear-speed varied at 2, 4, 6 and 10 cm/s, each throughout 10 repeated tests. (b) The translating linear-speed was fixed at 6 cm/s, whilst the distance changed at 5, 10 and 15 cm, each throughout 10 repeated tests. \textbf{The embedded DSNN represents higher firing rates by translations from shorter distances or at faster linear-speeds}.}
	\label{robot-sys-trans}
\end{figure}

First, we examined if the embedded DSNN shows good speed response to translational motion as explored in the off-line tests (Fig. \ref{simu-sys-speed} and \ref{simu-sys-grayscale}). We accumulated the elicited spikes of the HS system during each translation process throughout repeated tests, which were all with a nearly identical translating time window. The statistical results shown in Fig. \ref{robot-sys-trans}(a) demonstrate that, tested from a fixed distance of $8$ cm, the spiking frequency of DSNN increases along with the translations speeding up, and then peaks around the translating linear-speed of roughly $6$ cm/s. Intriguingly, the spiking frequency is not continuously increasing, i.e., it declines after the peak. The results are in accordance with the selection of sampling distance between each combination of ON/OFF motion detectors and the number of directional connections for each polarity cell in the dual-pathways. As mentioned above, such a structure improves the speed tuning of the HRC-detectors based translational motion sensitive systems. Its functionality nevertheless is restricted by the predefined parameters of the ensembles of motion detectors.

Second, we examined the influence of distance on the spiking rate. Intuitively, the results in Fig. \ref{robot-sys-trans}(b) represent the spiking rate shrinks dramatically as the distance between the translating and monitoring robots increases, i.e., the peak and valley of firing rate occurs from the smallest and largest distances respectively. Since the DSNs in the fly's visual system were well known to be mostly sensitive to wide-field movements in the visual field \cite{Schnell-widefield2012,Fisher-L3}, it is conceivable that the DSNs are not able to smoothly recognize the translating objects of a very small size, similarly to the situation that translations happen far away from the visual field. As mentioned in Section \ref{introduction}, there is another group of visual neurons, in the insects' visual pathways, specialized in the small targets movement detection \cite{Wiederman-2008,Circuit-correlation2013,STMD-IJCNN}.

\paragraph{Angular approach and recession tests}
\begin{figure}[t]
	\begin{minipage}[t]{0.5\linewidth}
		\centering
		\centerline{\includegraphics[width=2.6in]{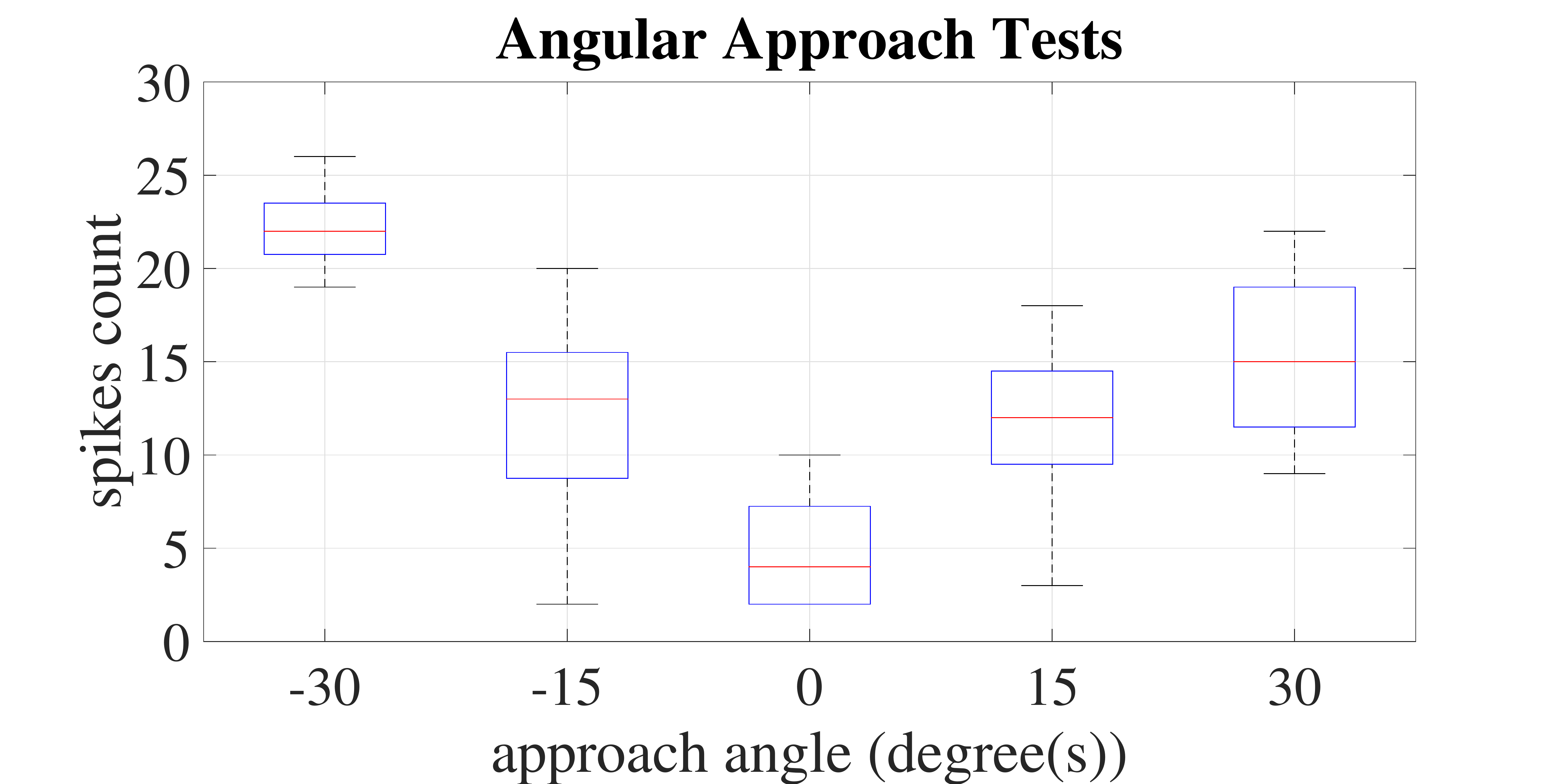}}
		\centerline{\scriptsize(a)}
	\end{minipage}
	\hfill
	\begin{minipage}[t]{0.5\linewidth}
		\centering
		\centerline{\includegraphics[width=2.6in]{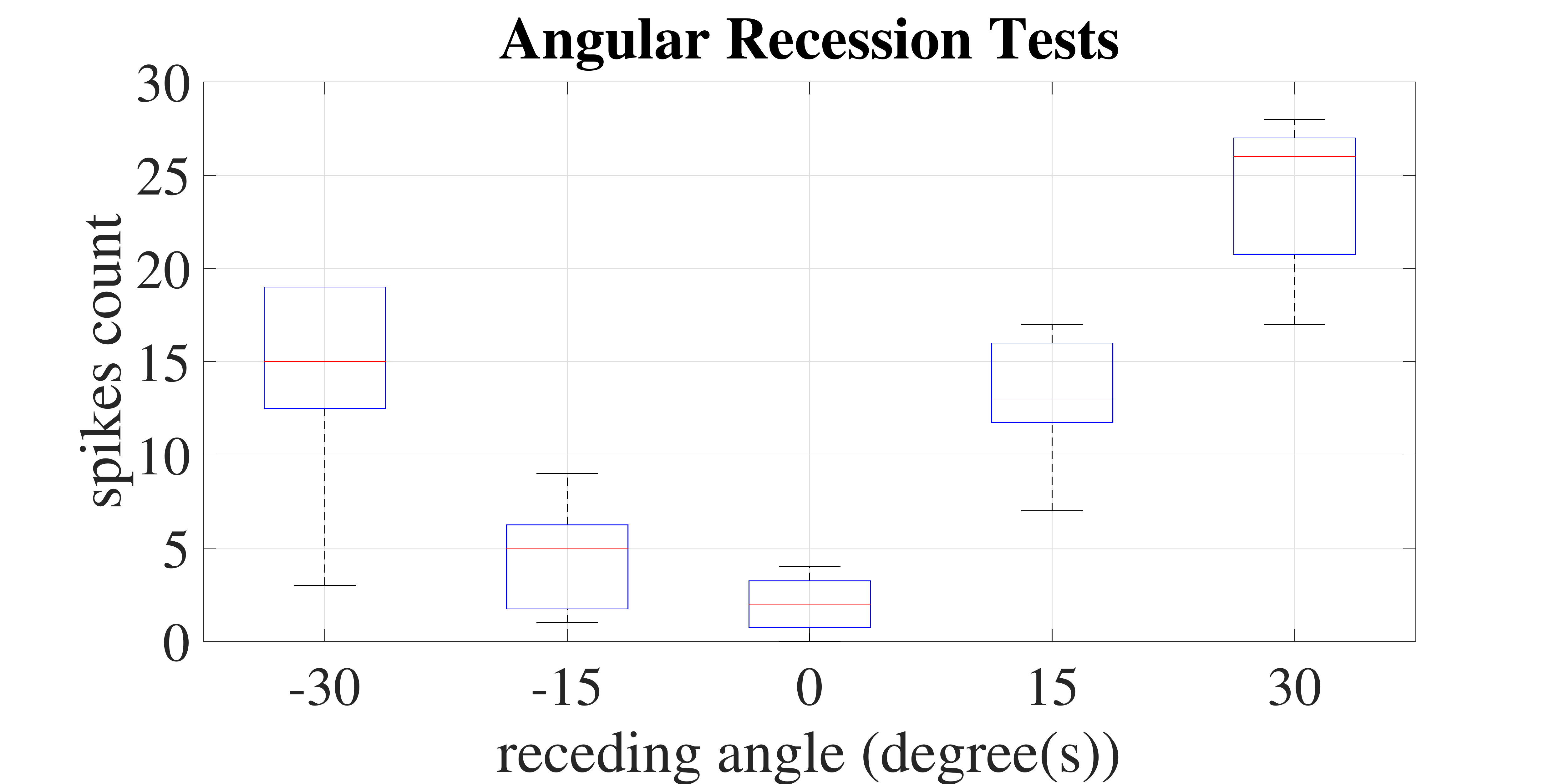}}
		\centerline{\scriptsize(b)}
	\end{minipage}
	\caption{The statistical results of real-time angular-approach (a) and angular-recession (b) tests: each kind of tests involved movements from five angles respectively, each throughout 10 repeated tests. \textbf{The embedded DSNN spikes at the lowest frequencies by movements of direct-approaching and direct-recession}.}
	\label{robot-sys-ar}
\end{figure}
In the third round of systematic robot experiments, we also challenged the embedded DSNN with angular approaching and receding stimuli, as shown in Fig. \ref{robot-exps-setting}(b) and \ref{robot-exps-setting}(c), in order to compare its functionality with the looming sensitive visual neural networks. Fig. \ref{robot-sys-ar} illustrates the statistical results of the spiking frequency under repeated angular-approaching and angular-receding courses. Concretely, the embedded DSNN spikes at the lowest rates with the direct approaching and receding stimuli from the angle of $0^{\circ}$. On the other hand, it is rigorously activated by the angular approaching and receding movements from other angles - the spike frequency gets higher if the angle of proximity and recession increases. As a matter of fact, for the monitoring Colias robot, the left angular approaching (angles $-30^{\circ}$ and $-15^{\circ}$) and the right angular approach (angles $15^{\circ}$ and $30^{\circ}$) gave rise to the rightward and leftward translating features respectively, and the opposite for the movements of angular recession. Interestingly, because a partially balanced structure of each pairwise ON-ON and OFF-OFF motion detectors within the dual-pathways, making the DSNN to respond more strongly to motion along the preferred versus null directions, the statistical results also indicate higher spiking frequency for the angular approaching/receding from the left/right sides of the monitoring Colias robot respectively. The robot experiments verify that the embedded DSNN mainly possesses the sensitivity to translational motion over other kinds of movements.

Interestingly, with similar ideas, we tested the looming (or collision) sensitive neuron models by the similar angular approach tests in our previous research \cite{IROS-LGMDs}. The results presented in Fig. \ref{robot-sys-ar} demonstrate an opposite but complementary performance of the embedded DSNN relatively to the looming detectors, which spike at the highest rate by the direct approaching. Therefore, combining the functionality of the two bio-plausible models can benefit the creation of more competitive motion sensitive systems.

\section{Conclusion and future work}
\label{conclusion}
In this article, we propose a directionally selective neural network for studying the characteristic of direction selective neurons in the fly's visual system, and mimicking the fly's preliminary motion-detecting pathways. DSNs are with unique sensitivity and direction-selectivity to wide-field translational motion. Compared with the former bio-inspired translational motion detectors, like the elementary motion detectors, the proposed framework splits motion information into ON and OFF visual pathways for parallel computation, encoding light-on and light-off responses separately. It finally integrates local excitations from four groups of lobula plate tangential cells, each one possessing certainly directional selectivity to form the horizontally and vertically sensitive systems. Importantly, the proposed computational architecture explains underlying fly's physiology. Through this modeling study, we emphasized the effectiveness of spatiotemporal computations for improving the velocity tuning of translational motion detectors by building ensembles of same-sign (ON-ON/OFF-OFF) polarity cells within the dual-pathways. We also demonstrated a temporal FDSR mechanism with biological plausibility, which contributes to filter out irrelevant motion from a visually cluttered and dynamic environment. The specific characteristic of direction selective neurons in the fly's visual system have been fully achieved by this computational model, and been demonstrated through our	 systematic and comparative experiments, ranging from off-line tests with synthetic and real-world scenarios to on-line robot tests.

This work opens several directions for future research. First, the above experiments give strong evidence that the functionality of the proposed DSNN can provide the perfect complement to the former collision-detecting neural networks (like LGMDs) with a similar structure of the separated ON/OFF pathways. Therefore, it is possible to construct a hybrid visual model integrating the functionality of direction and collision sensitive neural networks, both inspired by insects physiology, for motion perception of more complex scenarios. Moreover, its computational simplicity and robustness, as an embedded vision system validated by the real-time robot experiments, also allow us to extend the DSNN to the higher level of behaviors, simulating the fly's motion tracking and fixation behaviors, which may benefit various vision-based tasks in swarm robotics.

\section*{Acknowledgment}
This work was supported by the grants of EU Horizon 2020 project STEP2DYNA(691154). We thank Dr. Cheng Hu for the hardware set-ups of Colias robots.

\section*{References}

\bibliography{qinbingbibfile}

\end{document}